\documentclass[runningheads]{llncs}
\usepackage{eccv}
\usepackage{eccvabbrv}
\usepackage{graphicx}
\usepackage{booktabs}
\usepackage[accsupp]{axessibility}
\usepackage{hyperref}
\usepackage{orcidlink}
\usepackage{multirow}
\usepackage{colortbl}
\usepackage{xcolor}

\begin{document}
\title{LinStereo: Linear-Complexity Global Attention for Multi-Scale Iterative Stereo Matching}
\titlerunning{LinStereo: Linear-Complexity Global Attention for Stereo Matching}
\author{Yiran Wang\thanks{Corresponding author.} \and
Oliver Turner \and
Viorela Ila\thanks{Project lead.}}
\authorrunning{Y. Wang et al.}
\institute{Australian Centre for Robotics (ACFR), School of Aerospace, Mechanical and Mechatronic Engineering (AMME), Faculty of Engineering, The University of Sydney, Sydney, NSW, Australia}

\maketitle
\setcounter{footnote}{0}

\begin{abstract}
Existing Vision Foundation Model (VFM)-based iterative stereo pipelines under-exploit three information pathways: multi-scale backbone features are collapsed into single-level correlations, geometric priors remain untapped at initialization, and context propagates only locally. These gaps widen under degraded photometric cues, making underwater scenes a stringent generalization test. To address this, we propose LinStereo, built upon Depth Anything V3, whose core is a Position-Aware Linear Attention (PALA) module that replaces local recurrence with global aggregation at linear cost, propagating reliable estimates from well-matched regions into degraded areas while preserving disparity structure. PALA is made effective by two enabling components: Hierarchical Semantic Cost Volumes (HSCV), which supply scale-aligned correlations from the VFM feature hierarchy, and a Depth Prior Initialization (DPI) that converts monocular depth into a metrically calibrated warm start. LinStereo achieves state-of-the-art-level accuracy on standard benchmarks and strong cross-domain generalization, particularly on underwater scene where severe photometric degradation makes stereo matching particularly challenging, attaining the best overall accuracy with consistent gains (\textbf{28\%} lower AbsRel on TartanAir-UW, \textbf{26\%} on SQUID, a real-world underwater dataset).
\end{abstract}

\section{Introduction}
\label{sec:intro}

Stereo depth estimation recovers dense metric 3D geometry from calibrated image pairs and is fundamental for vision-based navigation, manipulation, and inspection, where both accuracy and inference speed directly affect performance~\cite{hartley2004multiple,geiger2012we}. Depth estimation remains difficult in textureless, occluded, or repetitive regions, which require strong contextual reasoning under limited computation. These challenges intensify underwater, where wavelength-dependent attenuation and volumetric scattering degrade visual signals, and limited annotated subsea data causes models trained on synthetic or in-air datasets to suffer significant performance drops~\cite{akkaynak2018revised,berman2020underwater,lv2025uwstereo}.

The core difficulty in dense stereo depth estimation lies in establishing reliable
correspondences over large, ambiguous regions while keeping computation
tractable. Some approaches construct explicit cost volumes and regularize
them with 3D aggregation
networks~\cite{chang2018pyramid,xu2023igev}; others adopt iterative
refinement, using a recurrent operator to progressively refine depth estimates by indexing a correlation volume~\cite{lipson2021raftstereo,wang2024selective}. The iterative
family has become dominant and has been further strengthened by replacing
conventional encoders with vision foundation model (VFM) backbones to
improve cross-domain robustness~\cite{wen2025foundationstereo,jiang2025defom}
and by fusing monocular depth priors to regularize ill-posed
regions~\cite{yao2025diving,bartolomei2025stereo}. These efforts have
substantially raised the accuracy ceiling, yet they focus on improving the
inputs to the update loop; the update mechanism itself has
received less attention. Crucially, VFM backbones provide semantically rich multi-scale features and implicit geometric priors for global reasoning, yet their effective use within the update loop remains underexplored.

We observe a growing mismatch between the capabilities of modern backbones and what the iterative update loop actually exploits. Even in recent VFM-based pipelines, three bottlenecks persist: updates reason primarily over local neighborhoods, multi-scale features are often reduced to single-level correlations, and the backbone’s geometric priors remain largely unused at initialization. Although recent works address some of these aspects individually~\cite{xu2023igev,yao2025diving,bartolomei2025stereo}, as backbone capacity continues to grow, the backbone–update interface remains a key bottleneck.

\textbf{LinStereo} closes this gap with a redesigned decoder whose \textbf{core is Position-Aware Linear Attention (PALA)}: a global, linear-complexity update operator that, unlike local recurrent operators~\cite{lipson2021raftstereo,wen2025foundationstereo}, propagates information across the entire feature map at every refinement step. Two components make this global aggregation effective: \textbf{Hierarchical Semantic Cost Volumes (HSCV)} supply each refinement level with a correlation volume built from the VFM's matching-scale features, giving PALA semantically aligned matching evidence; and \textbf{Depth Prior Initialization (DPI)} converts the backbone's monocular depth into a disparity-space initialization, so PALA refines from a geometrically plausible state. LinStereo is thus a single PALA-centred decoder enabled by HSCV and DPI, rather than three independent modules.

These components address complementary aspects of the backbone--update
interface and yield consistent improvements across all settings
(Sec.~\ref{sec:experiments}).
LinStereo achieves state-of-the-art accuracy among methods trained on
comparable data on standard stereo benchmarks and demonstrates strong
cross-domain generalization, particularly in underwater scenes with
severe photometric degradation, achieving \textbf{28\%} lower AbsRel on
TartanAir-UW~\cite{wang2020tartanair} and \textbf{26\%} on
SQUID~\cite{berman2020underwater}.

The main contributions of this work are summarized as follows:
\begin{itemize}
\item \textbf{LinStereo}, an iterative stereo framework whose core is \textbf{PALA}, a
position-aware linear-attention operator that replaces local recurrence with global
aggregation at linear complexity. PALA is paired with \textbf{HSCV}, a hierarchical
cost-volume design that contributes both a strategy for using the backbone's
multi-scale features and a new structure for each per-scale volume; and with
\textbf{DPI}, which provides a warm start for refinement from monocular depth. Together, these reach
state-of-the-art-level accuracy with significantly fewer refinement iterations at linear
per-iteration cost.
    
    \item Trained only on SceneFlow with no underwater data, LinStereo
    achieves state-of-the-art performance on underwater stereo benchmarks
    without domain-specific adaptation, attaining the best overall accuracy, demonstrating strong
    cross-domain generalization under severe photometric degradation.
\end{itemize}

The paper is complemented by open-source code and a new evaluation dataset with dense disparity ground truth and realistic subsea optical modeling, as described in the supplementary multimedia.

\section{Related Work}

\paragraph{Stereo Depth Estimation.}
Cost volume-based methods~\cite{kendall2017end,chang2018pyramid,zhang2019ga,guo2019gwcnet,shen2022pcwnet,xu2024acvnet} build explicit 3D/4D volumes regularized by 3D convolutions, but their cost scales with disparity range and single-scale construction limits multi-level semantic integration.
RAFT-Stereo~\cite{lipson2021raftstereo} introduced an iterative alternative: a recurrent operator repeatedly queries a fixed correlation volume to refine disparity through local updates.
Subsequent works enrich this paradigm with cascaded correlation~\cite{li2022practical}, geometry-encoded volumes~\cite{xu2023igev}, and frequency-adaptive units~\cite{wang2024selective}, yet the update operator remains spatially local, typically requiring tens of iterations for information to propagate across the image. \\
This locality becomes increasingly significant as vision foundation models (VFMs) enter the pipeline.
FoundationStereo~\cite{wen2025foundationstereo}, DEFOM-Stereo~\cite{jiang2025defom}, MGStereo~\cite{yao2025diving}, and Stereo Anywhere~\cite{bartolomei2025stereo} leverage pretrained encoders~\cite{oquab2024dinov2,yang2024depthanything,yang2024depthanythingv2} to extract globally rich features that substantially improve zero-shot generalization, yet these features are still consumed through a local recurrent loop, which creates an information utilization gap between backbone capacity and update reach.
Our work targets this mismatch, enabling the iterative pipeline to better exploit backbone representations and converge in fewer iterations.

\paragraph{Efficient Stereo Depth Estimation.}
Reducing inference cost while maintaining accuracy is a central challenge in stereo depth estimation.
Prior work primarily simplifies the architecture itself:
pruning the disparity search space~\cite{khamis2018stereonet,xu2021bgnet,shamsafar2022mobilestereonet,duggal2019deeppruner},
replacing 3D cost aggregation with lightweight 2D alternatives~\cite{xu2020aanet,xu2024acvnet,bangunharcana2021coex,xu2023cgistereo,wang2025adstereo,guo2025lightstereo},
or eliminating explicit volumes entirely~\cite{tankovich2021hitnet}.
While achieving impressive speed, such simplifications can limit accuracy in challenging regions.
Our method pursues efficiency from a complementary direction: rather than simplifying the architecture, we reduce the number of refinement iterations required for convergence, preserving rich contextual modeling while lowering inference time.
\paragraph{Underwater Depth Estimation.}
Underwater environments amplify the challenges of {stereo depth estimation}: wavelength-dependent attenuation, volumetric backscattering, and refractive distortions severely degrade photometric consistency, texture contrast, and color fidelity across stereo views~\cite{akkaynak2018revised,berman2020underwater}.
These degradations directly weaken the correspondence cues that stereo methods rely on, making underwater scenes a particularly demanding setting for evaluating contextual reasoning capability.
Domain-specific methods such as UWNet~\cite{zhu2024uwnet} introduce attention and cross-search modules tailored to underwater imagery but remain tightly coupled to specific degradation patterns.
On the data side, annotated underwater stereo corpora are scarce: UWStereo~\cite{lv2025uwstereo} provides dense synthetic annotations but exhibits a sim-to-real gap, and existing datasets generally lack physically grounded underwater optical modeling paired with dense stereo ground truth.
To address this data bottleneck, we introduce SeaStereo-Dataset, a physically rendered underwater stereo corpus with dense disparity annotations under realistic subsea optical conditions.

In summary, iterative stereo methods suffer from an information gap between powerful backbones and their local update interface, efficient methods sacrifice contextual capacity for speed, and underwater stereo remains limited by data scarcity.
LinStereo addresses all three: it widens the update interface to leverage modern backbones, achieves efficiency through rapid convergence rather than architectural simplification, and contributes a large-scale underwater training and evaluation corpus.

\section{The Proposed Method}
\label{sec:method}

\subsection{Overview}
\label{sec:overview}

\begin{figure}[t]

    \centering
    \includegraphics[width=\textwidth, page=1]{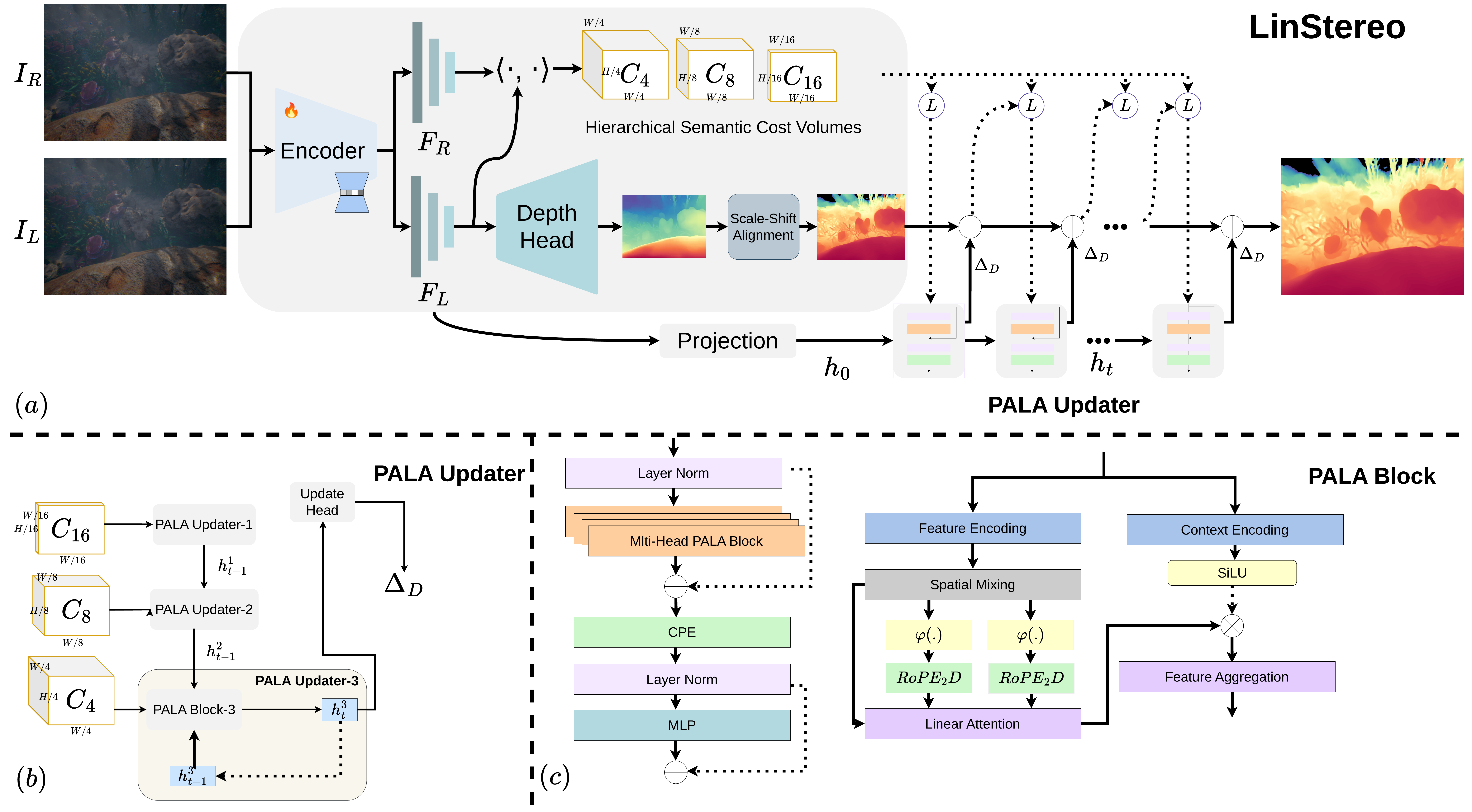}
    \caption{\textbf{Detailed Architecture of LinStereo:} Our model replaces the ConvGRU update operator with the Position-Aware Linear Attention (PALA) updater, which enables global spatial reasoning over cost volume features at linear complexity by attention. The attention path applies kernel-activated queries and keys with 2D rotary positional encoding, while a parallel context modulation branch provides adaptive gating for selective feature aggregation.}
    \label{main arch}
\end{figure}

We propose LinStereo, a stereo depth estimation framework that addresses the growing mismatch between VFM backbone capacity and iterative update utilization in stereo pipelines. Recent methods~\cite{wen2025foundationstereo, jiang2025defom} have demonstrated that VFM backbones significantly improve stereo depth estimation through richer feature representations, yet they inherit the narrow-bandwidth update architecture from pre-VFM designs~\cite{lipson2021raftstereo, xu2023igev}, creating a fundamental mismatch between a high-capacity encoder and a low-capacity iterative decoder. As illustrated in Fig.~\ref{main arch}, LinStereo closes this gap with three components that each target a distinct information pathway: \textbf{Hierarchical Semantic Cost Volumes (HSCV)} exploit the VFM's multi-scale hierarchy to provide semantically matched correlation at every refinement level (\textit{what} to match); \textbf{Depth Prior Initialization} converts the backbone's monocular depth into a geometrically informed warm start (\textit{where} to start); and \textbf{Position-Aware Linear Attention (PALA)} replaces local convolutions with global context aggregation at linear cost (\textit{how} to refine). These components target complementary information pathways and, as shown in Sec.~\ref{sec:experiments}, yield consistent gains across all evaluated settings. The following subsections present them in pipeline order: from feature construction through initialization to iterative refinement.

\subsection{Feature Extraction}
\label{sec:backbone}

Before describing the main architectural components, we introduce the shared feature extraction backbone. {Given a rectified stereo pair $(\mathbf{I}_L, \mathbf{I}_R) \in \mathbb{R}^{H \times W \times 3}$, where $H$ and $W$ denote the image height and width, we} employ Depth Anything V3 (DA3)~\cite{lin2025depthanything3} as a shared backbone for both stereo feature extraction and monocular depth estimation. As the latest Depth Anything release, DA3 offers more detailed and geometrically consistent monocular depth than earlier versions, motivating its use here. It consists of a DINOv2 ViT-B/14 encoder~\cite{oquab2024dinov2} and a Dual-DPT head~\cite{ranftl2021dpt}, which first reassembles multi-scale intermediate ViT representations through shared modules and then routes them to separate fusion branches for depth and ray prediction. We leverage the intermediate outputs of this architecture in two complementary ways from a single forward pass. For stereo correlation, we extract multi-scale feature representations after the shared reassembly layers but before fusion branches. This yields multi-scale features $\{\mathbf{F}_{L,s}, \mathbf{F}_{R,s}\}_{s \in \{4,8,16\}}$ with $\mathbf{F}_{L/R,s} \in \mathbb{R}^{\frac{H}{s} \times \frac{W}{s} \times D_s}$, where $s$ denotes the spatial downsampling factor. These features retain both fine geometric detail and high-level semantic structure from the VFM, and are subsequently projected to a uniform channel dimension {$c_f{=}128$} for HSCV (Sec.~\ref{sec:hscv}). Meanwhile, the depth fusion branch produces a dense monocular depth map that provides the initialization signal for iterative refinement (Sec.~\ref{sec:depth_init}), adding only the cost of the lightweight fusion layers.

\subsection{Hierarchical Semantic Cost Volumes{(HSCV)}}
\label{sec:hscv}

Existing iterative methods~\cite{lipson2021raftstereo, xu2023igev} build a single
correlation volume at a fixed resolution (e.g., $1/4$), restricting matching to a
narrow semantic bandwidth and ignoring the hierarchy of modern VFM backbones. HSCV
instead makes two contributions. First, \textit{utilization}: we build a per-scale
correlation volume from each VFM scale and keep the full set simultaneously
accessible, so PALA can re-query any scale at every iteration, unlike one-shot
multi-scale fusion~\cite{xu2020aanet} or coarse-to-fine cascades~\cite{gu2020cascade}
that consume coarser volumes once and discard them. Second, \textit{representation}:
each per-scale volume carries its own internal disparity-axis pyramid for
coarse-to-fine matching while preserving spatial resolution.

Given the stereo feature maps $\{\mathbf{F}_{L,s}, \mathbf{F}_{R,s}\}$ for each downsampling factor {$s \in \{4, 8, 16\}$} (as defined in Sec.~\ref{sec:backbone}), we first project each through dedicated modules consisting of instance-normalized residual blocks and $1{\times}1$ convolutions to obtain $\hat{\mathbf{F}}_{L/R,s} = \psi_s(\mathbf{F}_{L/R,s})$. We then compute the 1D correlation volume along the epipolar line at each scale:
\begin{equation}
    \mathbf{C}_s(i,j,d) = \frac{1}{\sqrt{C_f}} \left\langle \hat{\mathbf{F}}_{L,s}(i,j), \; \hat{\mathbf{F}}_{R,s}(i, j{-}d) \right\rangle,
    \label{eq:corr_volume}
\end{equation}
where $i, j$ index spatial positions and $d$ indexes disparity displacement. To capture correspondences at both fine-grained and large displacements, each correlation volume $\mathbf{C}_s$ is organized into a $4$-level pyramid by repeatedly downsampling along the disparity dimension, providing progressively larger search ranges while preserving spatial resolution.

This design introduces a two-level matching hierarchy: across semantic scales from the VFM backbone and within each scale's disparity pyramid.
{Each level of the iterative refinement thus receives semantically appropriate correlation signals at its own operating resolution, rather than relying solely on inter-scale hidden state propagation.}

\subsection{Depth Prior Initialization (DPI)}
\label{sec:depth_init}

VFM-based monocular depth estimators capture rich relative structure with strong cross-domain generalization, yet their predictions are affine-invariant and lack the metric scale of a specific stereo rig. While metric monocular models~\cite{hu2024metric3dv2, bochkovskii2025depthpro} exist, their absolute scale is a statistical prior that drifts substantially in out-of-distribution scenes (\eg, underwater). Affine-invariant models avoid this commitment and preserve more robust relative structure across domains; since a calibrated stereo pair already provides geometric correspondences, the metric scale can be recovered directly from the data.
The DA3 depth fusion branch produces an affine-invariant depth map $\hat{\mathbf{D}}_{\text{mono}}$ from the shared forward pass. To convert it into a metrically meaningful disparity initialization, we perform a lightweight scale-shift alignment, replacing the zero-disparity start used in~\cite{lipson2021raftstereo} and its descendants that wastes early iterations on gross layout recovery.

\paragraph{Scale-Shift Alignment.}
Since disparity is inversely proportional to depth ($d \propto 1/Z$), we convert the monocular prediction via $d = \alpha / \hat{\mathbf{D}}_{\text{mono}} + \beta$, where $\alpha$ absorbs the unknown metric scale and $\beta$ accounts for any additive offset.
We estimate $\alpha$ and $\beta$ from sparse correspondences: SIFT keypoints are extracted from the rectified pair and matched along epipolar lines, yielding references $\{(p_k, d_k^{\text{sp}})\}_{k=1}^{K}$. Compared to learning-based matchers~\cite{wen2025foundationstereo,bartolomei2025stereo}, SIFT is sufficient here since only coarse sparse correspondences are needed. The parameters are obtained by least-squares:
\begin{equation}
    (\alpha^*, \beta^*) = \arg\min_{\alpha, \beta} \sum_{k=1}^{K} \left\| d_k^{\text{sp}} - \left(\frac{\alpha}{\hat{\mathbf{D}}_{\text{mono}}(p_k)} + \beta\right) \right\|^2,
    \label{eq:scale_shift}
\end{equation}
giving the initial disparity map:
\begin{equation}
    \mathbf{d}^{(0)} = \frac{\alpha^*}{\hat{\mathbf{D}}_{\text{mono}}} + \beta^*.
    \label{eq:disp_init}
\end{equation}
When inlier matches fall below $K_{\min}{=}20$, we fall back to zero-disparity initialization (Fig.~\ref{main arch}-(a)). We note that $\mathbf{d}^{(0)}$ serves as a coarse warm start; VFM depths do not strictly follow a global affine model, and handcrafted matches introduce localization noise. The iterative refinement with PALA (Sec.~\ref{sec:pala}) is therefore essential to recover fine-grained disparity.

\subsection{Position-Aware Linear Attention {(PALA)} Update}
\label{sec:pala}

Existing recurrent update operators reason over a local spatial neighborhood per iteration, yet effective stereo refinement should propagate context across the entire spatial extent of the feature map, for instance, transferring reliable depth estimates from textured regions to nearby textureless areas. PALA achieves this through linear attention, which aggregates global spatial context with $\mathcal{O}(N)$ complexity per iteration, enabling each update step to attend to the full feature map without the quadratic cost of softmax attention.
LinStereo employs three PALA updaters, one per scale, each containing a single PALA block that processes features at three resolution scales.
\paragraph{Multi-Scale PALA Updater}
As shown in Fig.~\ref{main arch}-(b), at each refinement iteration $t$, given the current disparity estimate $\mathbf{d}_s^{(t)}$ for downsampling factor $s$, we perform a lookup into $\mathbf{C}_s$ around the predicted correspondence across all pyramid levels, and concatenate the retrieved values to form the per-scale correlation feature $\mathbf{c}_s^{(t)}$. A dedicated motion encoder then encodes the matching signal as $\mathbf{m}_s^{(t)} = \mathrm{MotionEnc}_s(\mathbf{d}_s^{(t)}, \mathbf{c}_s^{(t)})$. Unlike prior work~\cite{lipson2021raftstereo} where only the finest scale receives such inputs, our HSCV design supplies each scale with its own matching signal, and hidden states $\{\mathbf{h}_s\}_{s \in \{4,8,16\}}$ are updated in coarse-to-fine order:
\begin{align}
    \mathbf{h}_{16}^{(t+1)} &= \mathrm{PALA}_{16}\!\left(\mathbf{h}_{16}^{(t)},\, \mathbf{m}_{16}^{(t)},\, \mathbf{h}_8^{(t)}\right), \label{eq:gru32} \\
    \mathbf{h}_8^{(t+1)}  &= \mathrm{PALA}_8\!\left(\mathbf{h}_8^{(t)},\, \mathbf{m}_8^{(t)},\, \mathbf{h}_4^{(t)},\, \mathbf{h}_{16}^{(t+1)}\right), \label{eq:gru16} \\
    \mathbf{h}_4^{(t+1)}  &= \mathrm{PALA}_4\!\left(\mathbf{h}_4^{(t)},\, \mathbf{m}_4^{(t)},\, \mathbf{h}_8^{(t+1)}\right), \label{eq:gru08}
\end{align}
where $\mathbf{h}_4^{(t+1)}$ produces the disparity update for final prediction.

\paragraph{PALA Block Architecture}
Standard linear attention achieves $\mathcal{O}(N)$ complexity by exploiting the associativity of matrix multiplication: rather than an $N{\times}N$ attention matrix, one precomputes the key-value product $\mathbf{K}^\top \mathbf{V} \in \mathbb{R}^{C_h \times C_h}$ and shares it across all query positions. However, this very associativity collapses all pairwise spatial relationships into a single global summary, making the resulting attention {position-agnostic}. For stereo depth refinement this is especially problematic: disparity updates are inherently spatially structured. For example, nearby pixels generally share similar depth, and discontinuities align with object boundaries at specific locations. A position-blind mechanism cannot distinguish these cases.

In the proposed PALA block (Fig.~\ref{main arch}-$c$), we restore spatial awareness without sacrificing linear complexity by incorporating a relative position encoding through Rotary Position Embeddings (RoPE)~\cite{su2024roformer}. Given kernel-mapped queries and keys $\hat{\mathbf{Q}} = \phi(\mathbf{W}_Q \mathbf{x})$, $\hat{\mathbf{K}} = \phi(\mathbf{W}_K \mathbf{x})$ with the non-negative kernel $\phi(\mathbf{z})$, we apply 2D RoPE extended to both height and width axes to obtain position-aware variants $\tilde{\mathbf{Q}} = \mathrm{RoPE}(\hat{\mathbf{Q}})$ and $\tilde{\mathbf{K}} = \mathrm{RoPE}(\hat{\mathbf{K}})$. The linear attention output is then:
\begin{equation}
    \mathbf{O}_{\text{attn}} = \tilde{\mathbf{Q}} \!\left(\tilde{\mathbf{K}}^\top \mathbf{V}\right) \!\cdot\! \left(\hat{\mathbf{Q}} \, \overline{\hat{\mathbf{K}}}^\top + \epsilon \right)^{-1},
    \label{eq:linear_attn}
\end{equation}
where $\overline{\hat{\mathbf{K}}} {=} \frac{1}{N}\sum_{j}\hat{\mathbf{K}}_j$. Crucially, RoPE is applied{asymmetrically}: only the numerator uses position-augmented $\tilde{\mathbf{Q}}, \tilde{\mathbf{K}}$ to encode relative spatial relationships, while the denominator retains the original $\hat{\mathbf{Q}}, \hat{\mathbf{K}}$ for stable normalization. Applying RoPE uniformly would make the normalizing factor position-dependent, distorting attention magnitudes and destabilizing training. This preserves $\mathcal{O}(N {\cdot} C_h^2)$ complexity, while restoring the spatial structure that vanilla linear attention discards. Beyond the relative encoding within the attention mechanism, stereo disparity fields are locally smooth: neighboring pixels on the same surface share similar depth. We therefore complement the attention with local spatial encoding on the value branch to reinforce this neighborhood coherence, and absolute position encoding flanking the block to anchor features to their spatial coordinates, which helps the network distinguish systematic depth variation across different image regions.

Since PALA operates within an iterative refinement loop, we employ a gated update to control information flow across iterations:
\begin{equation}
    \mathbf{z}_g = \sigma\!\left(\mathrm{Conv}_{3\times3}([\mathbf{h} \| \mathbf{O}])\right), \quad \mathbf{h}_{\text{new}} = (1{-}\mathbf{z}_g) \odot \mathbf{h} + \mathbf{z}_g \odot \tanh(\mathbf{O}),
    \label{eq:gru_update}
\end{equation}
Here, $\|$ denotes channel-wise concatenation, $\sigma$ the sigmoid, and $\odot$ element-wise multiplication.
In stereo matching, correlation evidence is inherently non-uniform: textured regions yield strong, unambiguous matches, while occluded or textureless areas produce noisy signals. The gate $\mathbf{z}_g$ enables each location to aggressively incorporate new evidence where matching is confident and to preserve the accumulated estimate where it is not, allowing reliable disparity to gradually propagate into uncertain regions across iterations.

\subsection{Loss and Training Strategy}
\label{sec:output}

The finest-scale hidden state $\mathbf{h}_4^{(t+1)}$ is decoded into a disparity residual $\Delta\mathbf{d}^{(t)}$ through stacked depth-wise convolutional blocks. The training objective supervises each iterative refinement step with exponentially increasing weights:
\begin{equation}
    \mathcal{L} = \sum_{t=1}^{T} \gamma^{T-t} \Big( \left\| \mathbf{d}^{(t)} - \mathbf{d}_{\text{gt}} \right\|_1 + \lambda_s \mathcal{L}_{\text{smooth}}(\mathbf{d}^{(t)}) + \lambda_g \mathcal{L}_{\text{grad}}(\mathbf{d}^{(t)}) \Big),
    \label{eq:loss_total}
\end{equation}
where $\gamma$ assigns progressively higher weight to later iterations. $\mathcal{L}_{\text{smooth}}$ is an edge-aware smoothness term that penalizes disparity gradients in regions with low image intensity variation, encouraging piece-wise smooth predictions while preserving discontinuities at object boundaries. $\mathcal{L}_{\text{grad}}$~\cite{ranftl2020towards} computes $\ell_1$ differences between predicted and ground-truth disparity gradients at multiple spatial scales, guiding the network to recover accurate depth edges. Both auxiliary terms carry small weights ($\lambda_s$, $\lambda_g$). Specific values of all training hyperparameters are provided in Sec.~\ref{sec:implementation}.

\section{Experiments}
\label{sec:experiments}

\subsection{Datasets and Metrics}
\label{sec:datasets}

\noindent\textit{Training.}
LinStereo is trained exclusively on \textbf{SceneFlow}~\cite{mayer2016large}, a large-scale synthetic dataset comprising approximately 35{,}000 stereo pairs with dense disparity ground truth. No real-world, underwater, or domain-specific data is used at any stage of training.
\noindent\textit{Standard Benchmarks.}
Cross-domain generalization is evaluated on five standard stereo benchmarks: \textbf{KITTI 2015}~\cite{menze2015object} and \textbf{KITTI 2012}~\cite{geiger2012we}, which provide real-world driving scenes; \textbf{Middlebury (H)}~\cite{scharstein2014high}, an indoor benchmark at half-resolution; \textbf{ETH3D}~\cite{schops2017multi}, a high-precision LiDAR-aligned dataset; and \textbf{Booster (Q)}~\cite{ramirez2022open}, evaluated at quarter resolution and notable for containing transparent and specular surfaces. Following standard protocol, end-point error (EPE) and percentage of pixels with disparity error $> k$ pixels (bad-$k$) are reported.
\noindent\textit{Underwater Benchmarks.}
To assess generalization under photometric degradation, two underwater benchmarks are adopted. \textbf{TartanAir-UW} consists of 13{,}583 stereo pairs from 22 underwater sequences in TartanAir~\cite{wang2020tartanair}, representing simulated subsea scenes with realistic light scattering. \textbf{SQUID}~\cite{berman2020underwater} provides 57 real-world underwater stereo pairs across 4 different scenes, covering a range of water visibility conditions. Both benchmarks operate at moderate-to-large depth scales. Since these benchmarks provide depth rather than disparity ground truth, depth-domain metrics are adopted following prior work~\cite{bartolomei2025stereo}: AbsRel, SqRel, RMSE, LogRMSE, and accuracy thresholds $\delta < 1.25^k$ ($k\!=\!1,2,3$). Additionally, we contribute a synthetic underwater stereo corpus (\textbf{SeaStereo-Dataset}, ${\sim}$40K stereo pairs) with dense disparity ground truth under realistic subsea optical conditions; dataset details and evaluation are provided in the supplementary material. SeaStereo is built by rendering ShapeNetCore~\cite{shapenet2015} foreground objects (arbitrary \texttt{.obj} meshes) over 3D marine backgrounds (coral, fish, and shipwrecks) under varying Jerlov water types.

\noindent\textit{Real-World Evaluation.}
LinStereo is further validated on a controlled \textbf{laboratory tank} dataset collected in-house at close range ($<2$\,m). In contrast to TartanAir-UW and SQUID, which evaluate at moderate-to-large depth scales, this benchmark tests near-range precision under genuine underwater conditions, where small absolute errors translate to large relative depth deviations. Ground-truth depth is obtained via AprilTag (16h5) markers with PnP-based pose recovery; detailed experimental setup is described in the supplementary material. 

\subsection{Implementation Details}
\label{sec:implementation}

The DA3 backbone is kept \textbf{fully frozen}; only the decoder is trained. For DPI, SIFT matching uses a minimum inlier threshold $K_{\min}=20$, below which we fall back to zero-disparity initialization; this occurs for $0\%$ of TartanAir-UW and $3.7\%$ of SQUID frames and costs only $+0.08$\,px EPE. We train with AdamW for 200K steps on SceneFlow (batch size 8, $320\times640$ crops, one NVIDIA H100), with a one-cycle schedule peaking at $2\times10^{-4}$. The loss (Eq.~(\ref{eq:loss_total})) spans $T=8$ refinement iterations ($\gamma=0.9$) with smoothness ($\lambda_s=0.001$) and gradient ($\lambda_g=0.01$) terms. For real-time use, the budget can drop to $T=2$ (supplementary).

\subsection{Main Results}
\label{sec:main_results}
\begin{table*}[t]
\centering
\caption{%
  Cross-domain generalization on standard stereo benchmarks.
  Booster~(Q): quarter-resolution evaluation.
  Extra~Data~($\checkmark$): method uses additional training data beyond SceneFlow~\cite{mayer2016large}. \textbf{Bold}/\underline{underline}: best/second-best among methods without extra data.
}
\label{tab:general_benchmarks}
\resizebox{\textwidth}{!}{%
\begin{tabular}{l c cc cc cccccc cc cc}
\toprule
\multirow{3}{*}{Method}
  & \multirow{3}{*}{\begin{tabular}[c]{@{}c@{}}Extra\\Data\end{tabular}}
  & \multicolumn{2}{c}{KITTI 2015}
  & \multicolumn{2}{c}{KITTI 2012}
  & \multicolumn{6}{c}{Middlebury (H)}
  & \multicolumn{2}{c}{ETH3D}
  & \multicolumn{2}{c}{Booster (Q)} \\
\cmidrule(lr){3-4} \cmidrule(lr){5-6} \cmidrule(lr){7-12} \cmidrule(lr){13-14} \cmidrule(lr){15-16}
  &
  & \multirow{2}{*}{EPE} & \multirow{2}{*}{bad3}
  & \multirow{2}{*}{EPE} & \multirow{2}{*}{bad3}
  & \multicolumn{2}{c}{All}
  & \multicolumn{2}{c}{NonOcc}
  & \multicolumn{2}{c}{Occ}
  & \multirow{2}{*}{EPE} & \multirow{2}{*}{bad1}
  & \multirow{2}{*}{EPE} & \multirow{2}{*}{bad2} \\
\cmidrule(lr){7-8} \cmidrule(lr){9-10} \cmidrule(lr){11-12}
  &  &  &  &  &  & EPE & bad2 & EPE & bad2 & EPE & bad2 &  &  &  &  \\

\midrule
PSMNet~\cite{chang2018pyramid}
  & ---
  & 4.05 & 28.51 & 3.77 & 27.34
  & 11.48 & 36.06 & 10.56 & 32.11 & 17.69 & 62.58
  & 2.15 & 15.13
  & 14.17 & 49.79 \\

RAFT-Stereo~\cite{lipson2021raftstereo}
  & ---
  & 1.13 & 5.69 & 0.90 & 4.35
  & 1.92 & 12.6 & 1.09 & 8.65 & 3.31 & 26.39
  & 0.36 & 3.3
  & 4.18 & 17.64 \\

UniMatch~\cite{xu2023unifying}
  & ---
  & 1.63 & 11.01 & 1.71 & 12.04
  & 4.21 & 34.72 & 3.96 & 31.52 & 5.70 & 56.51
  & 0.69 & 18.61
  & 8.57 & 47.30 \\

MoCha-Stereo~\cite{chen2024mocha}
  & ---
  & 1.29 & 5.97 & 1.02 & 4.83
  & 2.66 & 10.18 & 2.49 & 7.96 & 3.84 & 24.16
  & 0.28 & 3.47
  & 3.88 & 16.82 \\

NMRF~\cite{guan2024nmrf}
  & ---
  & 1.17 & 5.31 & 0.92 & 4.63
  & 2.91 & 13.36 & 2.73 & 10.90 & 4.22 & 29.27
  & 0.31 & 3.8
  & 5.05 & 26.22 \\

MGStereo~\cite{yao2025diving}
  & ---
  & 1.13 & 5.62 & 0.87 & 4.16
  & 1.15 & 8.39 & \underline{0.85} & \underline{5.67} & 2.89 & 26.50
  & 0.25 & 1.88
  & 2.26 & 11.02 \\

Stereo~Anywhere(ViT-L)~\cite{bartolomei2025stereo}
  & ---
  & 1.07 & 4.93 & 0.83 & 3.90
  & 0.94 & 6.96 & \textbf{0.79} & \textbf{4.75} & 2.67 & 20.34
  & \underline{0.24} & 2.66
  & \underline{2.21} & \textbf{9.91} \\

DEFOM-Stereo~(ViT-L)~\cite{jiang2025defom}
  & ---
  & 1.06 & 4.99 & 0.84 & 3.76
  & \underline{0.84} & \textbf{5.91} & 0.90 & 5.76 & \underline{2.11} & \underline{19.34}
  & 0.35 & 2.35
  & 3.52 & 13.20 \\

MonSter~(ViT-L)~\cite{cheng2025monster}
  & ---
  & \textbf{0.89} & \textbf{3.58} & \textbf{0.75} & \underline{3.58}
  & 1.98 & 10.41 & 2.03 & 8.79 & 2.62 & 21.56
  & \textbf{0.23} & \underline{1.41}
  & 3.08 & 17.45 \\

BridgeDepth~(ViT-L)~\cite{guan2025bridgedepth}
  & ---
  & 1.13 & 4.90 & 0.86 & 4.39
  & 2.21 & 10.27 & 2.25 & 8.66 & 2.68 & 20.42
  & \underline{0.24} & \textbf{1.33}
  & 4.71 & 18.79 \\
\midrule
\rowcolor[rgb]{.886,.937,.855}
\textbf{Ours(ViT-B)}
  & ---
  & \underline{1.01} & \underline{4.54} & \underline{0.76} & \textbf{3.49}
  & \textbf{0.83} & \underline{6.01} & 0.89 & \underline{5.67} & \textbf{1.33} & \textbf{8.69}
  & \underline{0.24} & 2.41
  & \textbf{2.14} & \underline{9.93} \\
\midrule
IGEV-Stereo~\cite{xu2023igev}
  & $\checkmark$
  & 1.21 & 6.03 & 1.03 & 5.13
  & 2.63 & 11.93 & 2.27 & 9.49 & 5.02 & 26.04
  & 0.33 & 4.0
  & 4.26 & 17.58 \\

Selective-RAFT~\cite{wang2024selective}
  & $\checkmark$
  & 1.27 & 6.68 & 1.08 & 5.19
  & 2.34 & 12.04 & 2.05 & 9.45 & 4.17 & 27.4
  & 0.34 & 4.36
  & 4.14 & 19.52 \\

Selective-IGEV~\cite{wang2024selective}
  & $\checkmark$
  & 1.25 & 6.06 & 1.08 & 5.64
  & 2.59 & 11.79 & 2.31 & 9.22 & 4.35 & 28.10
  & 0.33 & 4.05
  & 4.62 & 19.28 \\

IGEV++~\cite{xu2025igevpp}
  & $\checkmark$
  & 1.27 & 6.22 & 1.20 & 6.81
  & 3.21 & 10.85 & 2.41 & 7.86 & 6.83 & 28.68
  & 0.35 & 4.60
  & 5.00 & 18.62 \\

FoundationStereo~(ViT-L)~\cite{wen2025foundationstereo}
  & $\checkmark$
  & 0.95 & 3.65 & 0.71 & 3.18
  & 0.71 & 3.47 & 0.42 & 1.46 & 2.67 & 16.44
  & 0.21 & 1.14
  & 1.77 & 6.89 \\
\bottomrule
\end{tabular}}
\end{table*}

\hspace*{\parindent}\textit{Standard-Benchmark Generalization.}
Tab.~\ref{tab:general_benchmarks} reports \emph{zero-shot} cross-domain results. Each method uses its official released weights (trained on SceneFlow unless marked $\checkmark$) and is evaluated on the training splits with public ground truth (Middlebury at half resolution~(H)). FoundationStereo~\cite{wen2025foundationstereo} (ViT-L, with substantial extra training data) is the strongest baseline; its larger backbone and extra data account for its lead on these standard benchmarks. We deliberately keep LinStereo a ViT-B\footnote{ViT-B keeps training feasible on modest hardware; our design is backbone-agnostic.}, SceneFlow-only model to isolate the decoder's contribution. Even in this smaller-backbone, same-data setting, it remains highly competitive: it matches or beats the ViT-L BridgeDepth~\cite{guan2025bridgedepth} on 13/14 metrics (trailing only on ETH3D bad-1) and outperforms MonSter~\cite{cheng2025monster} on 9/14. The gains thus stem from the proposed architectural components rather than backbone capacity or extra data.
The most pronounced improvement appears on occluded regions of Middlebury~(H), where EPE is $37\%$ lower than the previous best~\cite{jiang2025defom}. This validates PALA's global context propagation: reliable disparity estimates from well-matched visible areas are propagated to occluded pixels that lack direct photometric evidence, which local-window updaters struggle to achieve.
Booster~(Q) is the most challenging benchmark, with transparent and specular surfaces that violate photometric consistency; even here, LinStereo remains stable, trailing the best by only a marginal Bad-2 gap (9.93 vs.\ 9.91~\cite{bartolomei2025stereo}) that lies within statistical variation.

\begin{table*}[t]
\centering
\caption{
  Quantitative comparison on underwater depth estimation.
  TartanAir-UW: underwater images from TartanAir~\cite{wang2020tartanair}.
  SQUID: real-world underwater dataset~\cite{berman2020underwater}.
  $\checkmark$: uses extra training data beyond SceneFlow~\cite{mayer2016large}.
}
\label{tab:quant_underwater}
\resizebox{\textwidth}{!}{%
\begin{tabular}{l c ccccccc ccccccc}
\toprule
\multirow{2}{*}{Method}
  & \multirow{2}{*}{\begin{tabular}[c]{@{}c@{}}Extra\\Data\end{tabular}}
  & \multicolumn{7}{c}{TartanAir UW}
  & \multicolumn{7}{c}{SQUID} \\
\cmidrule(lr){3-9} \cmidrule(lr){10-16}
  &
  & AbsRel$\downarrow$ & SqRel$\downarrow$ & RMSE$\downarrow$ & LogRMSE$\downarrow$
  & A1$\uparrow$ & A2$\uparrow$ & A3$\uparrow$
  & AbsRel$\downarrow$ & SqRel$\downarrow$ & RMSE$\downarrow$ & LogRMSE$\downarrow$
  & A1$\uparrow$ & A2$\uparrow$ & A3$\uparrow$ \\
\midrule
UniMatch~\cite{xu2023unifying}
  & ---
  & 0.17 & 2.59 & 6.35 & 0.26 & 0.759 & 0.830 & 0.837
  & 1.38 & 20.91 & 6.42 & 0.66 & 0.561 & 0.660 & 0.734 \\
MoCha-Stereo~\cite{chen2024mocha}
  & ---
  & 0.15 & 2.30 & 6.02 & 0.25 & 0.769 & 0.835 & 0.843
  & 0.15 & 0.61 & 1.55 & 0.20 & 0.875 & 0.931 & 0.958 \\
NMRF~\cite{guan2024nmrf}
  & ---
  & 0.11 & 1.08 & 4.63 & 0.21 & 0.811 & 0.856 & 0.870
  & 0.43 & 5.27 & 2.57 & 0.33 & 0.844 & 0.904 & 0.935 \\
PSMNet~\cite{chang2018pyramid}
  & ---
  & 0.10 & 1.02 & 4.56 & 0.21 & 0.813 & 0.857 & 0.871
  & 0.46 & 4.31 & 3.43 & 0.51 & 0.736 & 0.813 & 0.853 \\
RAFT-Stereo~\cite{lipson2021raftstereo}
  & ---
  & 0.08 & 0.65 & 4.36 & 0.20 & 0.902 & 0.963 & 0.984
  & 0.07 & 0.27 & 1.25 & 0.12 & 0.937 & 0.971 & 0.987 \\
MGStereo~\cite{yao2025diving}
  & ---
  & 0.08 & 0.55 & 3.69 & 0.19 & 0.911 & 0.968 & 0.987
  & 0.09 & 0.71 & 1.99 & 0.16 & 0.925 & 0.958 & 0.976 \\
Stereo~Anywhere~\cite{bartolomei2025stereo}
  & ---
  & 0.06 & 0.41 & 3.24 & 0.18 & 0.946 & 0.980 & 0.989
  & 0.07 & 0.40 & 1.46 & 0.13 & 0.937 & 0.976 & 0.985 \\
DEFOM-Stereo~\cite{jiang2025defom}
  & ---
  & 0.05 & 0.41 & 3.13 & 0.17 & 0.953 & 0.982 & 0.990
  & 0.09 & 0.63 & 2.00 & 0.16 & 0.915 & 0.954 & 0.977 \\
\midrule
\rowcolor[rgb]{.886,.937,.855}
\textbf{Ours}
  & ---
  & \textbf{0.04} & \textbf{0.19} & \textbf{2.08} & \textbf{0.09}
  & \textbf{0.959} & \textbf{0.992} & \textbf{0.998}
  & \textbf{0.04} & \textbf{0.12} & \textbf{0.90} & \textbf{0.08}
  & \textbf{0.970} & \textbf{0.990} & \textbf{0.996} \\
\midrule
Selective-Stereo~(RAFT)~\cite{wang2024selective}
  & $\checkmark$
  & 0.09 & 0.67 & 4.31 & 0.21 & 0.902 & 0.962 & 0.981
  & 0.11 & 0.22 & 1.16 & 0.16 & 0.877 & 0.940 & 0.968 \\
Selective-Stereo~(IGEV)~\cite{wang2024selective}
  & $\checkmark$
  & 0.11 & 1.02 & 5.01 & 0.23 & 0.856 & 0.942 & 0.976
  & 0.08 & 0.21 & 1.05 & 0.14 & 0.932 & 0.966 & 0.980 \\
IGEV-Stereo~\cite{xu2023igev}
  & $\checkmark$
  & 0.10 & 0.90 & 4.68 & 0.21 & 0.891 & 0.955 & 0.979
  & 0.20 & 1.35 & 2.68 & 0.46 & 0.760 & 0.821 & 0.863 \\
IGEV++~\cite{xu2025igevpp}
  & $\checkmark$
  & 0.09 & 0.81 & 4.37 & 0.20 & 0.905 & 0.962 & 0.983
  & 0.06 & 0.22 & 1.11 & 0.12 & 0.950 & 0.981 & 0.990 \\
FoundationStereo~\cite{wen2025foundationstereo}
  & $\checkmark$
  & 0.05 & 0.40 & 3.01 & 0.16 & \textbf{0.959} & 0.984 & 0.991
  & 0.07 & 0.30 & 1.36 & 0.13 & 0.940 & 0.976 & 0.987 \\
\bottomrule
\end{tabular}}
\end{table*}

\textit{Underwater Generalization.}
Underwater scenes are a stringent cross-domain test, with photometric degradation from scattering and attenuation. On TartanAir-UW and SQUID (Tab.~\ref{tab:quant_underwater}, $T\!=\!8$), LinStereo, trained only on SceneFlow, achieves the best results on both, surpassing even methods trained on real-world or domain-specific data~\cite{wang2024selective,xu2023igev,xu2025igevpp,wen2025foundationstereo}. The largest gains track each benchmark's dominant degradation: $31\%$ lower RMSE than FoundationStereo~\cite{wen2025foundationstereo} on TartanAir-UW (long-range backscatter) and $26\%$ lower AbsRel than IGEV++~\cite{xu2025igevpp} on SQUID (color attenuation), indicating PALA propagates reliable estimates to the most degraded regions.

\begin{figure*}[t]
\centering
\begin{subfigure}[t]{\textwidth}
  \centering
  \includegraphics[width=\columnwidth]{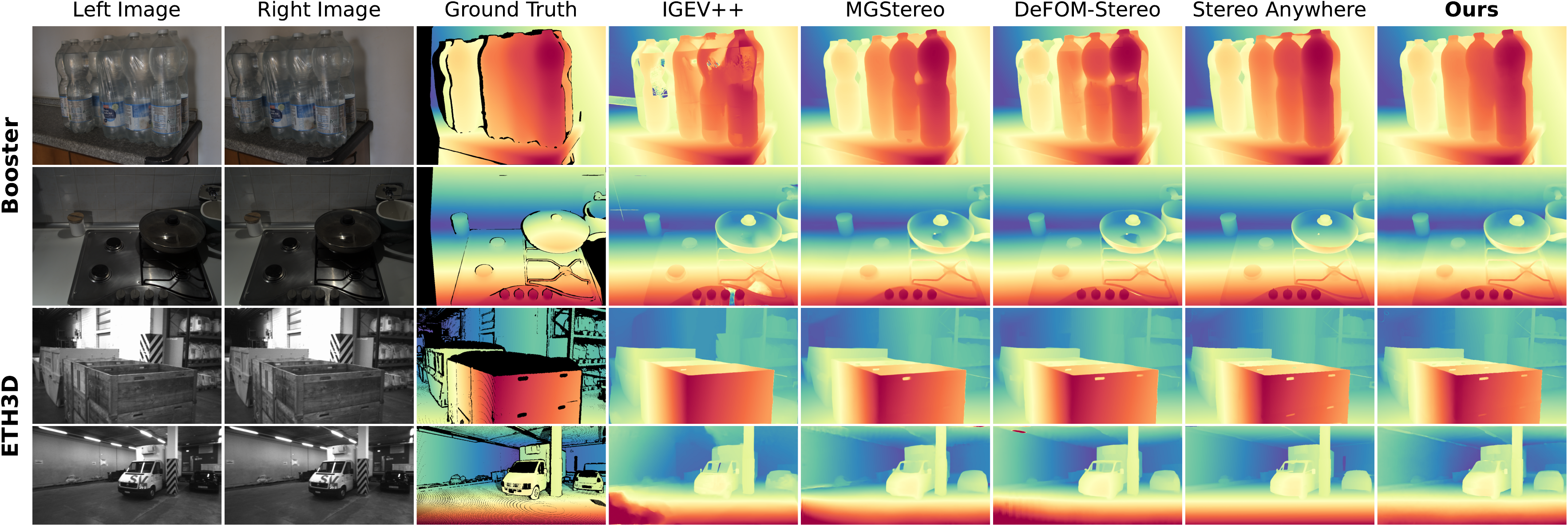}
  \caption{%
    \textbf{Booster(Q) and ETH3D benchmarks.}
    Black regions in ground truth denote unavailable disparity.
  }
  \label{fig:qualitative_standard}
\end{subfigure}

\begin{subfigure}[t]{\textwidth}
  \centering
  \includegraphics[width=\columnwidth]{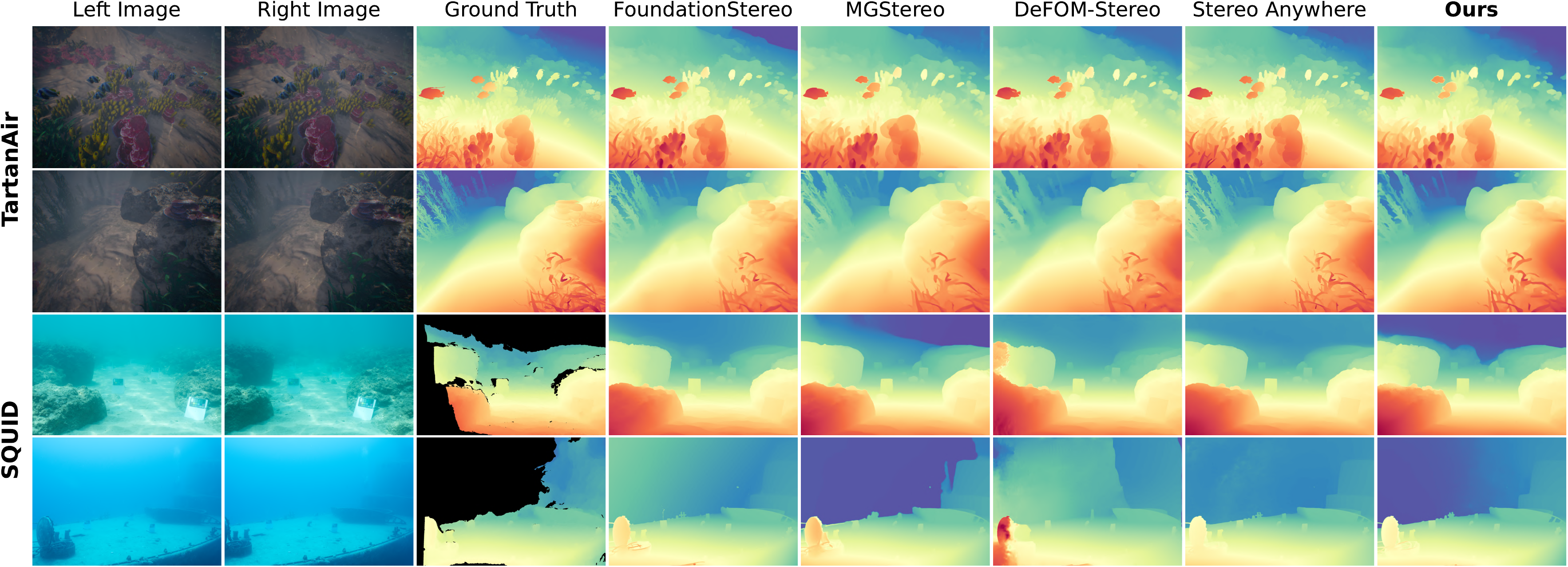}
  \caption{%
    \textbf{Underwater benchmarks.}
    Black regions in SQUID ground truth denote unavailable depth.
  }
  \label{fig:qualitative_underwater}
\end{subfigure}
\caption{\textbf{Qualitative comparison} on standard and underwater stereo benchmarks.}
\label{fig:qualitative_all}
\end{figure*}

\textit{Qualitative Evaluation.}
Fig.~\ref{fig:qualitative_standard} compares predictions on Booster~(Q) and ETH3D. Recent SOTA methods produce visually near-identical depth maps on these well-conditioned scenes; LinStereo shows sharper thin structures and handles non-Lambertian surfaces (\eg, transparent bottles and car windows), consistent with its quantitative lead. This visual near-parity motivates our underwater evaluation, where photometric degradation amplifies inter-method differences. As shown in Fig.~\ref{fig:qualitative_underwater}, competing methods tend to overestimate depth in distant regions and produce inconsistent near-field estimates, whereas LinStereo preserves accurate depth scale at both ranges.

\subsection{Real-World Evaluation}
\label{sec:realworld}
LinStereo is additionally evaluated on a controlled laboratory water tank at close range ($<2$\,m) with AprilTag-based ground truth (Sec.~\ref{sec:datasets}). As shown in Tab.~\ref{tab:quant_labtank}, LinStereo outperforms all compared methods by a substantial margin, including those leveraging multiple additional training datasets. We note that the reference depth comes from a CAD model of the rig (frame and ropes), localized in each view via AprilTags; only these valid-GT pixels are scored, so the larger background errors produced by some baselines fall outside this quantitative evaluation. The result confirms that the proposed architecture generalizes robustly even under extreme near-range, real-world underwater conditions. Full-precision table and qualitative results are provided in supplementary materials. 

\begin{table}[t]
\centering
\caption{
  Quantitative comparison on close-range laboratory tank images ($<2$\,m depth).
  Extra Data ($\checkmark$) indicates training data beyond SceneFlow~\cite{mayer2016large}.
  Accuracy thresholds rounded to two decimal places; full-precision results in supplimentary materials. 
  Best results in \textbf{bold}.
}
\label{tab:quant_labtank}
\resizebox{0.64\columnwidth}{!}{%
\begin{tabular}{l c ccccccc}
\toprule
\multirow{2}{*}{Method}
  & \multirow{2}{*}{\begin{tabular}[c]{@{}c@{}}Extra\\Data\end{tabular}}
  & \multicolumn{7}{c}{Laboratory Tank} \\
\cmidrule(lr){3-9}
  &
  & AbsRel$\downarrow$ & SqRel$\downarrow$ & RMSE$\downarrow$ & LogRMSE$\downarrow$
  & A1$\uparrow$ & A2$\uparrow$ & A3$\uparrow$ \\
\midrule
PSMNet~\cite{chang2018pyramid}
  & ---
  & 0.18 & 0.16 & 0.34 & 0.28 & 0.87 & 0.90 & 0.93 \\
UniMatch~\cite{xu2023unifying}
  & ---
  & 0.08 & 0.04 & 0.17 & 0.16 & 0.93 & 0.96 & 0.98 \\
Stereo~Anywhere~\cite{bartolomei2025stereo}
  & ---
  & 0.09 & 0.06 & 0.20 & 0.19 & 0.92 & 0.93 & 0.97 \\
RAFT-Stereo~\cite{lipson2021raftstereo}
  & ---
  & 0.06 & 0.03 & 0.15 & 0.14 & 0.94 & 0.96 & 0.99 \\
MGStereo~\cite{yao2025diving}
  & ---
  & 0.08 & 0.05 & 0.18 & 0.17 & 0.93 & 0.94 & 0.98 \\
\midrule
\rowcolor[rgb]{.886,.937,.855}
\textbf{Ours} ($T\!=\!8$)
  & ---
  & \textbf{0.04} & \textbf{0.01} & \textbf{0.07} & \textbf{0.07} & \textbf{0.98} & \textbf{0.99} & \textbf{1.00} \\
\midrule
Selective-Stereo~(RAFT)~\cite{wang2024selective}
  & $\checkmark$
  & 0.07 & 0.03 & 0.14 & 0.14 & 0.94 & 0.97 & 0.99 \\
Selective-Stereo~(IGEV)~\cite{wang2024selective}
  & $\checkmark$
  & 0.08 & 0.04 & 0.16 & 0.16 & 0.91 & 0.95 & 0.99 \\
IGEV-Stereo~\cite{xu2023igev}
  & $\checkmark$
  & 0.06 & 0.02 & 0.13 & 0.12 & 0.94 & 0.97 & 0.99 \\
IGEV++~\cite{xu2025igevpp}
  & $\checkmark$
  & 0.05 & 0.02 & 0.12 & 0.12 & 0.96 & 0.97 & 0.99 \\
FoundationStereo~\cite{wen2025foundationstereo}
  & $\checkmark$
  & 0.07 & 0.04 & 0.18 & 0.17 & 0.93 & 0.94 & 0.98 \\
\bottomrule
\end{tabular}}
\end{table}

\begin{table}[t]
\centering
\caption{%
  PALA design ablation: progressive addition of components to global linear attention (A).
  (B) adds global spatial encoding; (C) further adds local spatial encoding; (D) adds adaptive gating to obtain Full PALA.
  Evaluated on KITTI~2015 and TartanAir-UW ($T\!=\!8$).
  Best in \textbf{bold}.
}
\label{tab:ablation_pala_design}
\resizebox{0.7\columnwidth}{!}{%
\begin{tabular}{cl cc cc}
\toprule
 & \multirow{2}{*}{Experiment}
  & \multicolumn{2}{c}{KITTI 2015}
  & \multicolumn{2}{c}{TartanAir-UW} \\
\cmidrule(lr){3-4} \cmidrule(lr){5-6}
  &   & EPE$\downarrow$ & bad3$\downarrow$ & AbsRel$\downarrow$ & RMSE$\downarrow$ \\
\midrule
(A)  & Global linear attention                             & 1.18 & 5.32 & 0.052 & 2.55 \\
(B)  & (A) + Global spatial encoding                      & 1.12 & 5.05 & 0.047 & 2.38 \\
(C)  & (B) + Local spatial encoding                       & 1.06 & 4.76 & 0.043 & 2.22 \\
\rowcolor[rgb]{.886,.937,.855}
(D)  & (C) + Adaptive gating \textbf{[Full PALA]}         & \textbf{1.01} & \textbf{4.54} & \textbf{0.04} & \textbf{2.08} \\
 & \quad with symmetric RoPE & 1.05 & 4.76 & 0.042 & 2.18 \\
\bottomrule
\end{tabular}}
\end{table}

\begin{table*}[t]
\centering
\caption{%
  \textbf{Ablation studies.}
  All models are trained on SceneFlow only and evaluated on KITTI~2015 and TartanAir-UW.
  Best results in \textbf{bold}.
  In~(a), \checkmark{} = enabled; ``HSCV (pooled)'' uses a single pooled multi-scale feature.
}
\label{tab:ablation_factorial}
\label{tab:ablation_train}
\begin{minipage}[t]{0.52\textwidth}
\centering
{\small\textbf{(a) Component ablation}}

\scalebox{0.68}{%
\setlength{\tabcolsep}{4pt}
\begin{tabular}{ccc cc cc}
\toprule
HSCV & DPI & PALA
  & \multicolumn{2}{c}{KITTI 2015}
  & \multicolumn{2}{c}{TartanAir-UW} \\
\cmidrule(lr){4-5} \cmidrule(lr){6-7}
  &   &   & EPE$\downarrow$ & bad3$\downarrow$ & AbsRel$\downarrow$ & RMSE$\downarrow$ \\
\midrule
           &            &            & 1.42 & 6.58 & 0.068 & 3.21 \\
\checkmark &            &            & 1.33 & 6.14 & 0.061 & 2.91 \\
           & \checkmark &            & 1.32 & 6.02 & 0.059 & 2.85 \\
           &            & \checkmark & 1.26 & 5.78 & 0.055 & 2.64 \\
\checkmark & \checkmark &            & 1.21 & 5.51 & 0.052 & 2.52 \\
\checkmark &            & \checkmark & 1.10 & 4.95 & 0.045 & 2.28 \\
           & \checkmark & \checkmark & 1.13 & 5.12 & 0.047 & 2.35 \\
\rowcolor[rgb]{.886,.937,.855}
\checkmark & \checkmark & \checkmark & \textbf{1.01} & \textbf{4.54} & \textbf{0.04} & \textbf{2.08} \\
\multicolumn{3}{l}{HSCV (pooled) + DPI + PALA}
                                     & 1.07 & 4.82 & 0.043 & 2.21 \\
\bottomrule
\end{tabular}}
\end{minipage}%
\hspace{1em}%
\begin{minipage}[t]{0.45\textwidth}
\centering
{\small\textbf{(b) Training hyperparameters}}

\scalebox{0.68}{%
\setlength{\tabcolsep}{4pt}
\begin{tabular}{ccc cc cc}
\toprule
\multirow{2}{*}{LR} & \multirow{2}{*}{BS} & \multirow{2}{*}{$T$}
  & \multicolumn{2}{c}{KITTI 2015}
  & \multicolumn{2}{c}{TartanAir-UW} \\
\cmidrule(lr){4-5} \cmidrule(lr){6-7}
  &   &   & EPE$\downarrow$ & bad3$\downarrow$ & AbsRel$\downarrow$ & RMSE$\downarrow$ \\
\midrule
$1\!\times\!10^{-4}$ & 8 & 8  & 1.09 & 4.88 & 0.045 & 2.27 \\
\rowcolor[rgb]{.886,.937,.855}
\boldmath$2\!\times\!10^{-4}$ & \textbf{8} & \textbf{8} & \textbf{1.01} & \textbf{4.54} & \textbf{0.04} & \textbf{2.08} \\
$4\!\times\!10^{-4}$ & 8 & 8  & 1.16 & 5.22 & 0.050 & 2.45 \\
$2\!\times\!10^{-4}$ & 4  & 8 & 1.07 & 4.79 & 0.044 & 2.22 \\
$2\!\times\!10^{-4}$ & 16 & 8 & 1.05 & 4.68 & 0.042 & 2.16 \\
$2\!\times\!10^{-4}$ & 8 & 2  & 1.24 & 5.65 & 0.055 & 2.62 \\
$2\!\times\!10^{-4}$ & 8 & 4  & 1.10 & 4.93 & 0.046 & 2.30 \\
$2\!\times\!10^{-4}$ & 8 & 12 & 1.03 & 4.61 & 0.041 & 2.12 \\
$2\!\times\!10^{-4}$ & 8 & 16 & 1.02 & 4.58 & 0.041 & 2.10 \\
\bottomrule
\end{tabular}}
\end{minipage}
\end{table*}

\subsection{Ablation Study}
\label{sec:ablation}

All ablations are evaluated on KITTI~2015 and TartanAir-UW to capture the model's behavior on different domains.
\noindent\textit{PALA Design Ablation.}
Tab.~\ref{tab:ablation_pala_design} traces a progressive build-up of the PALA block from a position-agnostic linear attention baseline~(A).
Each addition yields consistent gains on both benchmarks, with global spatial encoding (A$\to$B) providing the largest single improvement, confirming that restoring positional structure is the most critical enhancement to vanilla linear attention. Within this positional encoding, an asymmetric RoPE further outperforms its symmetric counterpart (Tab.~\ref{tab:ablation_pala_design}). Local spatial encoding and adaptive gating contribute complementary refinements, and the full PALA configuration~(D) achieves the best overall performance.

\noindent\textit{Component Ablation.}
Tab.~\ref{tab:ablation_factorial} reports the results for all combinations of the three contributions.
Each component individually improves over the baseline, with PALA contributing the largest single gain, consistent with its role as the primary architectural change.
Notably, the full combination yields super-additive improvements. This is because PALA's global reasoning benefits from the richer per-scale correlations that HSCV provides, while DPI's warm start allows convergence within fewer iterations.
Among pairwise combinations, HSCV\,+\,PALA achieves the second best performance (EPE $1.10$), suggesting that global attention benefits most from multi-scale matching signals.
Replacing the hierarchical design with a single pooled multi-scale feature (HSCV pooled) degrades accuracy, confirming that scale-aligned correlation better exploits VFM representations.

\noindent\textit{Training Hyperparameters.} Tab.~\ref{tab:ablation_train}(b) sweeps peak learning rate, batch size, and refinement iterations.
Performance is relatively robust across the tested ranges, with the optimal setting (LR $= 2\!\times\!10^{-4}$, BS $= 8$, $T = 8$) achieving the best accuracy. The most notable finding is that $T\!=\!12$ yields no improvement over $T\!=\!8$, indicating diminishing returns beyond 8 iterations.

\noindent\textit{Inference Efficiency.}
LinStereo's iterative design lets the refinement budget shrink at test time: at $T\!=\!2$ it runs at $80$\,ms per frame ($12.5$\,FPS at $480\!\times\!640$ on a single RTX~4500) while still surpassing every efficient stereo baseline on both underwater benchmarks, most strikingly reaching AbsRel $0.05$ on SQUID where the strongest competitor stays at $\geq\!0.71$ (Tab.~\ref{tab:realtime_efficiency}; the full efficient-method comparison is given in the supplementary material).
Of the $127$\,M parameters, ${\sim}120$\,M ($94\%$) lie in the frozen DA3 backbone, while the proposed PALA, HSCV, and DPI together add only ${\sim}7$\,M; per refinement step PALA's global update is no costlier than a local ConvGRU operator (Tab.~\ref{tab:per_iter_latency}), so linear attention supplies global reasoning without the quadratic cost of softmax attention.
\begin{table}[t]
\centering
\caption{
Computational efficiency comparison.
Measured at $480 \times 640$ resolution on a single NVIDIA RTX~4500 GPU (24\,GB).
}
\label{tab:realtime_efficiency}
\resizebox{0.6\columnwidth}{!}{%
\begin{tabular}{l c c c c}
\toprule
Method
& Params (M)$\downarrow$
& GFLOPs$\downarrow$
& Time (ms)$\downarrow$
& FPS$\uparrow$ \\
\midrule
LightStereo-S~\cite{guo2025lightstereo}            & 3.44          & \textbf{45.4}  & \textbf{9.9}  & \textbf{101.0} \\
CoEx~\cite{bangunharcana2021coex}                   & 2.73          & 70.6           & 11.0          & 91.0  \\
CGI-Stereo~\cite{xu2023cgistereo}                   & 3.50          & 81.6           & 12.8          & 78.1  \\
Fast-ACVNet~\cite{xu2024acvnet}                     & 3.08          & 104.6          & 13.2          & 76.0  \\
ADStereo~\cite{wang2025adstereo}                    & 7.10          & 201.6          & 15.4          & 64.7  \\
RAFT-Stereo$^\dagger$~\cite{lipson2021raftstereo}   & 9.87          & 296.4          & 18.0          & 55.7  \\
RT-IGEV~\cite{xu2025igevpp}                         & 4.17          & 354.2          & 24.7          & 40.5  \\
MobileStereoNet~\cite{shamsafar2022mobilestereonet} & \textbf{2.35} & 175.6          & 37.7          & 26.5  \\
AANet~\cite{xu2020aanet}                            & 3.93          & --             & 65.8          & 15.2  \\
\midrule
\rowcolor[rgb]{.886, .937, .855}
\textbf{Ours}                                        & 127.0         & 770.4          & 80         & 12.5  \\
\bottomrule
\end{tabular}}
\end{table}

\begin{table}[t]
\centering
\caption{
Per-iteration update-operator latency. Measured at $480 \times 640$ resolution on a single NVIDIA RTX~4500 GPU (24\,GB); $\pm$ denotes the 95\% confidence interval.
}
\label{tab:per_iter_latency}
\resizebox{0.5\columnwidth}{!}{%
\begin{tabular}{l c}
\toprule
Update operator & Latency (ms)$\downarrow$ \\
\midrule
\rowcolor[rgb]{.886, .937, .855}
\textbf{PALA (Ours)} & $3.50 \pm 0.05$ \\
RAFT-Stereo ConvGRU~\cite{lipson2021raftstereo} & $3.63 \pm 0.06$ \\
IGEV ConvGRU~\cite{xu2023igev} & $3.43 \pm 0.03$ \\
\bottomrule
\end{tabular}}
\end{table}

\vspace{-0.3cm}
\section{Conclusions, Limitations and Future Work}
We proposed LinStereo, an iterative stereo depth estimation framework that addresses the information flow between VFM backbones and the iterative update through three complementary mechanisms: Position-Aware Linear Attention for global context aggregation at linear complexity, Hierarchical Semantic Cost Volumes for scale-aligned semantic matching, and Depth Prior Initialization for geometrically informed warm start. Trained exclusively on SceneFlow, LinStereo achieves state-of-the-art-level accuracy on standard stereo benchmarks and demonstrates strong cross-domain generalization to underwater environments, achieving the best overall accuracy on both simulated and real-world underwater benchmarks.

LinStereo builds upon a frozen DA3 backbone, which accounts for most of the inference cost and can limit deployment on resource-constrained platforms at the full iteration budget. Distilling the backbone into a lightweight architecture could reduce latency while preserving cross-domain feature quality. In addition, the DPI module currently relies on handcrafted SIFT features for scale--shift alignment. Replacing this component with a lightweight learned sparse matcher may provide more robust initialization and improved adaptability across domains in the feature.

\vspace{-0.2cm}
\section*{Acknowledgements}
We thank Jay Zhang, Dr.~Gideon Billings for collecting the underwater dataset. This research was funded by the Australian Government through the Australian Research Council (ARC) Research Hub for Intelligent Robotic Systems for Real-Time Asset Management (IH210100030).

\bibliographystyle{splncs04}
\bibliography{review_version/main}

\clearpage

\appendix
\renewcommand{\theHsection}{appendix.\arabic{section}}
\renewcommand{\theHsubsection}{appendix.\arabic{section}.\arabic{subsection}}

\section*{Appendix}

\section{Additional Ablation Studies}
\label{sec:appendix_ablation}

We supplement the ablation studies in the main text with additional hyperparameter sweeps covering HSCV structure (Tab.~\ref{tab:ablation_arch}) and PALA block design (Tab.~\ref{tab:ablation_pala_hyperparam}).

Tab.~\ref{tab:ablation_arch} varies the correlation pyramid depth ($L_\text{corr}$) and the number of active update scales ($n_\text{scales}$).
Increasing $L_\text{corr}$ from 2 to 4 yields monotonic gains with diminishing returns, as each additional pyramid level extends the disparity search range at progressively smaller marginal benefit. The number of update scales has a larger effect: reducing $n_\text{scales}$ from 3 to 1 degrades EPE by 12\%, with underwater metrics disproportionately affected (AbsRel +20\%, RMSE +13\%), since coarse-to-fine propagation is particularly important for resolving near-range structure under photometric degradation.

\begin{table}[h]
\centering
\caption{%
  HSCV hyperparameters.
  Evaluated on KITTI~2015 and TartanAir-UW ($T\!=\!8$).
  Best in \textbf{bold}.
}
\label{tab:ablation_arch}
\scalebox{0.78}{%
\setlength{\tabcolsep}{5pt}
\begin{tabular}{ll cc cc}
\toprule
\multirow{2}{*}{Sweep} & \multirow{2}{*}{Value}
  & \multicolumn{2}{c}{KITTI 2015}
  & \multicolumn{2}{c}{TartanAir-UW} \\
\cmidrule(lr){3-4} \cmidrule(lr){5-6}
  &  & EPE$\downarrow$ & bad3$\downarrow$ & AbsRel$\downarrow$ & RMSE$\downarrow$ \\
\midrule
\multirow{3}{*}{$L_\text{corr}$}
  & 2                          & 1.09 & 4.88 & 0.045 & 2.27 \\
  & 3                          & 1.04 & 4.69 & 0.042 & 2.16 \\
  & \cellcolor[rgb]{.886,.937,.855}\textbf{4 (ours)} & \cellcolor[rgb]{.886,.937,.855}\textbf{1.01} & \cellcolor[rgb]{.886,.937,.855}\textbf{4.54} & \cellcolor[rgb]{.886,.937,.855}\textbf{0.04} & \cellcolor[rgb]{.886,.937,.855}\textbf{2.08} \\
\midrule
\multirow{3}{*}{$n_\text{scales}$}
  & 1                          & 1.13 & 5.14 & 0.048 & 2.36 \\
  & 2                          & 1.06 & 4.74 & 0.043 & 2.20 \\
  & \cellcolor[rgb]{.886,.937,.855}\textbf{3 (ours)} & \cellcolor[rgb]{.886,.937,.855}\textbf{1.01} & \cellcolor[rgb]{.886,.937,.855}\textbf{4.54} & \cellcolor[rgb]{.886,.937,.855}\textbf{0.04} & \cellcolor[rgb]{.886,.937,.855}\textbf{2.08} \\
\bottomrule
\end{tabular}}
\end{table}

Tab.~\ref{tab:ablation_pala_hyperparam} examines three design choices within the PALA block: stacked depth ($N_\text{block}$), attention heads, and hidden dimension.
Stacking multiple blocks per updater increases parameters without improving accuracy, as the outer iterative loop ($T$ passes) already provides sufficient representational depth; a single block per scale is optimal.
The number of heads exhibits an inverted-U pattern: too few heads limit the diversity of attended spatial patterns, while too many reduce the per-head dimensionality below a useful threshold ($128/8 = 16$); four heads provides the best trade-off.
The hidden dimension is most effective when matched to the HSCV projection width ($c_f\!=\!128$): smaller values create an information bottleneck that amplifies outlier errors, while larger values add capacity without commensurate signal from the correlation features.

\begin{table}[h]
\centering
\caption{%
  PALA block design hyperparameters.
  Evaluated on KITTI~2015 and TartanAir-UW ($T\!=\!8$).
  Best in \textbf{bold}.
}
\label{tab:ablation_pala_hyperparam}
\scalebox{0.78}{%
\setlength{\tabcolsep}{5pt}
\begin{tabular}{ll cc cc}
\toprule
\multirow{2}{*}{Sweep} & \multirow{2}{*}{Value}
  & \multicolumn{2}{c}{KITTI 2015}
  & \multicolumn{2}{c}{TartanAir-UW} \\
\cmidrule(lr){3-4} \cmidrule(lr){5-6}
  &  & EPE$\downarrow$ & bad3$\downarrow$ & AbsRel$\downarrow$ & RMSE$\downarrow$ \\
\midrule
\multirow{3}{*}{$N_\text{block}$}
  & \cellcolor[rgb]{.886,.937,.855}\textbf{1 (ours)} & \cellcolor[rgb]{.886,.937,.855}\textbf{1.01} & \cellcolor[rgb]{.886,.937,.855}\textbf{4.54} & \cellcolor[rgb]{.886,.937,.855}\textbf{0.04} & \cellcolor[rgb]{.886,.937,.855}\textbf{2.08} \\
  & 2                  & 1.03          & 4.65          & 0.041         & 2.12 \\
  & 3                  & 1.05          & 4.70          & 0.042         & 2.19 \\
\midrule
\multirow{4}{*}{Heads}
  & 1                  & 1.09          & 4.94          & 0.045         & 2.29 \\
  & 2                  & 1.05          & 4.73          & 0.043         & 2.17 \\
  & \cellcolor[rgb]{.886,.937,.855}\textbf{4 (ours)} & \cellcolor[rgb]{.886,.937,.855}\textbf{1.01} & \cellcolor[rgb]{.886,.937,.855}\textbf{4.54} & \cellcolor[rgb]{.886,.937,.855}\textbf{0.04} & \cellcolor[rgb]{.886,.937,.855}\textbf{2.08} \\
  & 8                  & 1.04          & 4.66          & 0.042         & 2.16 \\
\midrule
\multirow{4}{*}{Hidden dim}
  & 64                 & 1.10          & 4.95          & 0.046         & 2.33 \\
  & 96                 & 1.05          & 4.74          & 0.043         & 2.20 \\
  & \cellcolor[rgb]{.886,.937,.855}\textbf{128 (ours)} & \cellcolor[rgb]{.886,.937,.855}\textbf{1.01} & \cellcolor[rgb]{.886,.937,.855}\textbf{4.54} & \cellcolor[rgb]{.886,.937,.855}\textbf{0.04} & \cellcolor[rgb]{.886,.937,.855}\textbf{2.08} \\
  & 256                & 1.03          & 4.66          & 0.041         & 2.12 \\
\bottomrule
\end{tabular}}
\end{table}

\section{SeaStereo-Dataset}
\label{sec:appendix_data_synth}

This work introduces SeaStereo-Dataset, a synthetic underwater stereo dataset containing 40,320 image pairs with depth maps of simulated subsea environments. The dataset can support the development and evaluation of stereo vision methods in underwater settings.

\begin{figure}[t]
    \centering
    \includegraphics[width=0.7\textwidth, page=1]{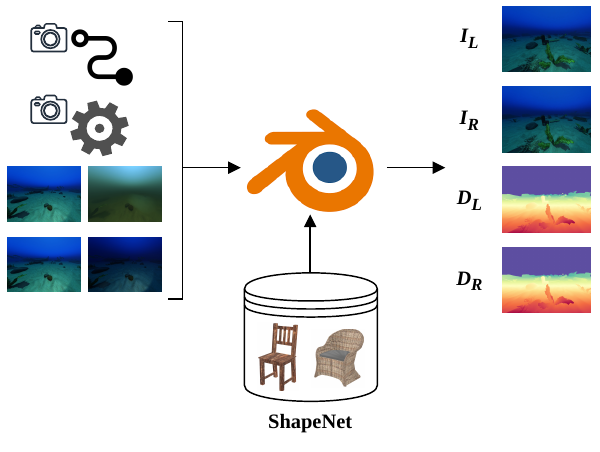}
    \caption{\textbf{Synthesis of the SeaStereo-Dataset:} The Blender Python API automated the rendering pipeline by iterating through each configuration of features (camera path, camera model, water type, depth and ShapeNet object subset). Each configuration generated stereo image pairs (\textit{\textbf{I\textsubscript{L}, I\textsubscript{R}}}) and disparity maps (\textit{\textbf{D\textsubscript{L}, D\textsubscript{R}}}).}
    \label{data synth}
\end{figure}

\paragraph{Methodology.}
\autoref{data synth} illustrates the synthesis pipeline of the SeaStereo-Dataset in Blender. The Blender Python API was used to automate the dataset generation. For each configuration, the features, including camera path, camera model, ShapeNetCore~\cite{shapenet2015} object subset, water type and seafloor depth, were applied to the scene. Stereo RGB images and corresponding depth maps were then rendered for the left and right camera views. The pipeline iterated through combinations of the dataset features to produce a diverse set of scene conditions.

\paragraph{Dataset Features.}
A synthetic dataset is advantageous because of the customisability it affords. The underwater scene consists of submerged foreground objects from ShapeNetCore, with realistic marine objects in the background, including fish, coral, shipwrecks, seaweed and rocks. Similar to UWStereo~\cite{lv2025uwstereo}, we introduce variations in the camera, illumination conditions and environment to construct a diverse stereo corpus. The camera path, camera model (focal length, interocular distance), water type, seafloor depth and ShapeNet object selection can all vary between configurations. An axis-aligned bounding box constraint was also applied to prevent overlapping placements of ShapeNet objects.

\begin{figure}[t]
    \centering
    \includegraphics[width=0.32\textwidth]{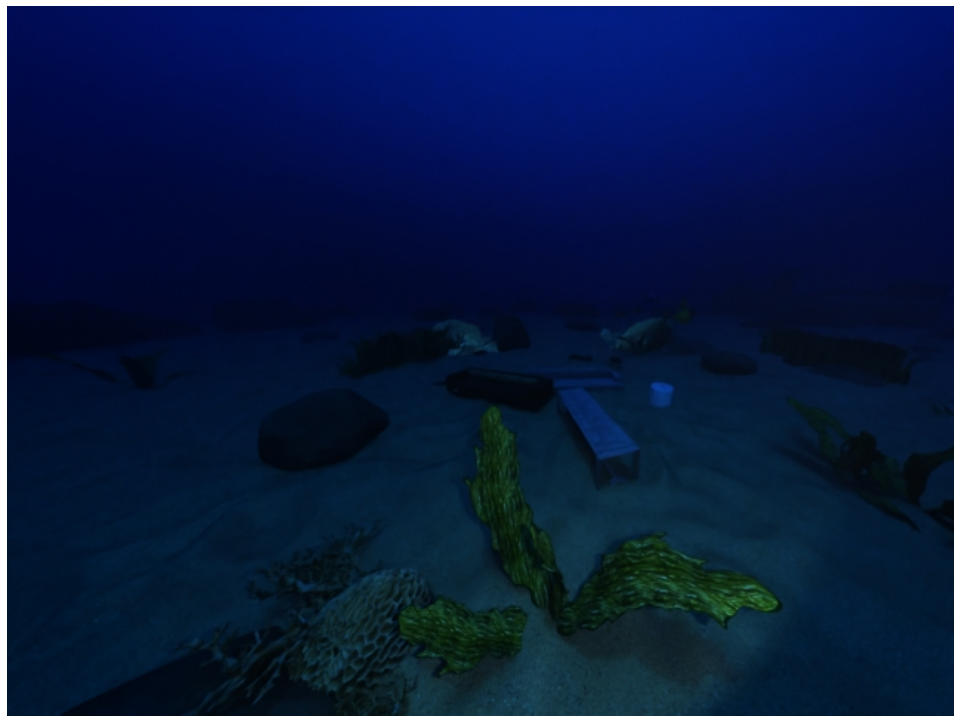}
    \hfill
    \includegraphics[width=0.32\textwidth]{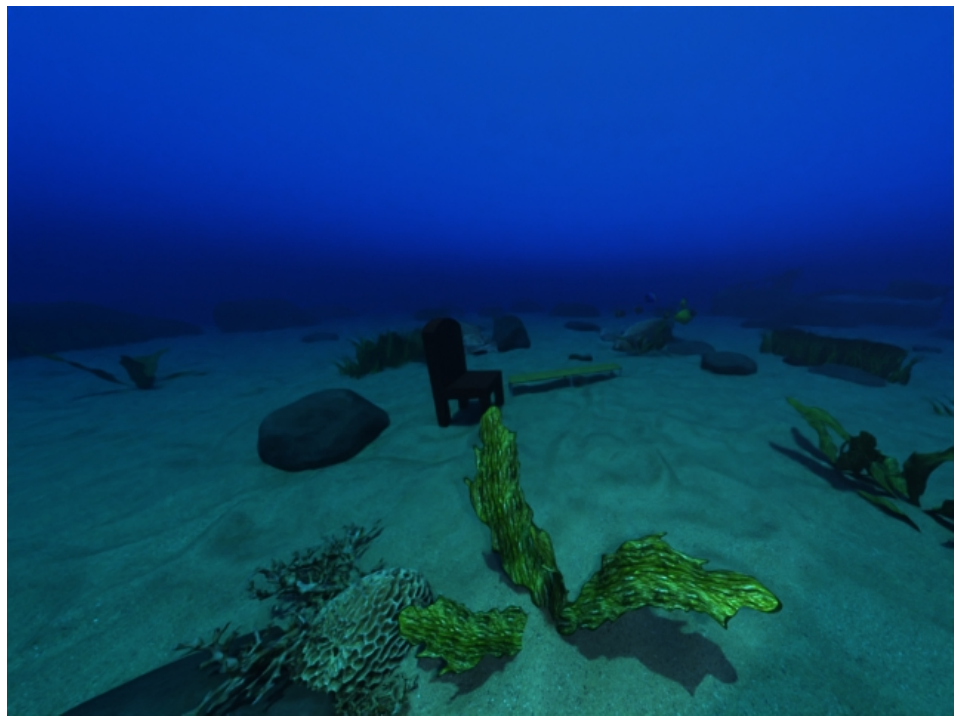}
    \hfill
    \includegraphics[width=0.32\textwidth]{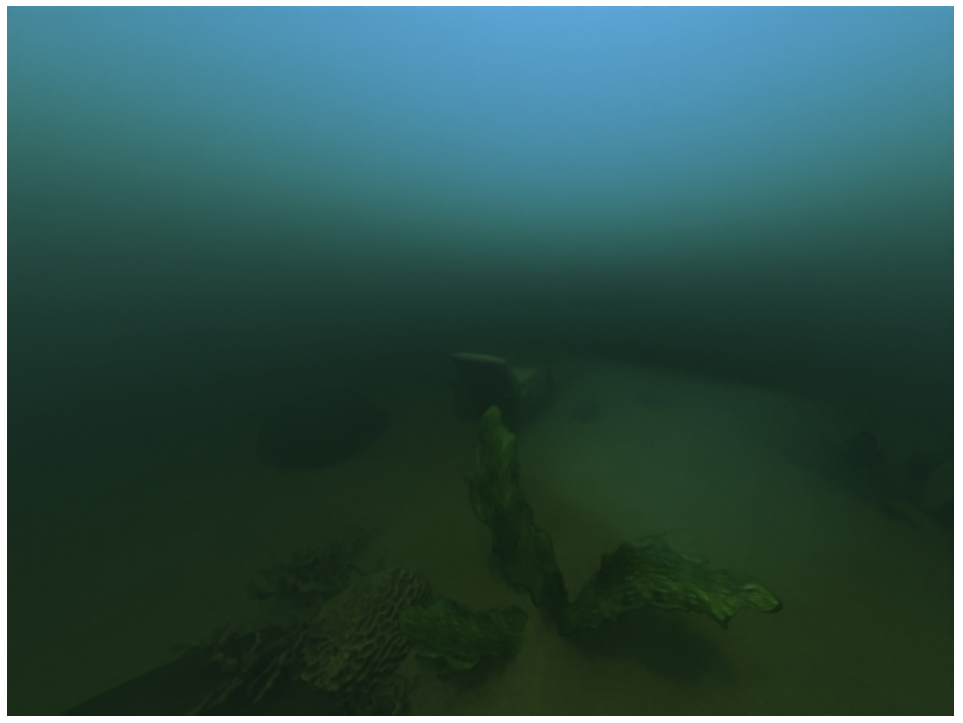}

    \caption{\textbf{SeaStereo-Dataset Examples:} Left camera renders from SeaStereo-Dataset, illustrating variation in seafloor depth and water type. These examples demonstrate the controlled changes in underwater visibility and environmental configuration used during dataset generation.}
    \label{fig:sea_synth_examples}
\end{figure}

The available water conditions are modelled by the Jerlov types I, IA, IB, II, IC, III and 3C. These are distinguished by their light absorption and scattering characteristics, leading to clear and turbid water, as shown in \autoref{fig:sea_synth_examples}.

\paragraph{Dataset Details.}

In our implementation, 1008 configurations were animated for 40 frames each, leading to 40,320 stereo image pairs. Seven camera trajectories were manually defined. Four camera models were applied, which were derived from combining two focal length values with two interocular distance values. A seafloor depth was then randomly sampled from a predefined range. Highly turbid water types (Jerlov III and 3C) were restricted to shallow depths to maintain image visibility, whereas clearer water types were sampled across both shallow and deep ranges to introduce variation in lighting conditions. At each selected depth, between three and five ShapeNet objects were randomly chosen and placed within the scene. This process was repeated for three iterations per configuration, with different randomly sampled depths within the corresponding depth range.

\begin{figure*}[t]
    \centering

    \resizebox{\textwidth}{!}{%
\begin{tabular}{ccccccccc}
        \textbf{Left} & \textbf{Right} & \textbf{Disparity} &
        \textbf{Left} & \textbf{Right} & \textbf{Disparity} &
        \textbf{Left} & \textbf{Right} & \textbf{Disparity} \\[1mm]

        \includegraphics[width=0.1\textwidth]{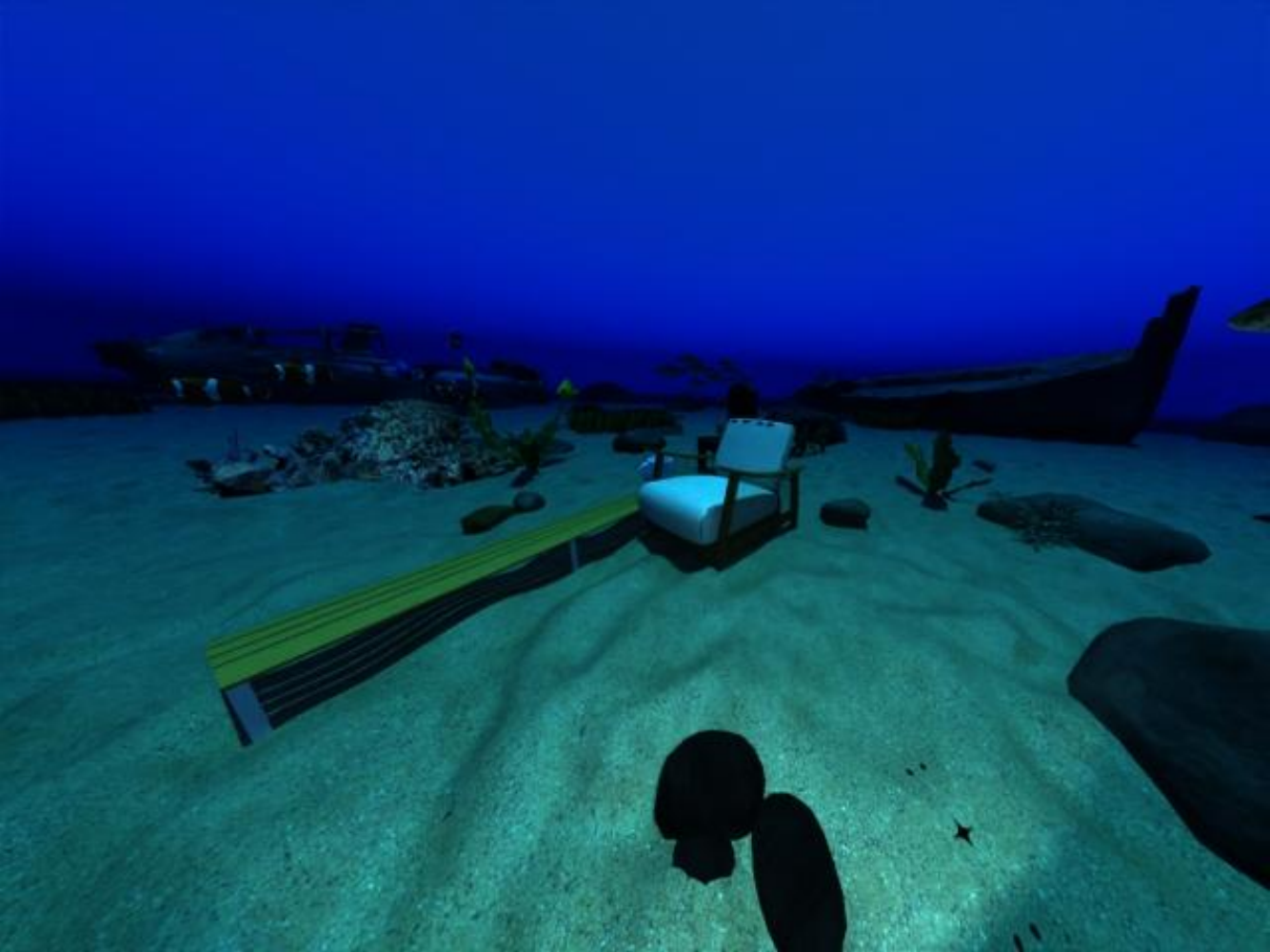} &
        \includegraphics[width=0.1\textwidth]{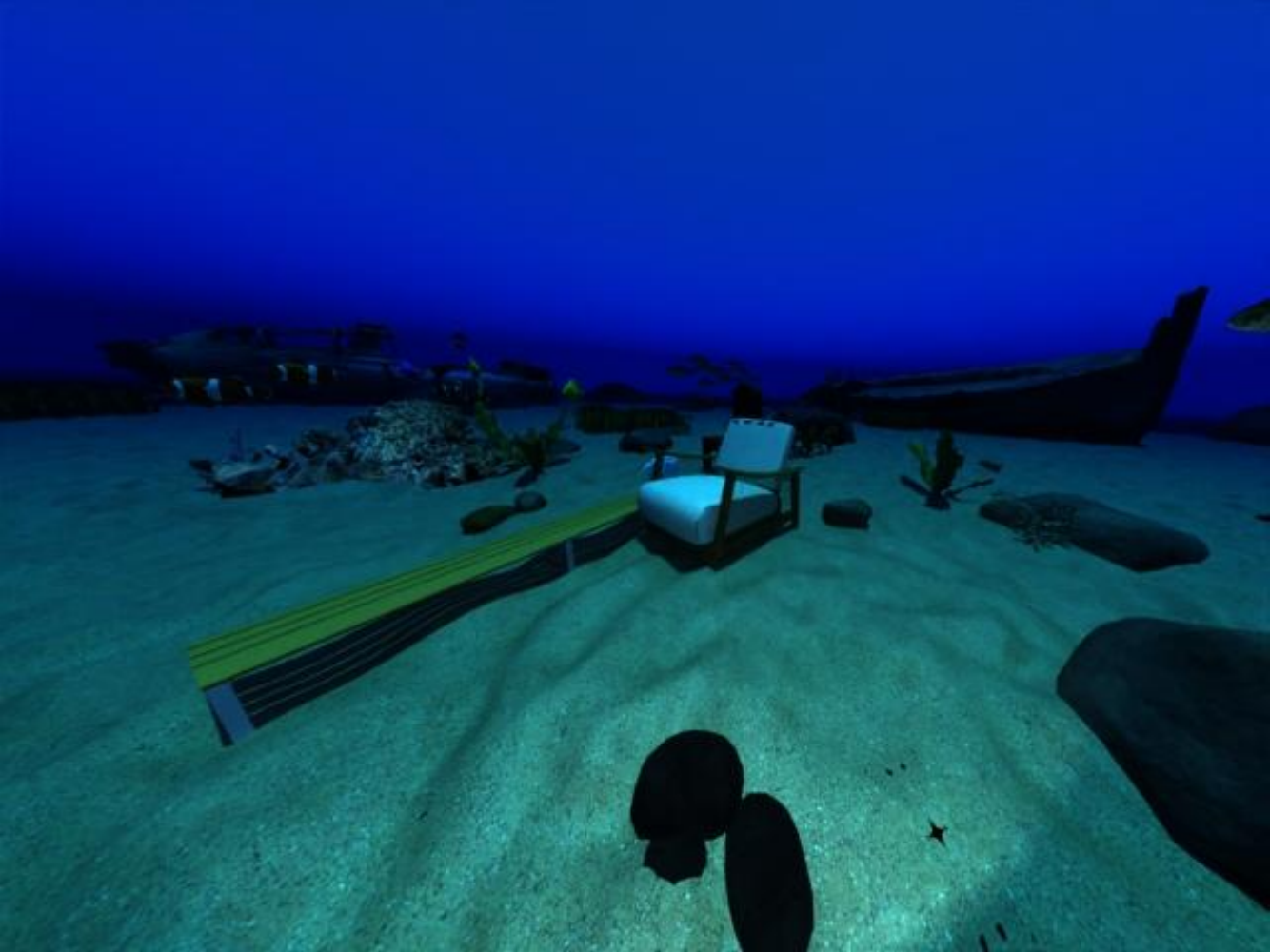} &
        \includegraphics[width=0.1\textwidth]{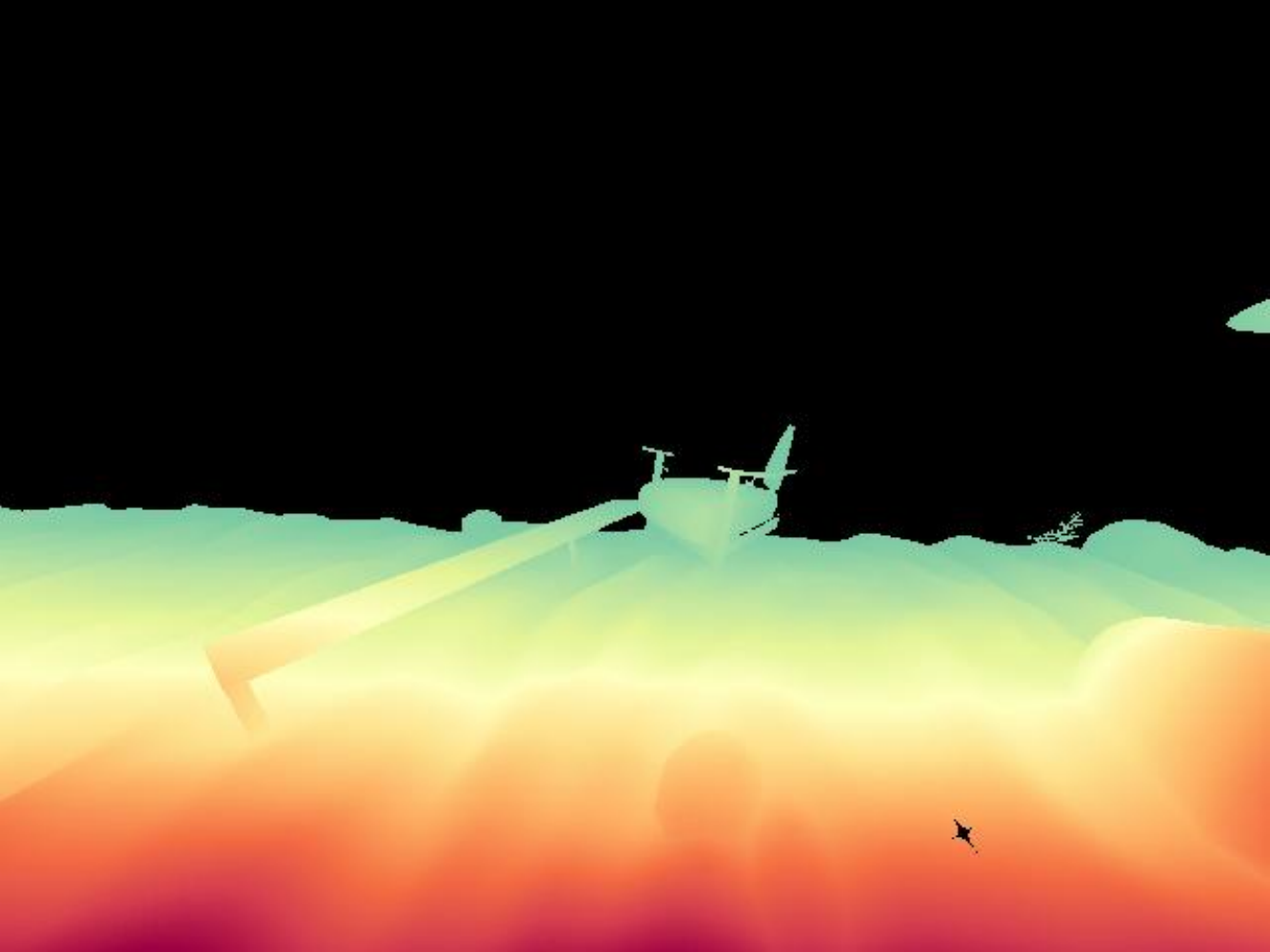} &

        \includegraphics[width=0.1\textwidth]{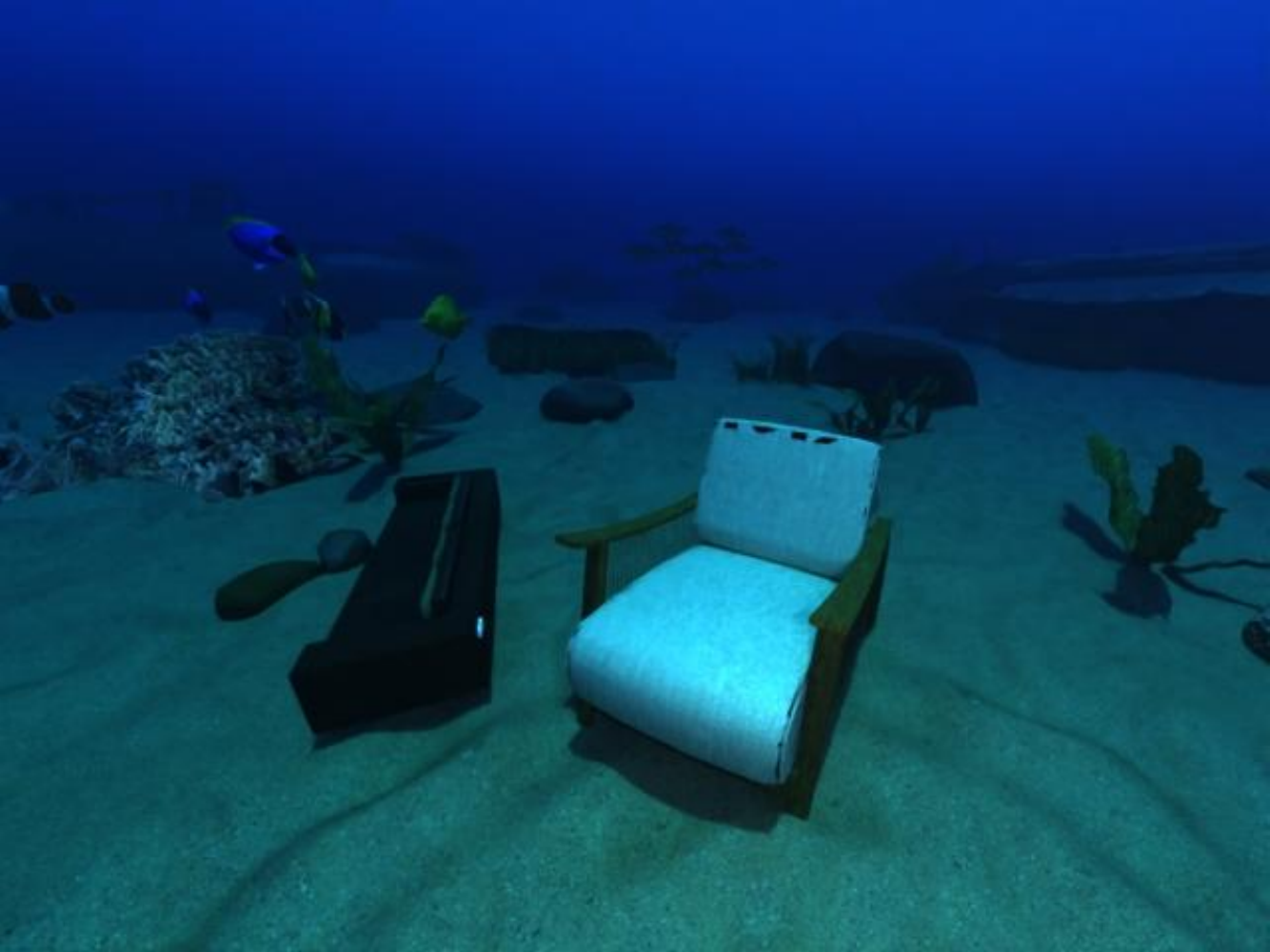} &
        \includegraphics[width=0.1\textwidth]{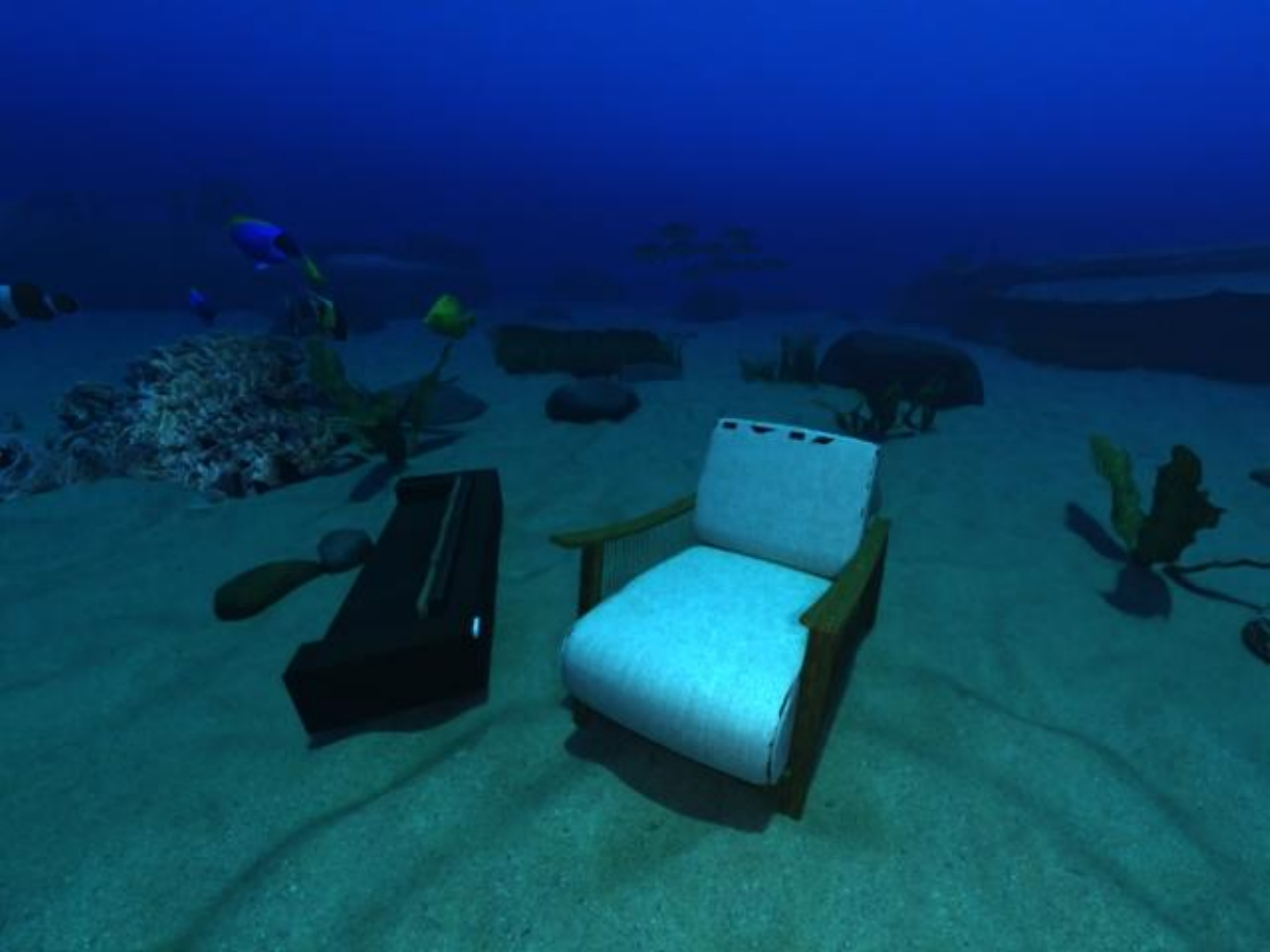} &
        \includegraphics[width=0.1\textwidth]{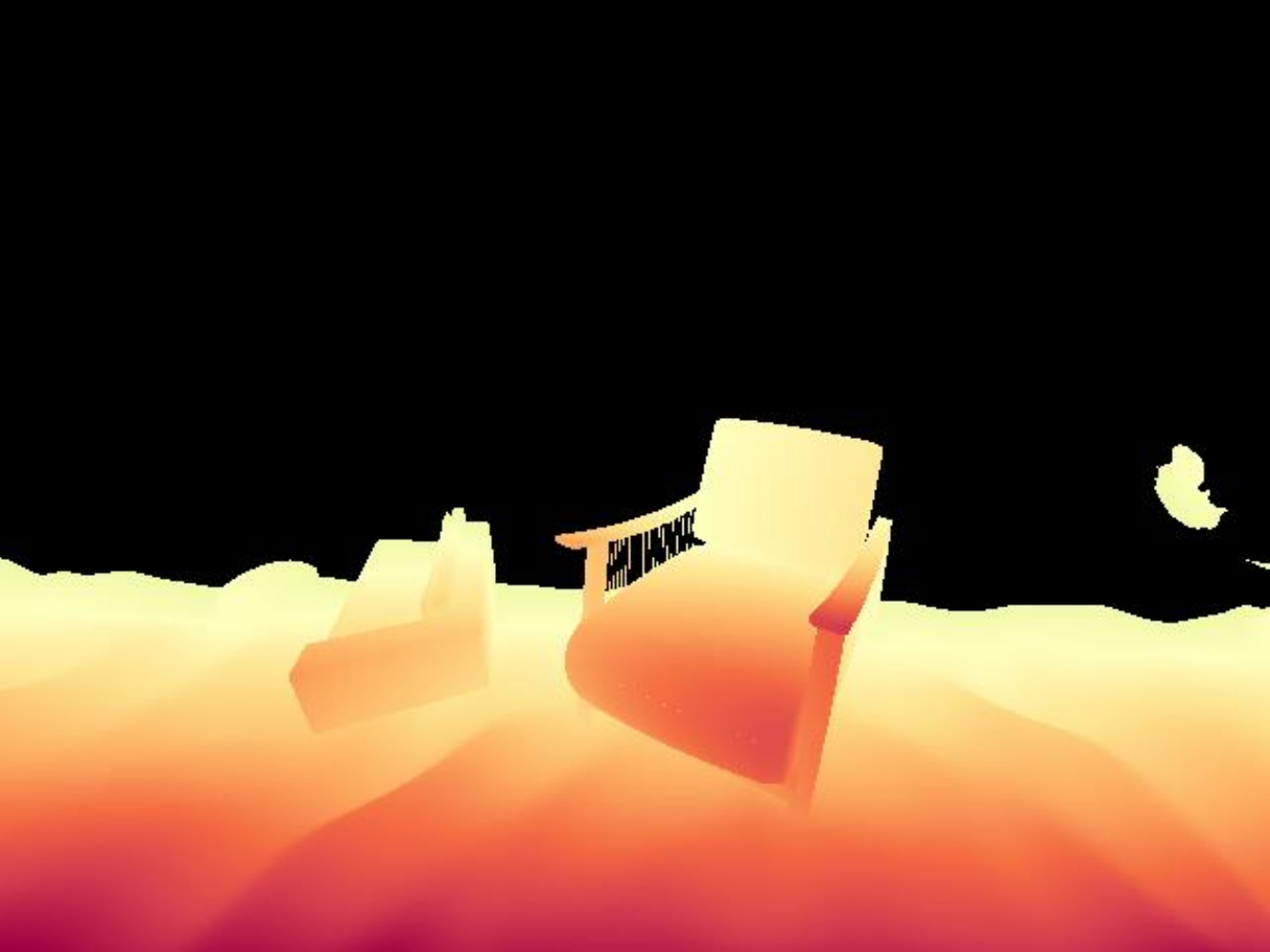} &

        \includegraphics[width=0.1\textwidth]{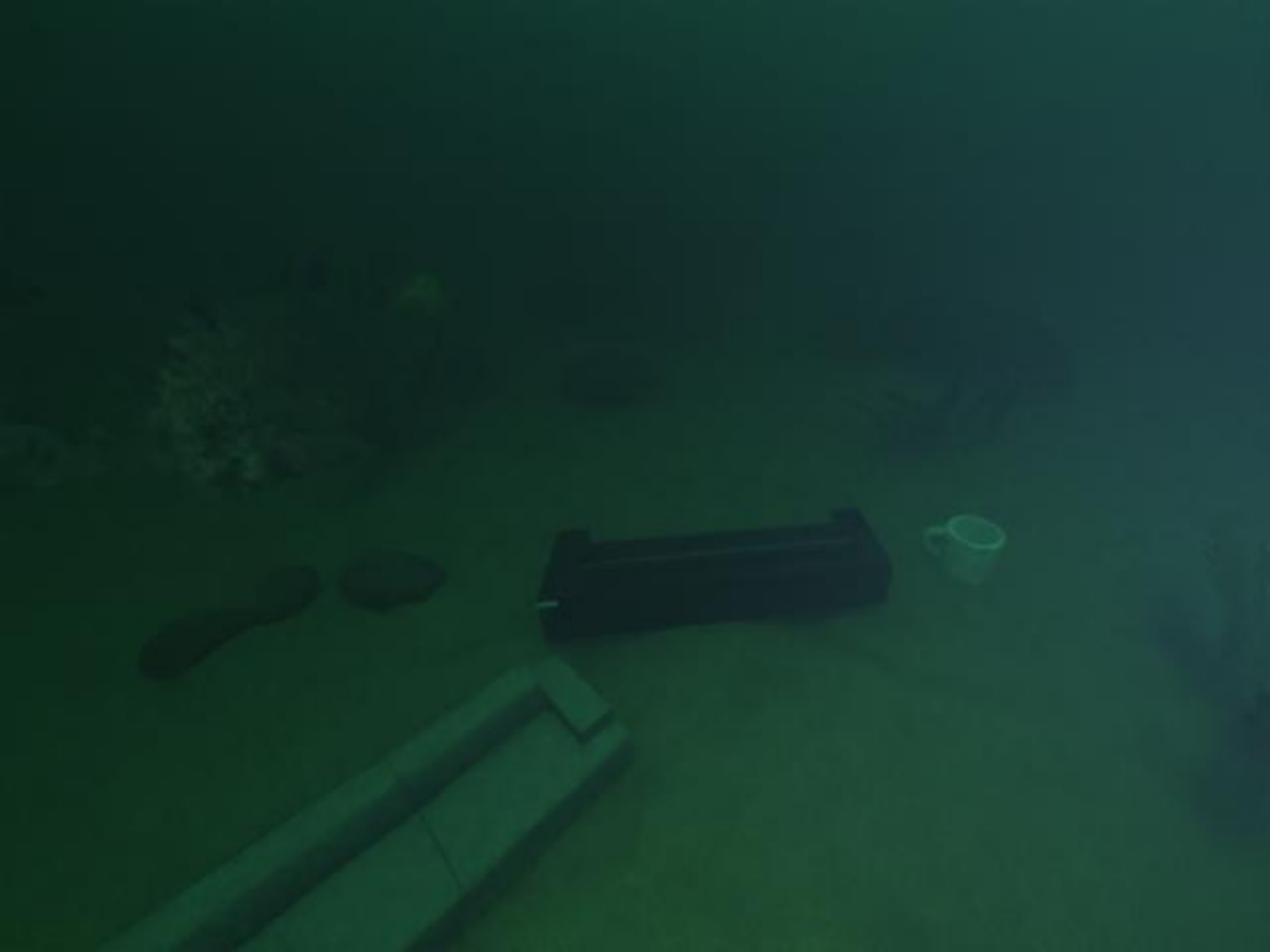} &
        \includegraphics[width=0.1\textwidth]{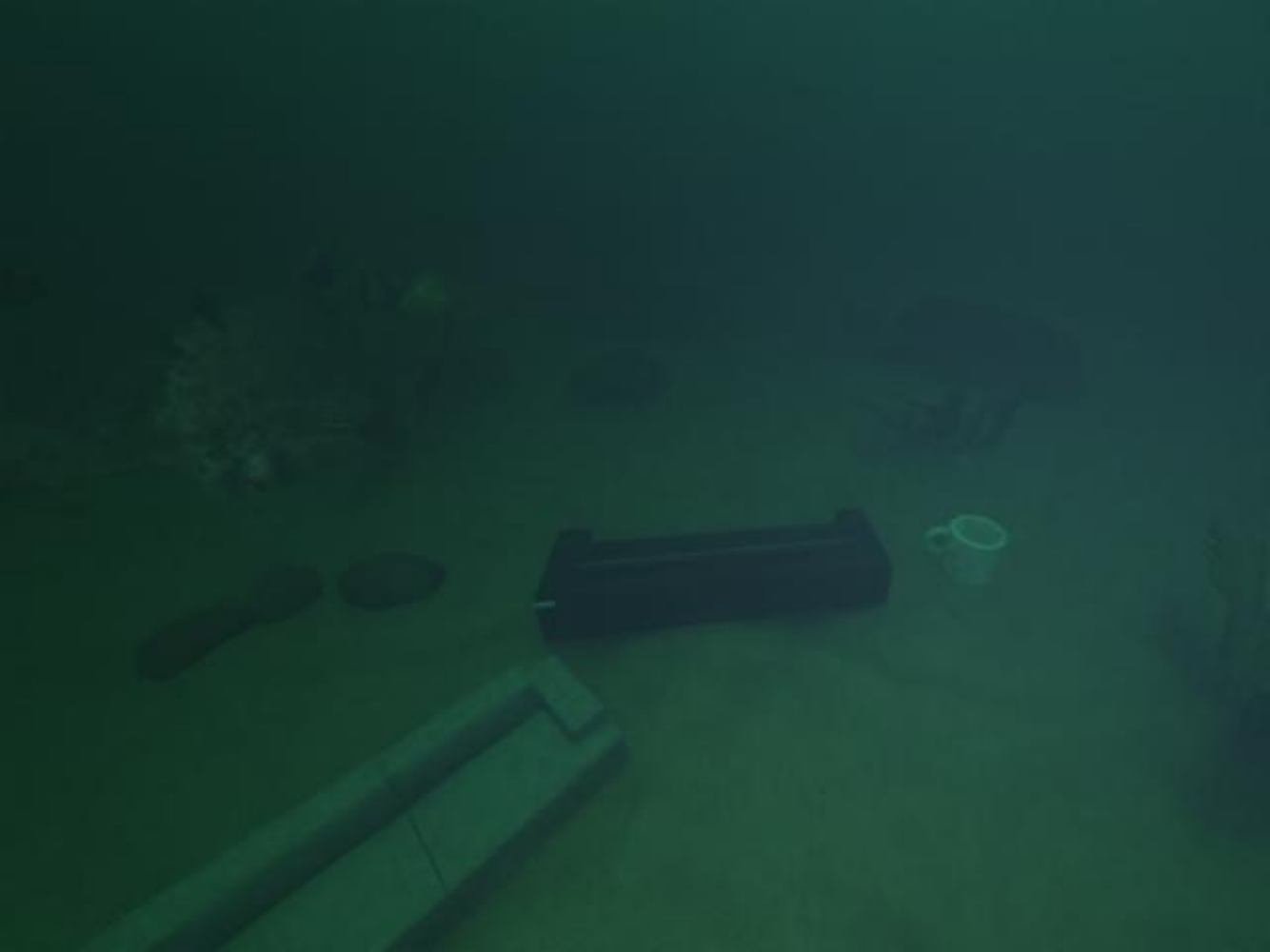} &
        \includegraphics[width=0.1\textwidth]{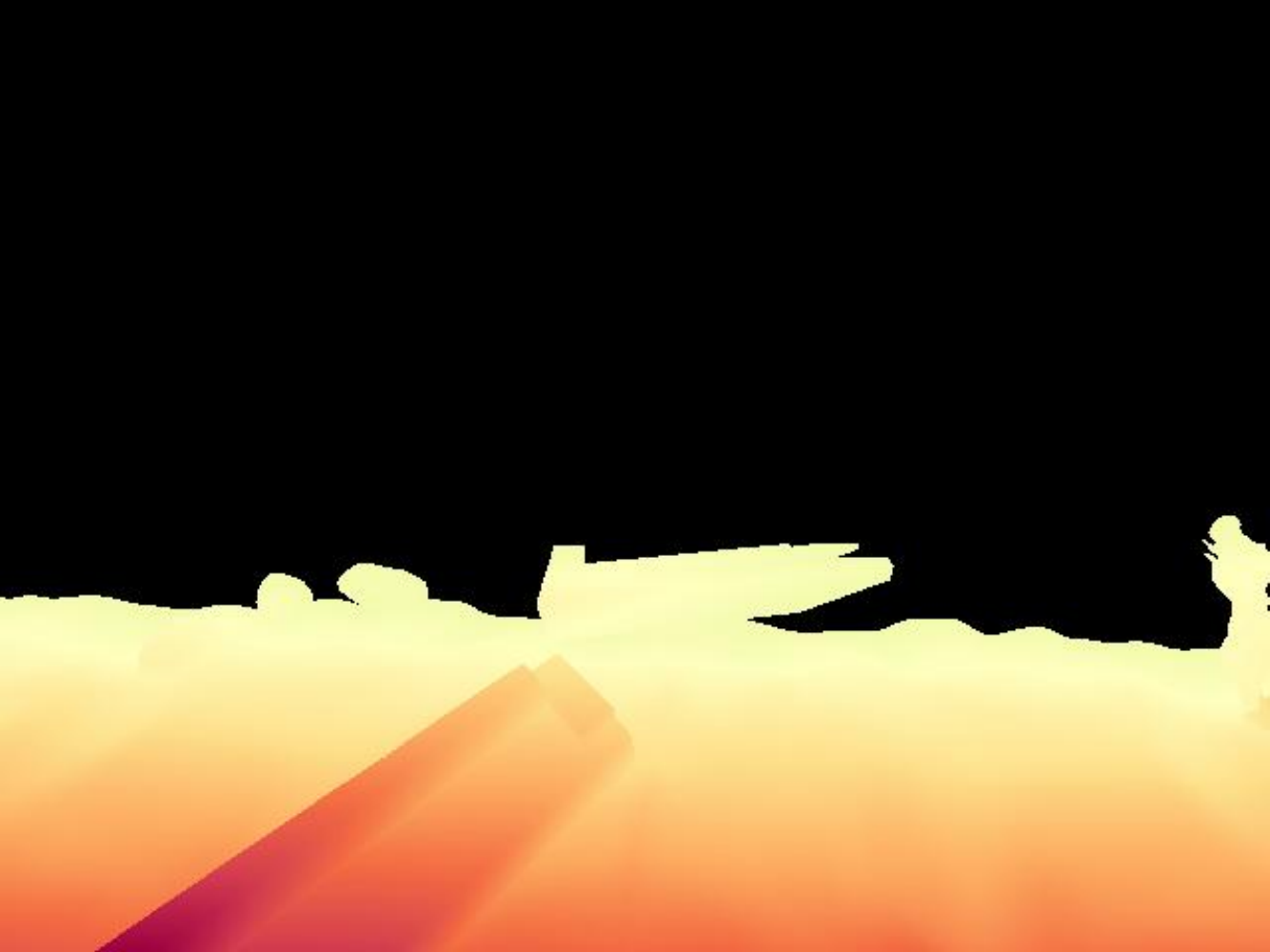} \\

        \includegraphics[width=0.1\textwidth]{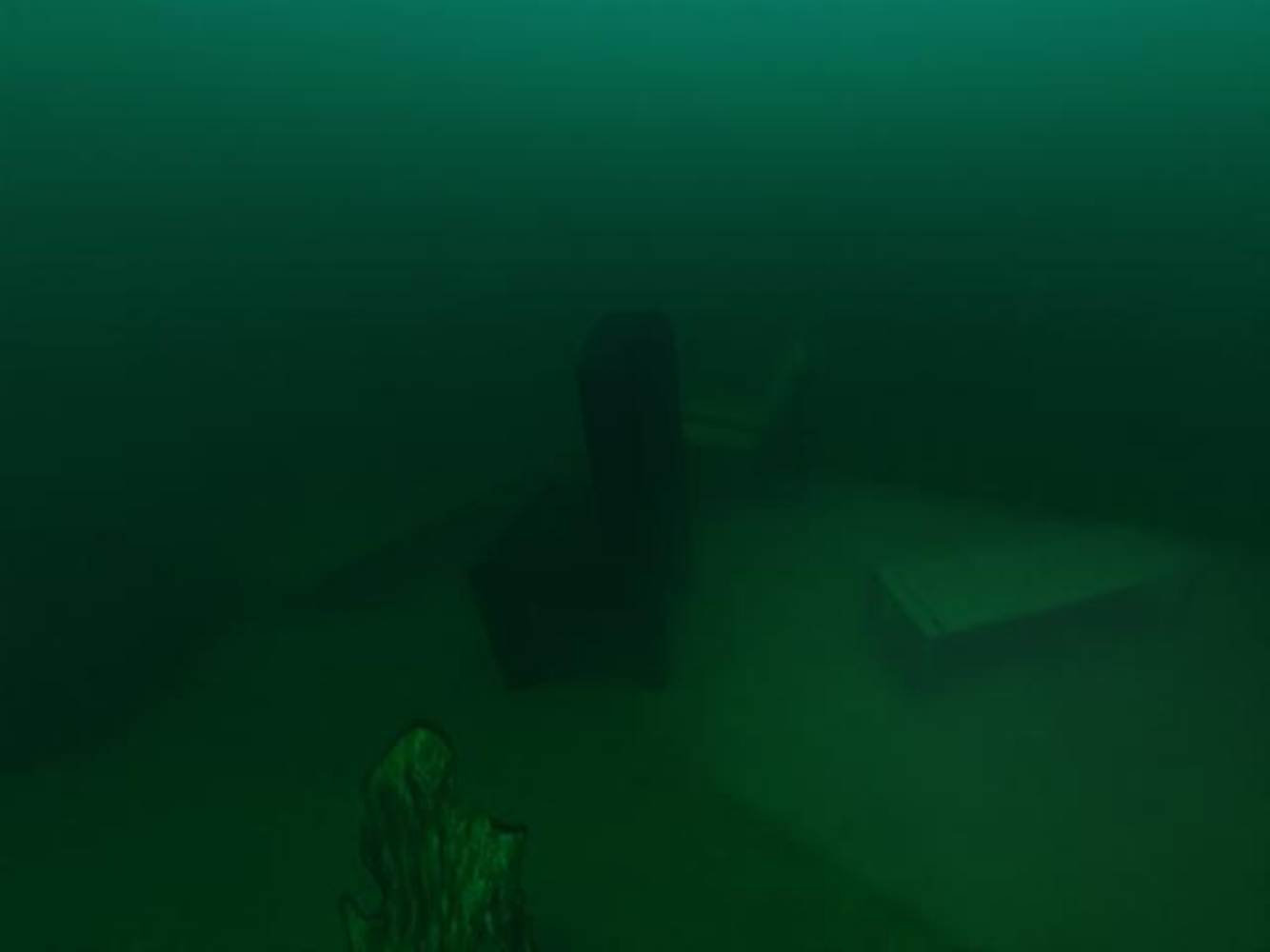} &
        \includegraphics[width=0.1\textwidth]{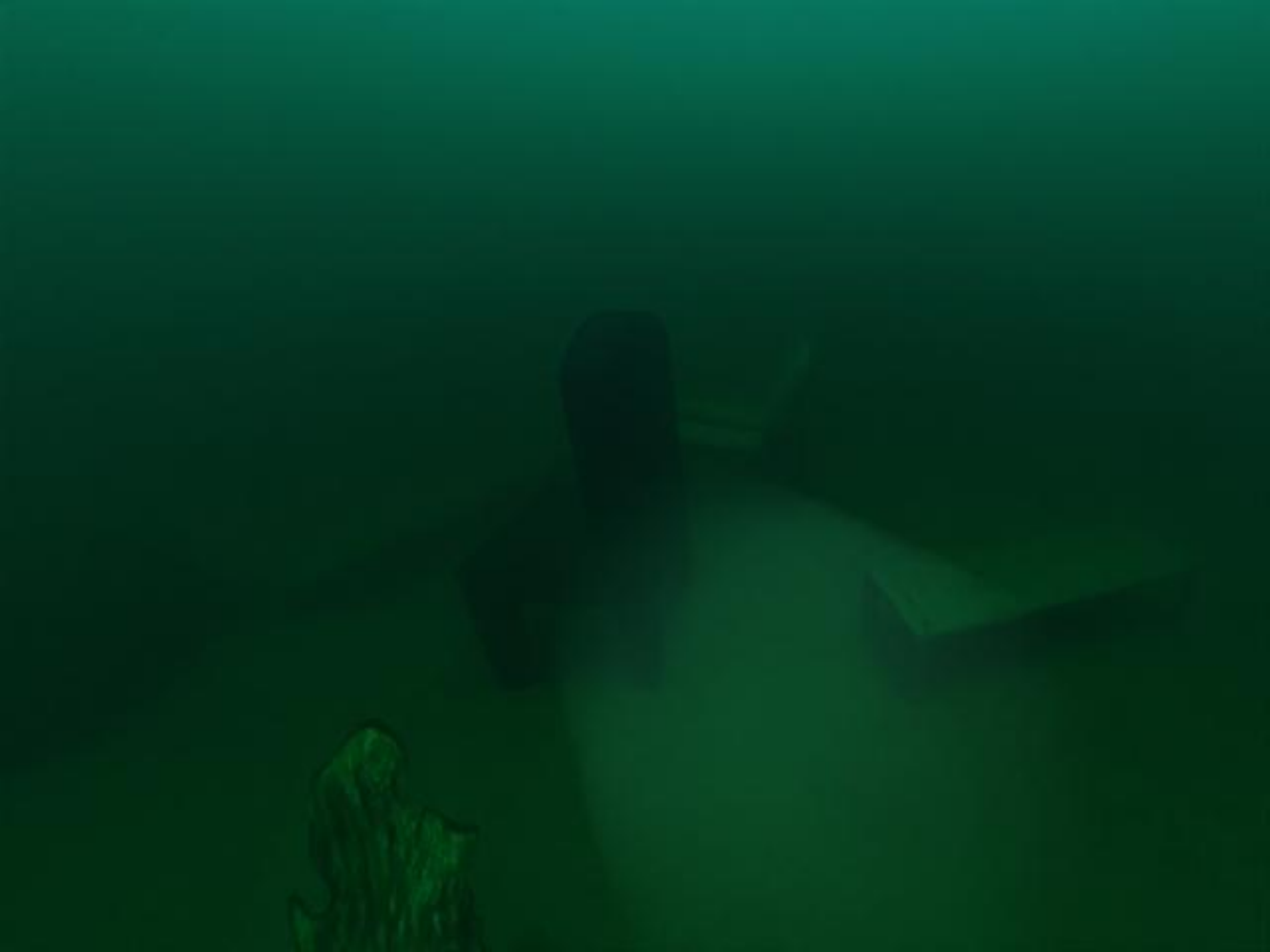} &
        \includegraphics[width=0.1\textwidth]{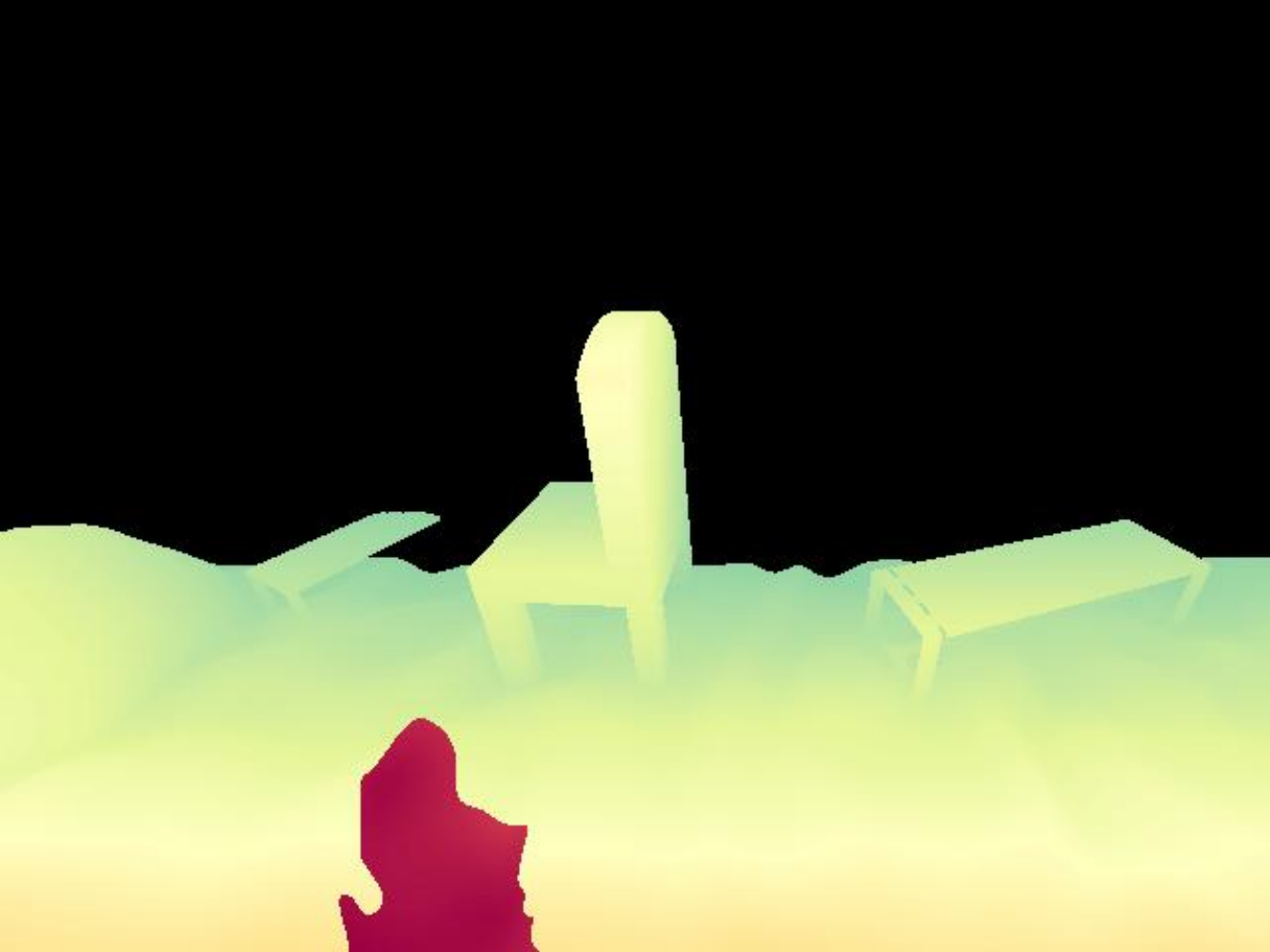} &

        \includegraphics[width=0.1\textwidth]{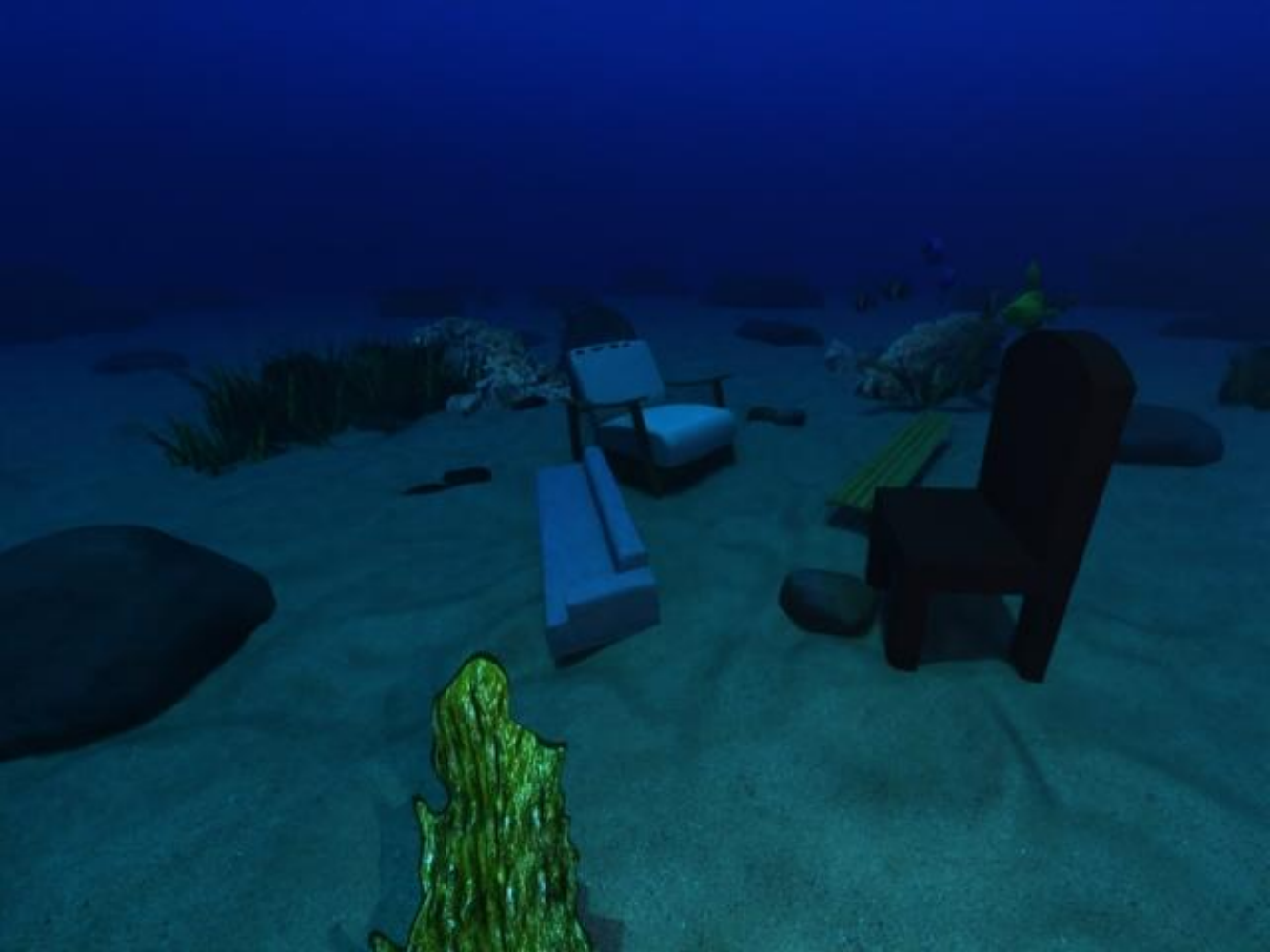} &
        \includegraphics[width=0.1\textwidth]{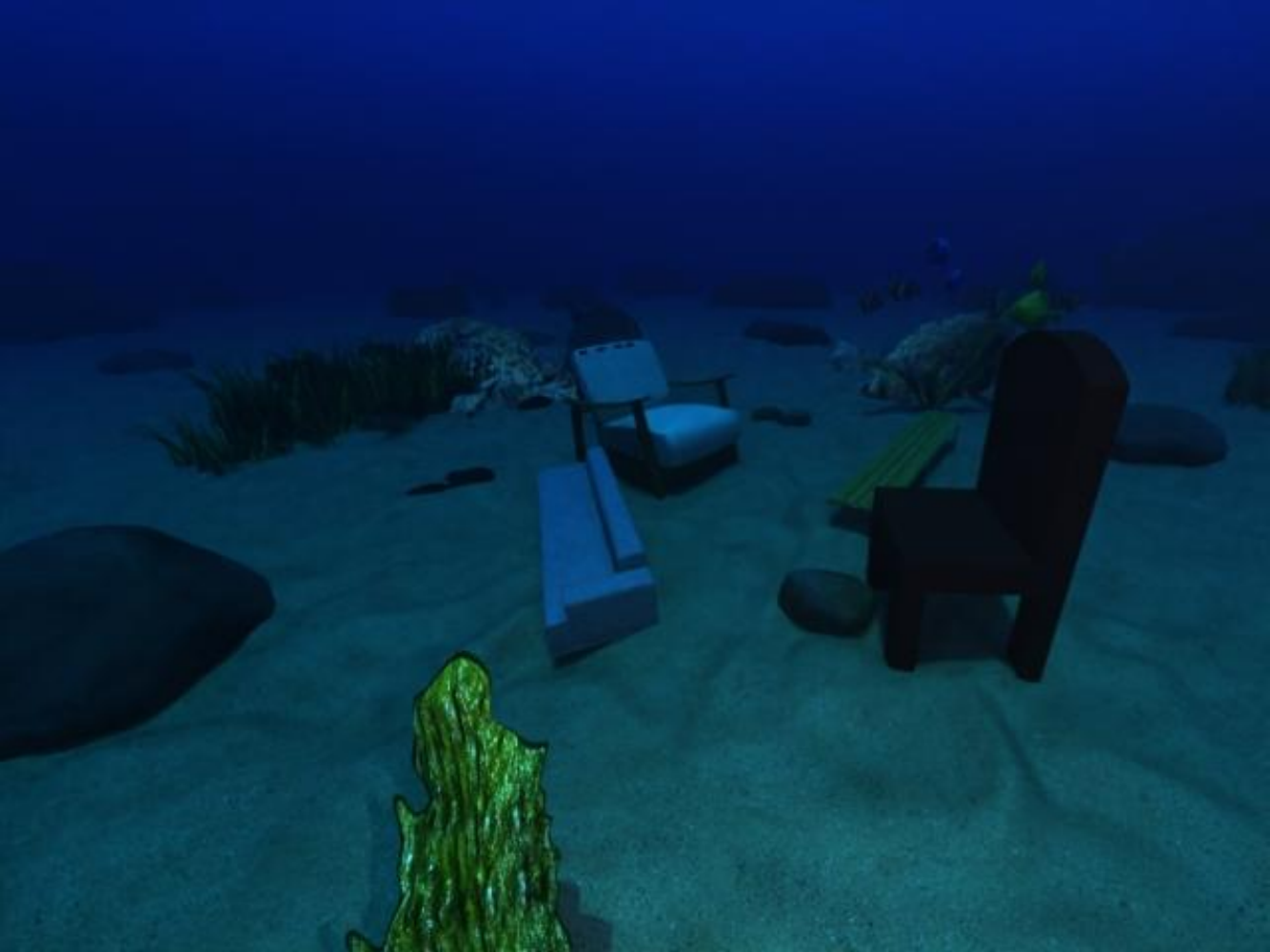} &
        \includegraphics[width=0.1\textwidth]{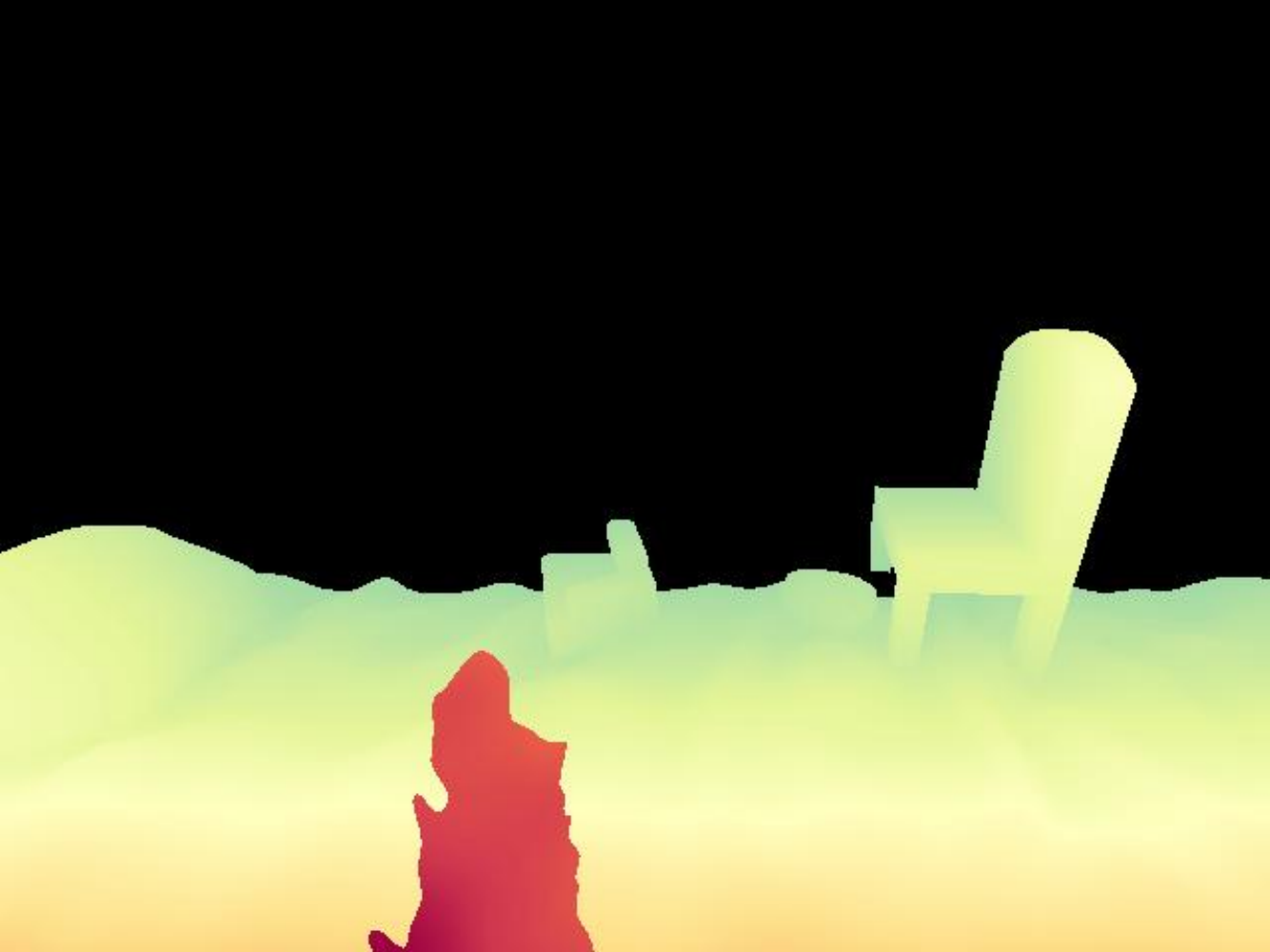} &

        \includegraphics[width=0.1\textwidth]{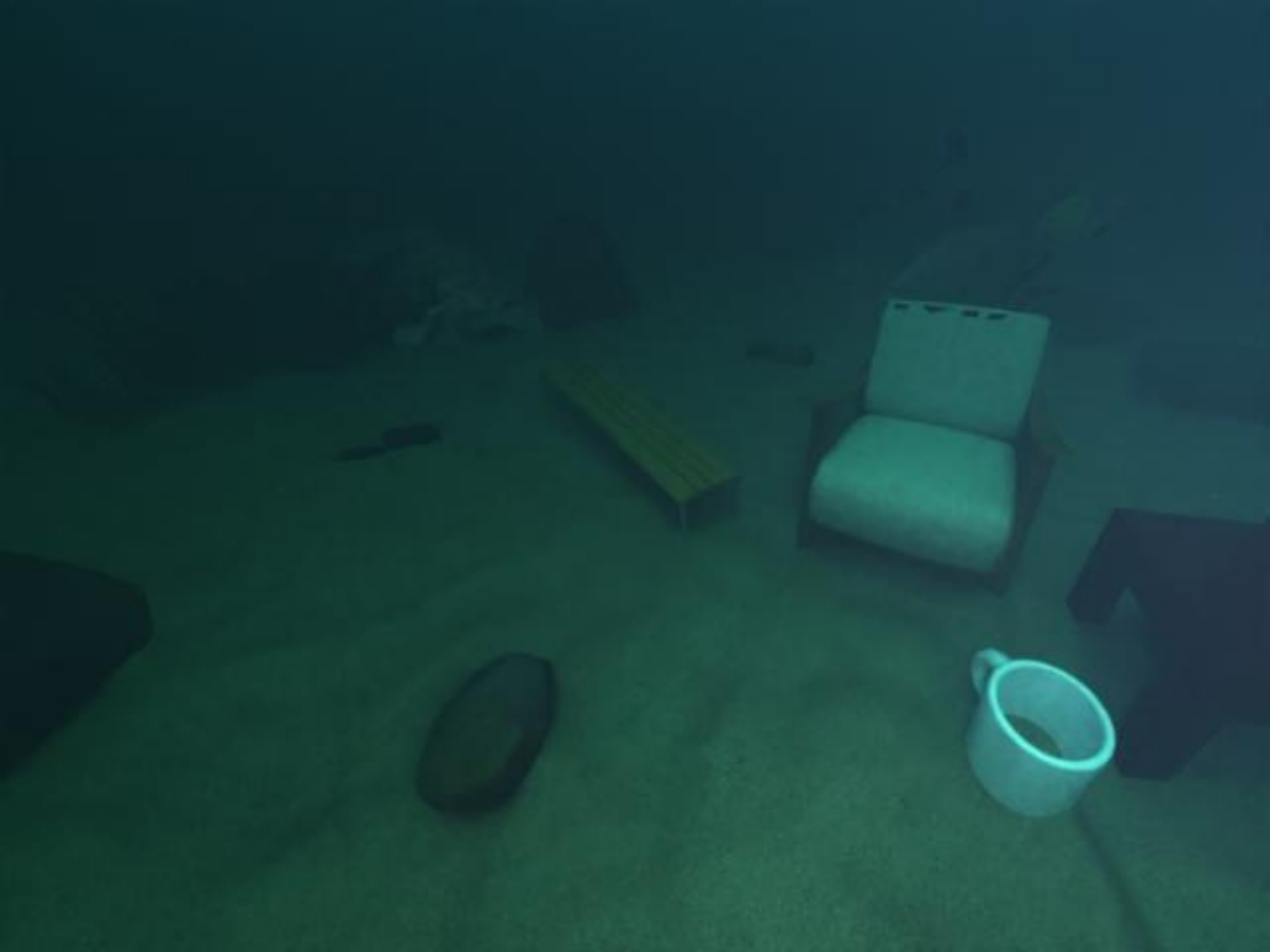} &
        \includegraphics[width=0.1\textwidth]{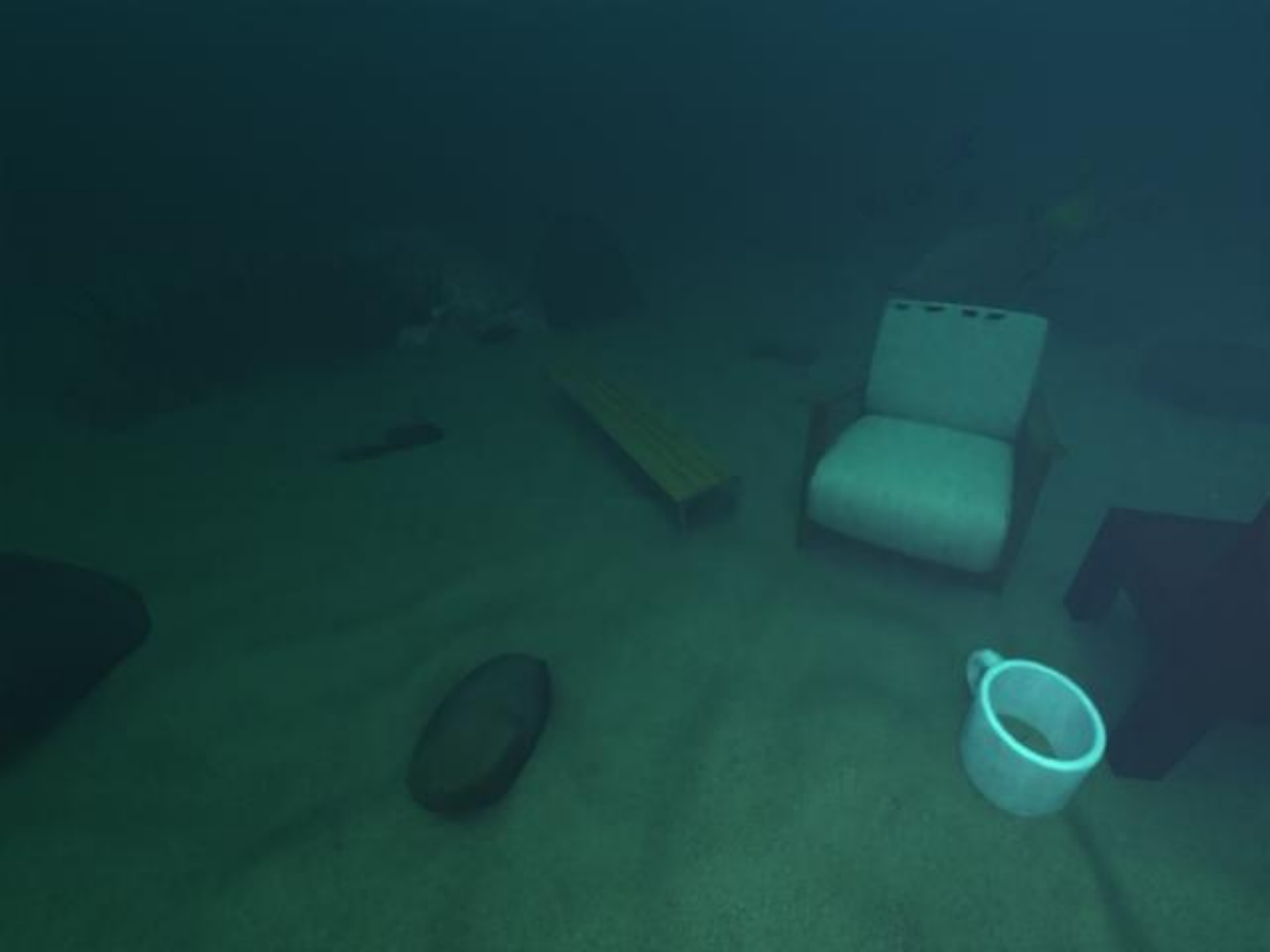} &
        \includegraphics[width=0.1\textwidth]{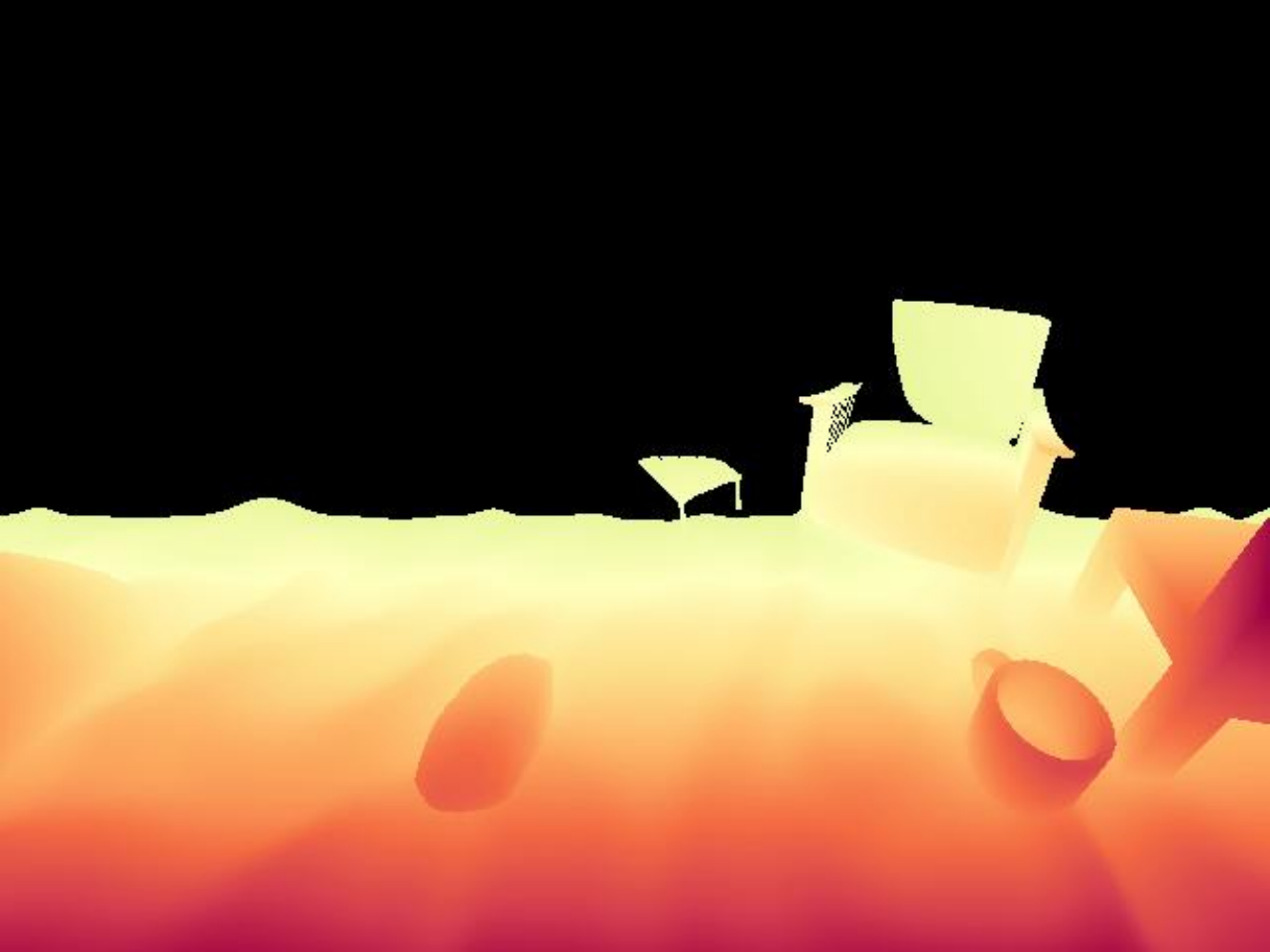} \\

        \includegraphics[width=0.1\textwidth]{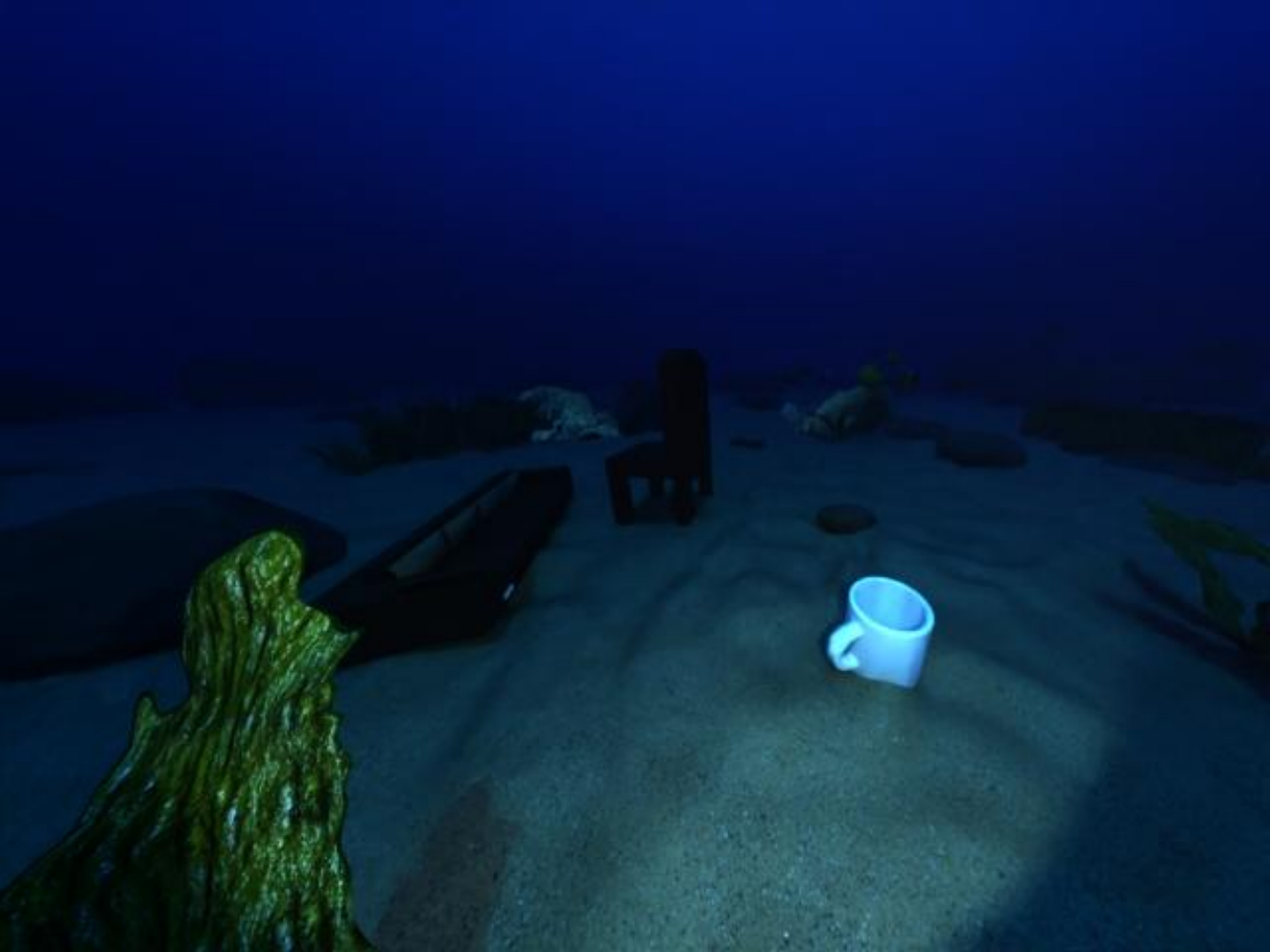} &
        \includegraphics[width=0.1\textwidth]{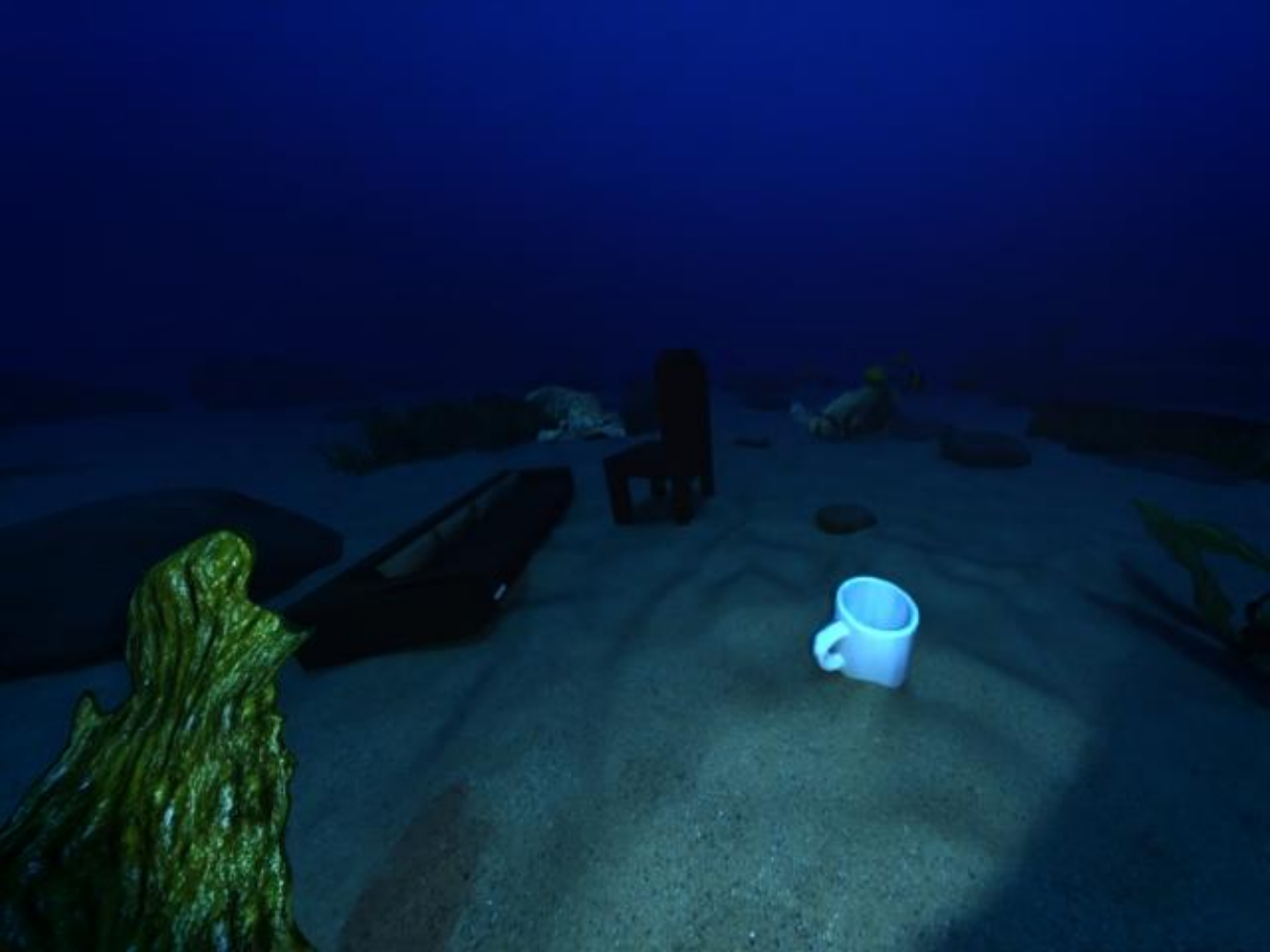} &
        \includegraphics[width=0.1\textwidth]{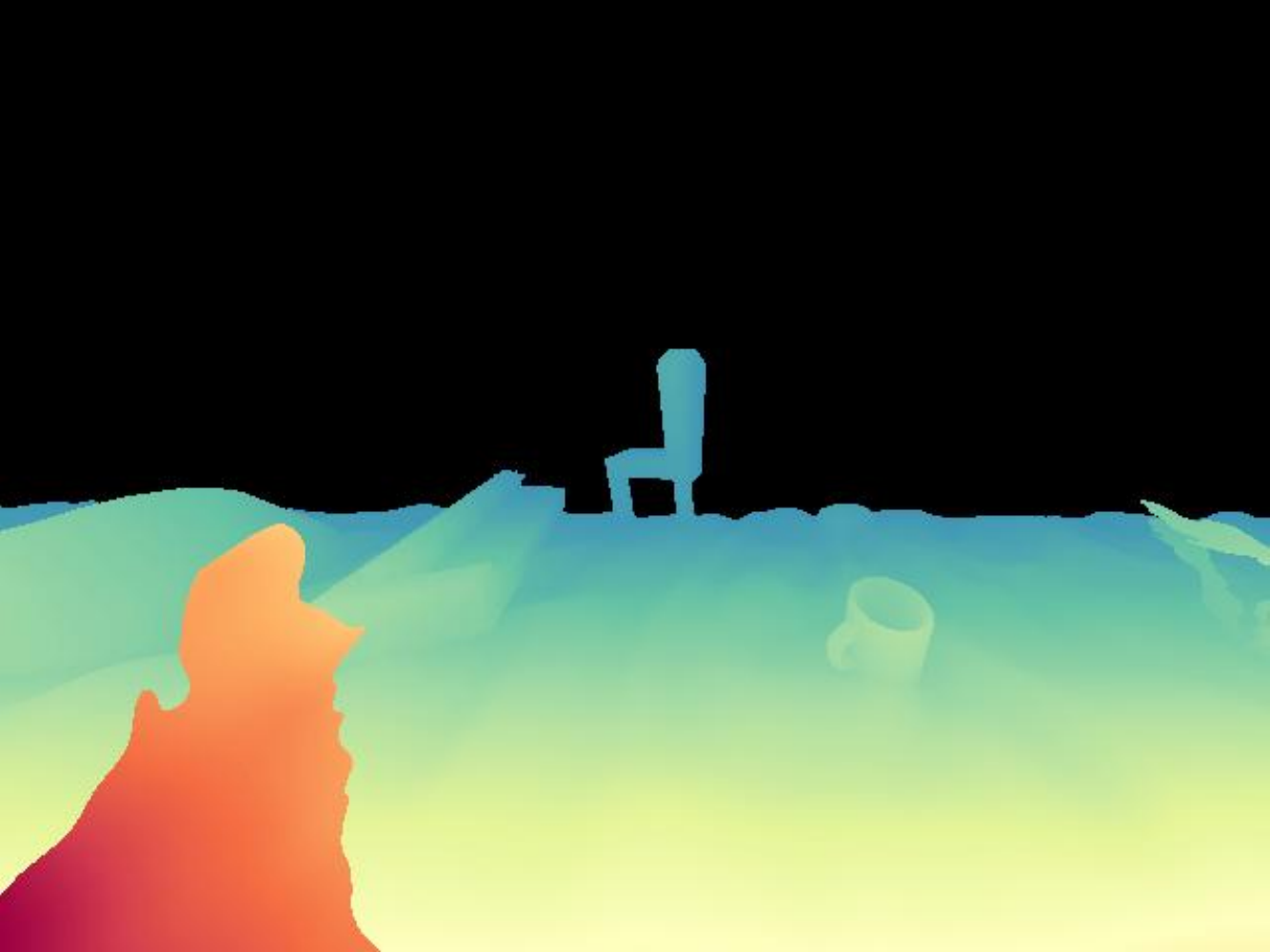} &

        \includegraphics[width=0.1\textwidth]{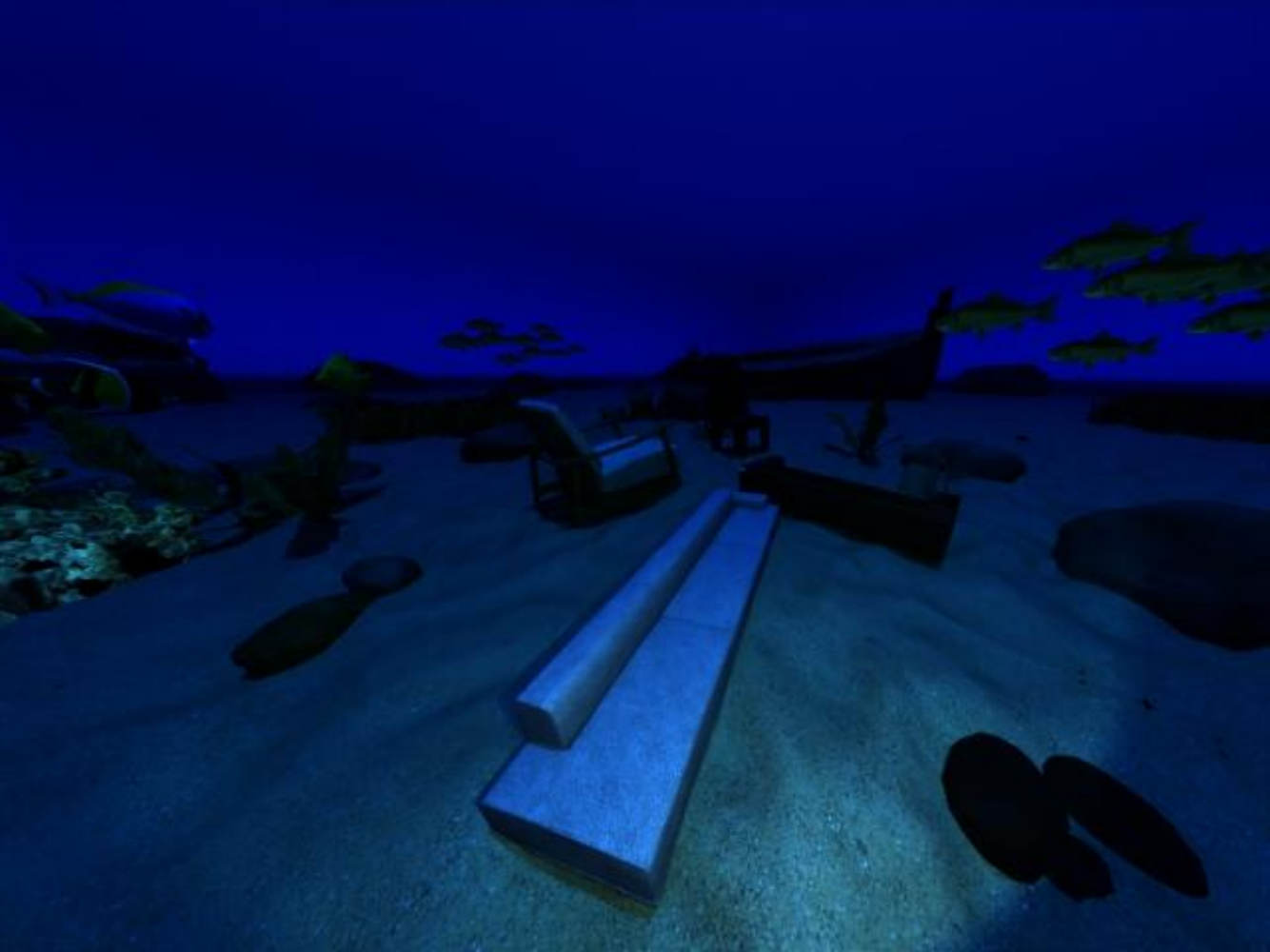} &
        \includegraphics[width=0.1\textwidth]{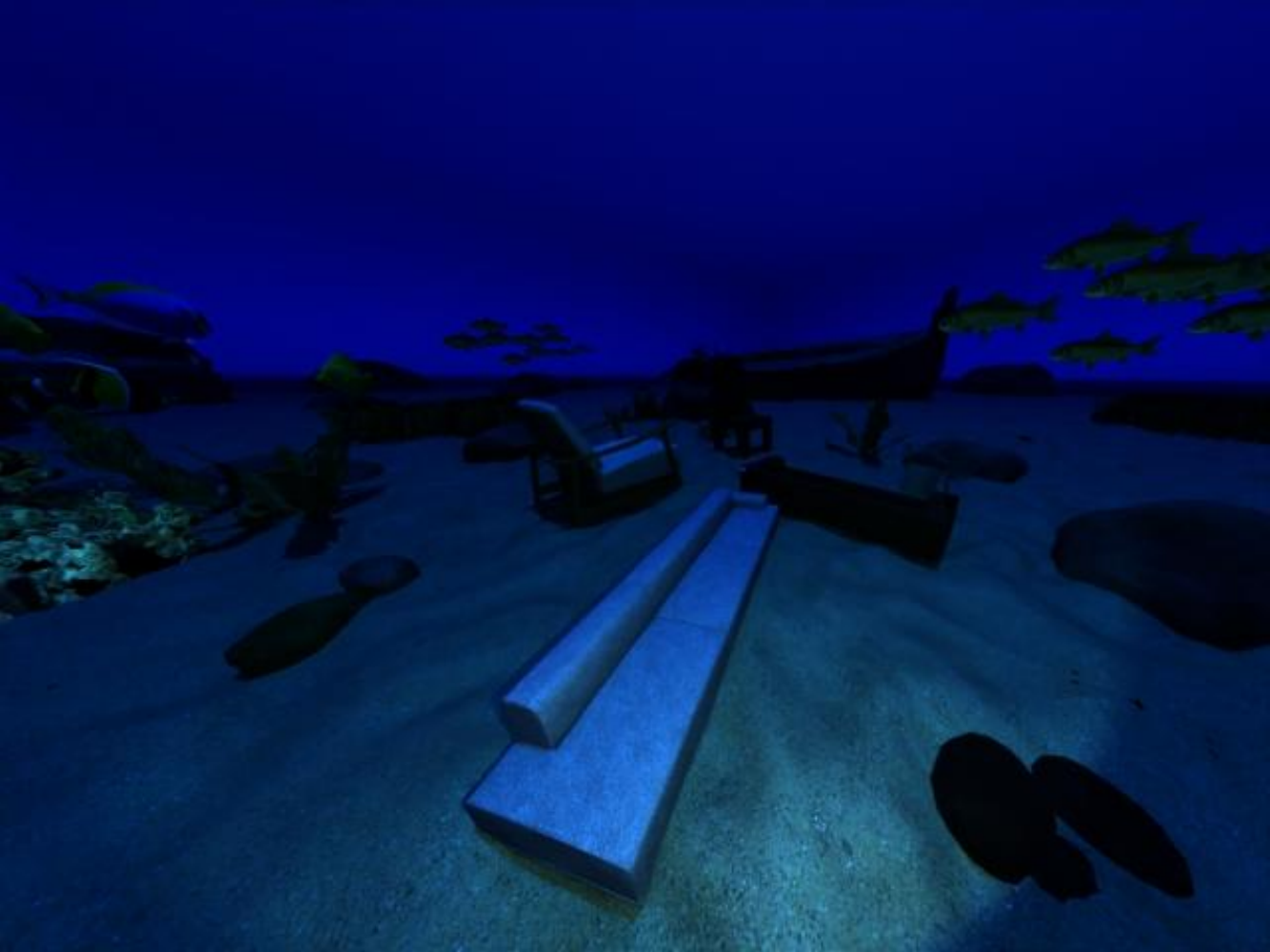} &
        \includegraphics[width=0.1\textwidth]{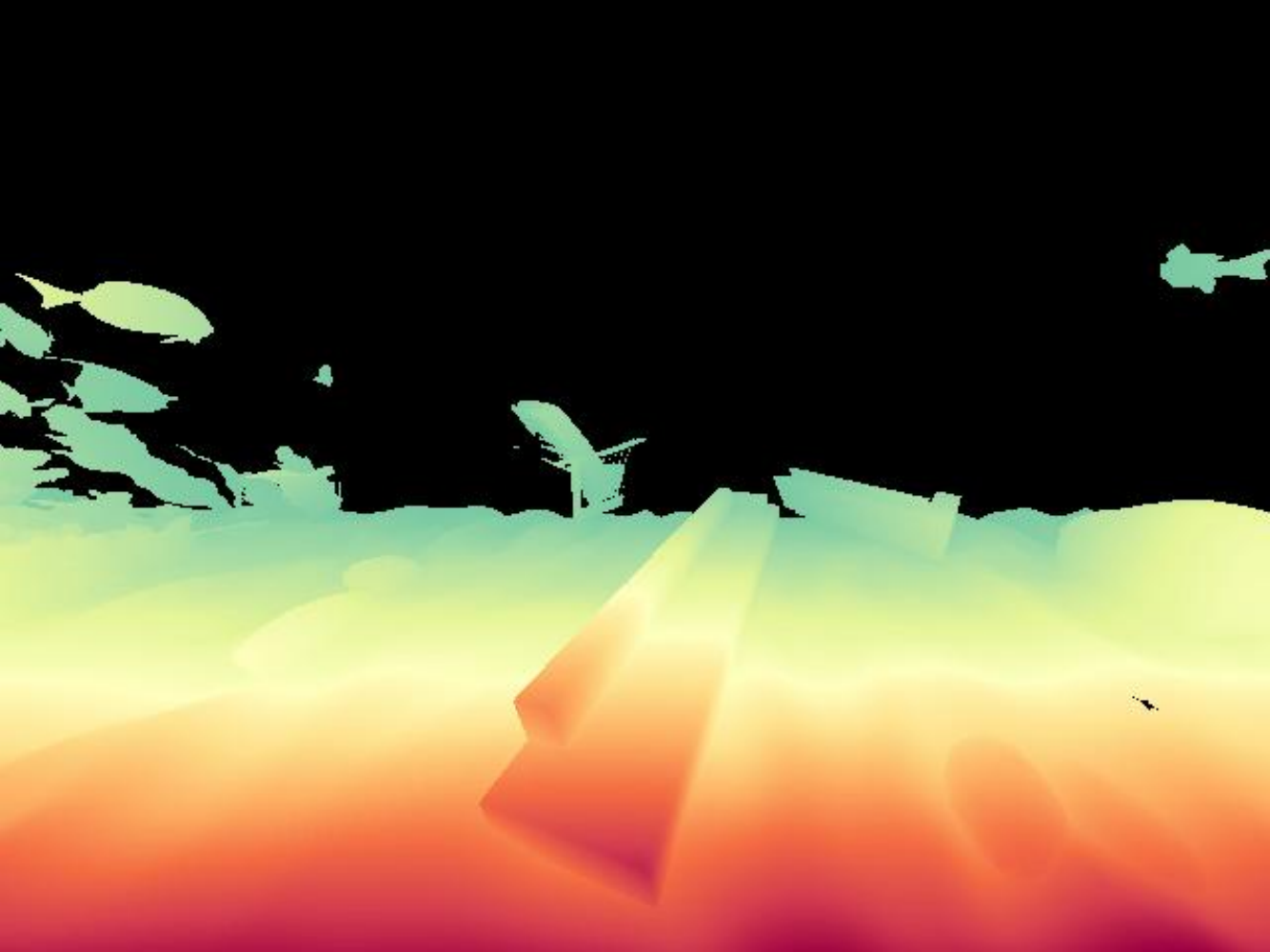} &

        \includegraphics[width=0.1\textwidth]{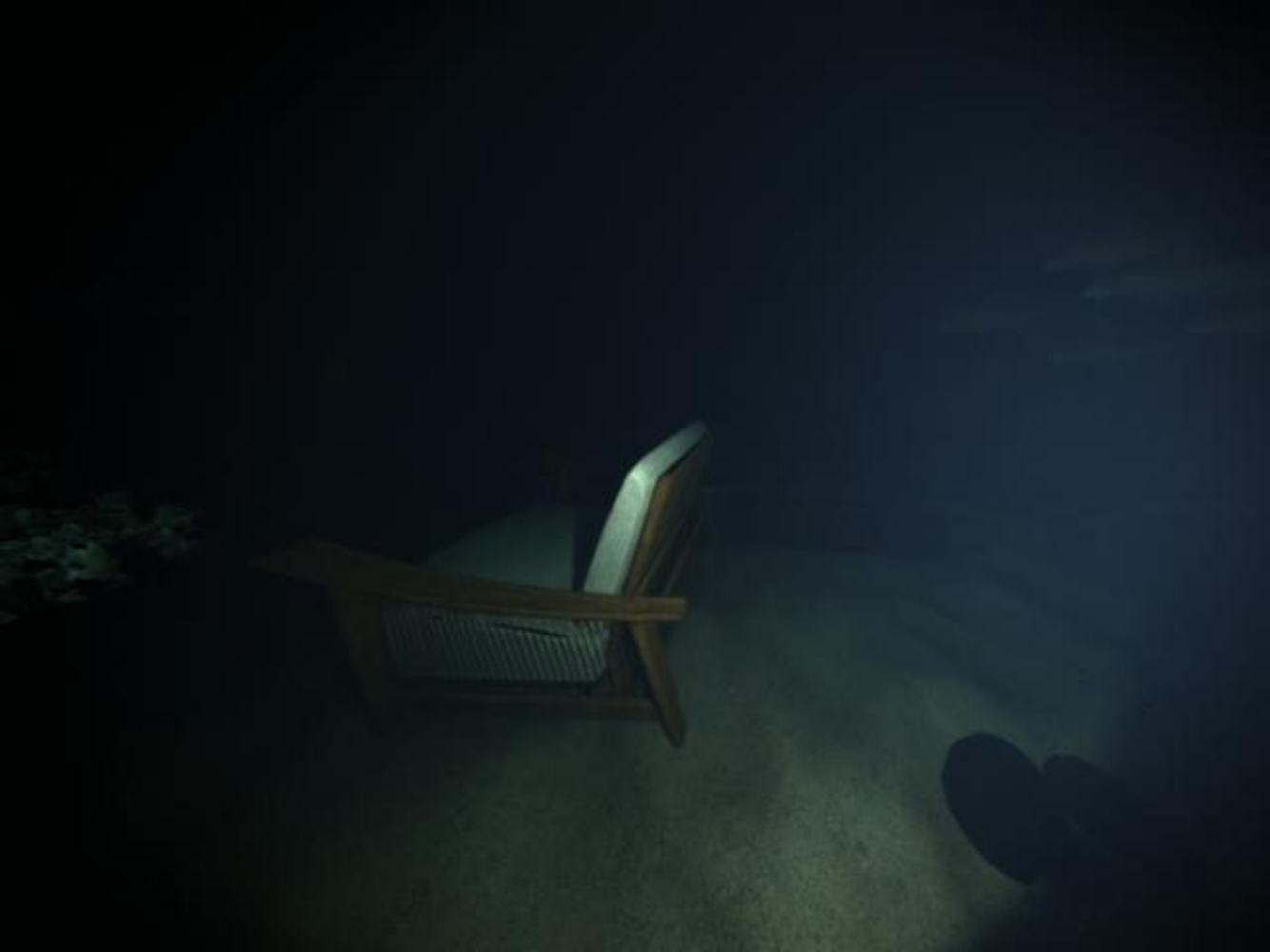} &
        \includegraphics[width=0.1\textwidth]{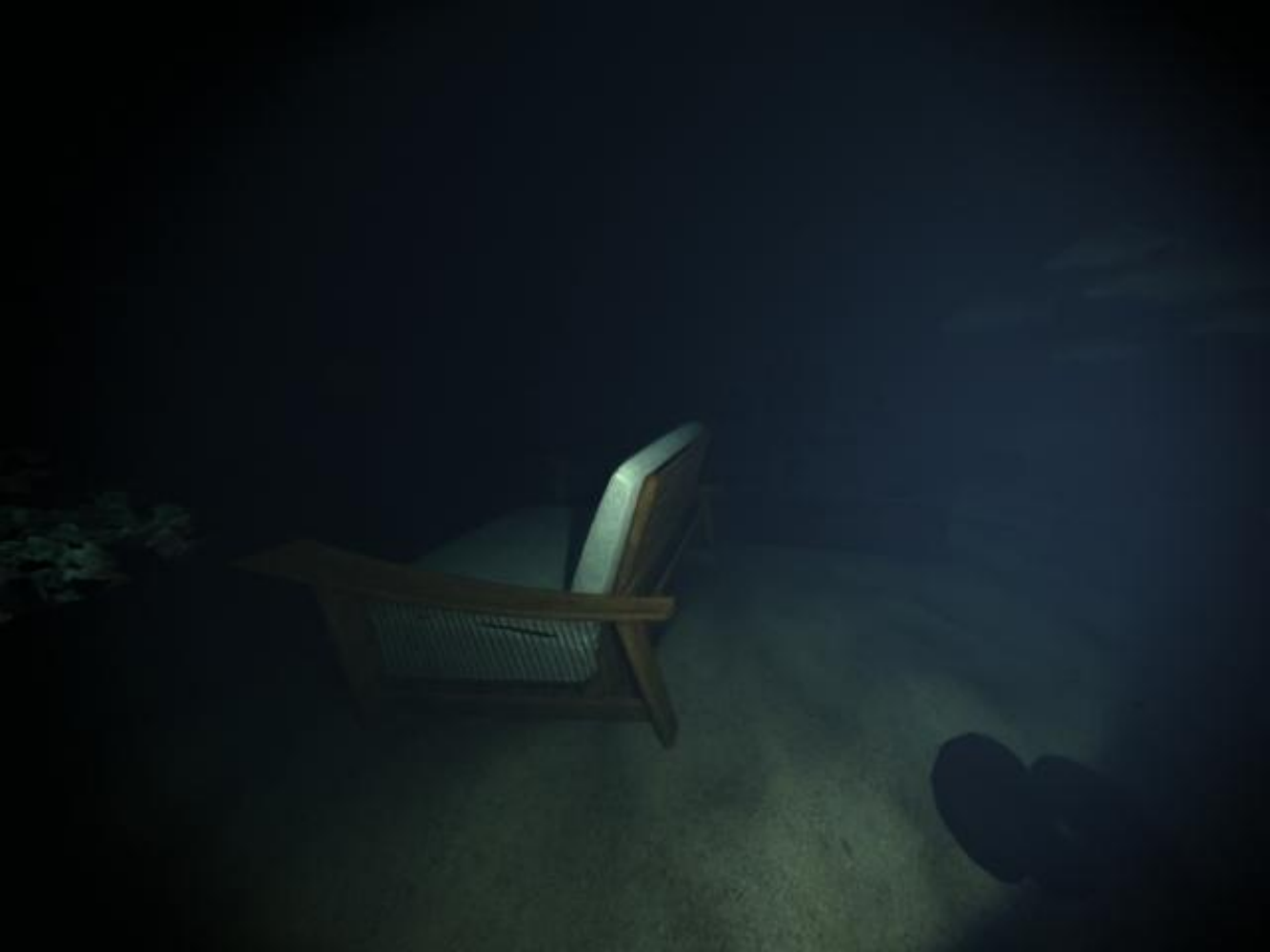} &
        \includegraphics[width=0.1\textwidth]{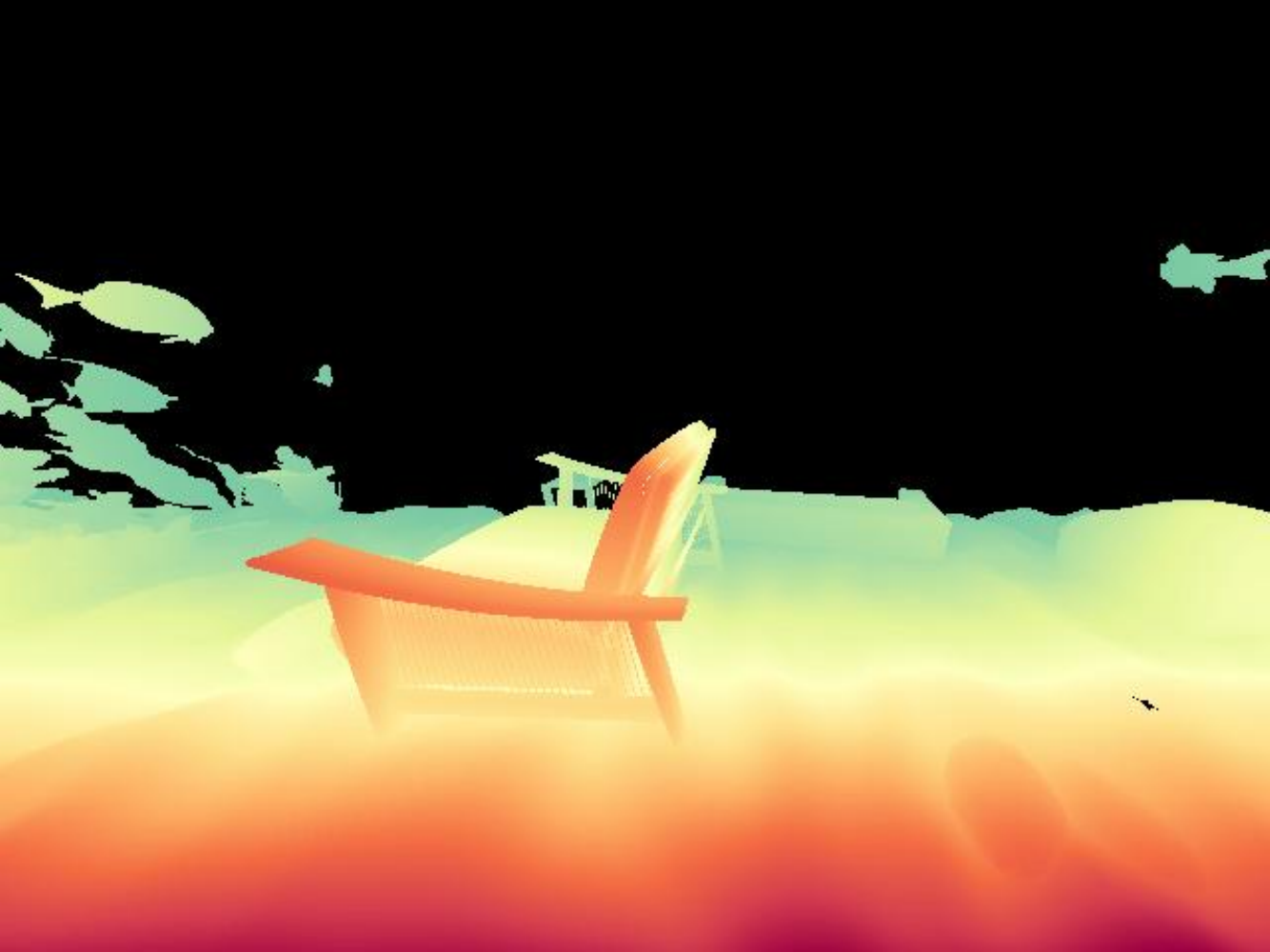} \\

        \includegraphics[width=0.1\textwidth]{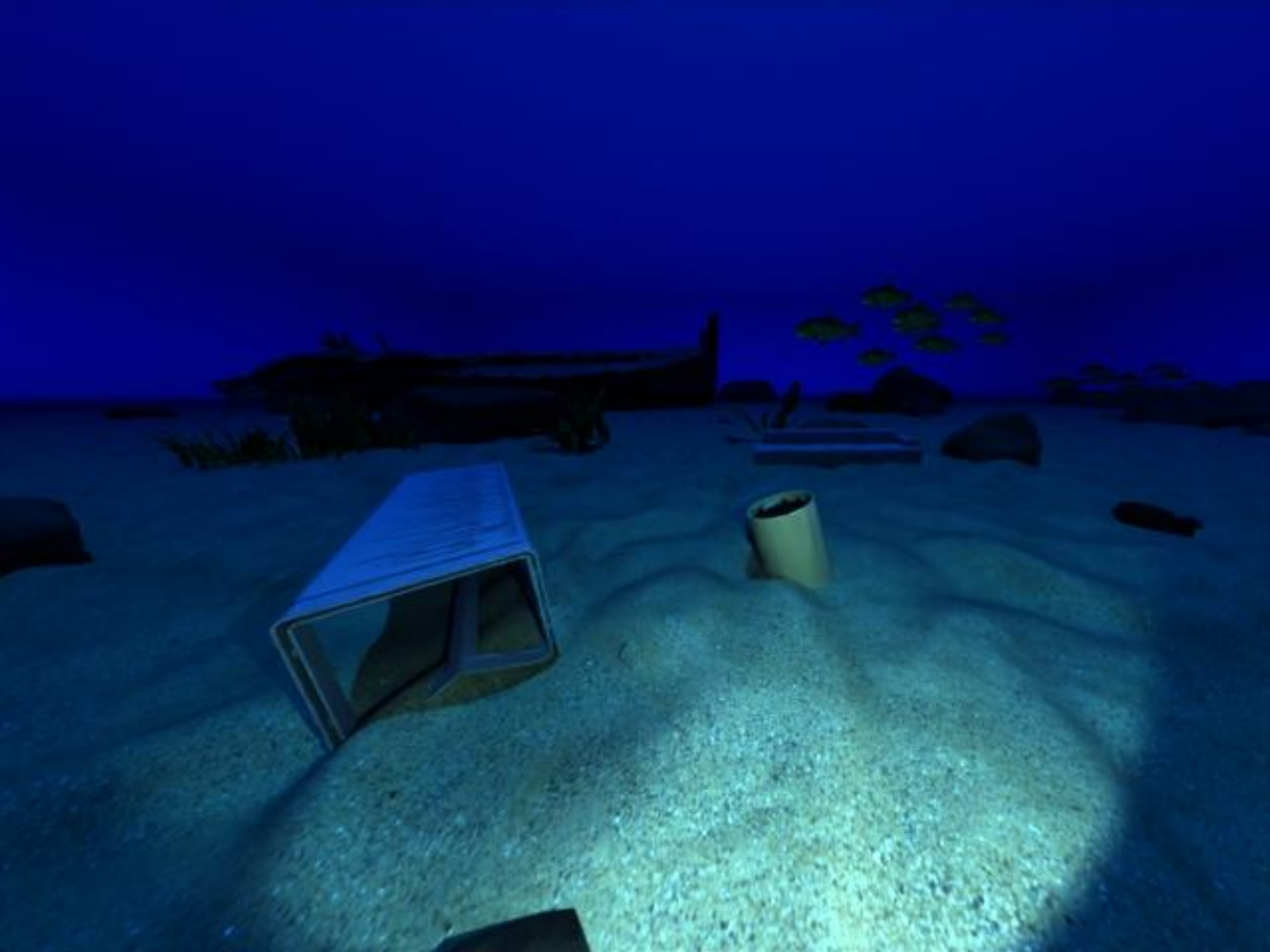} &
        \includegraphics[width=0.1\textwidth]{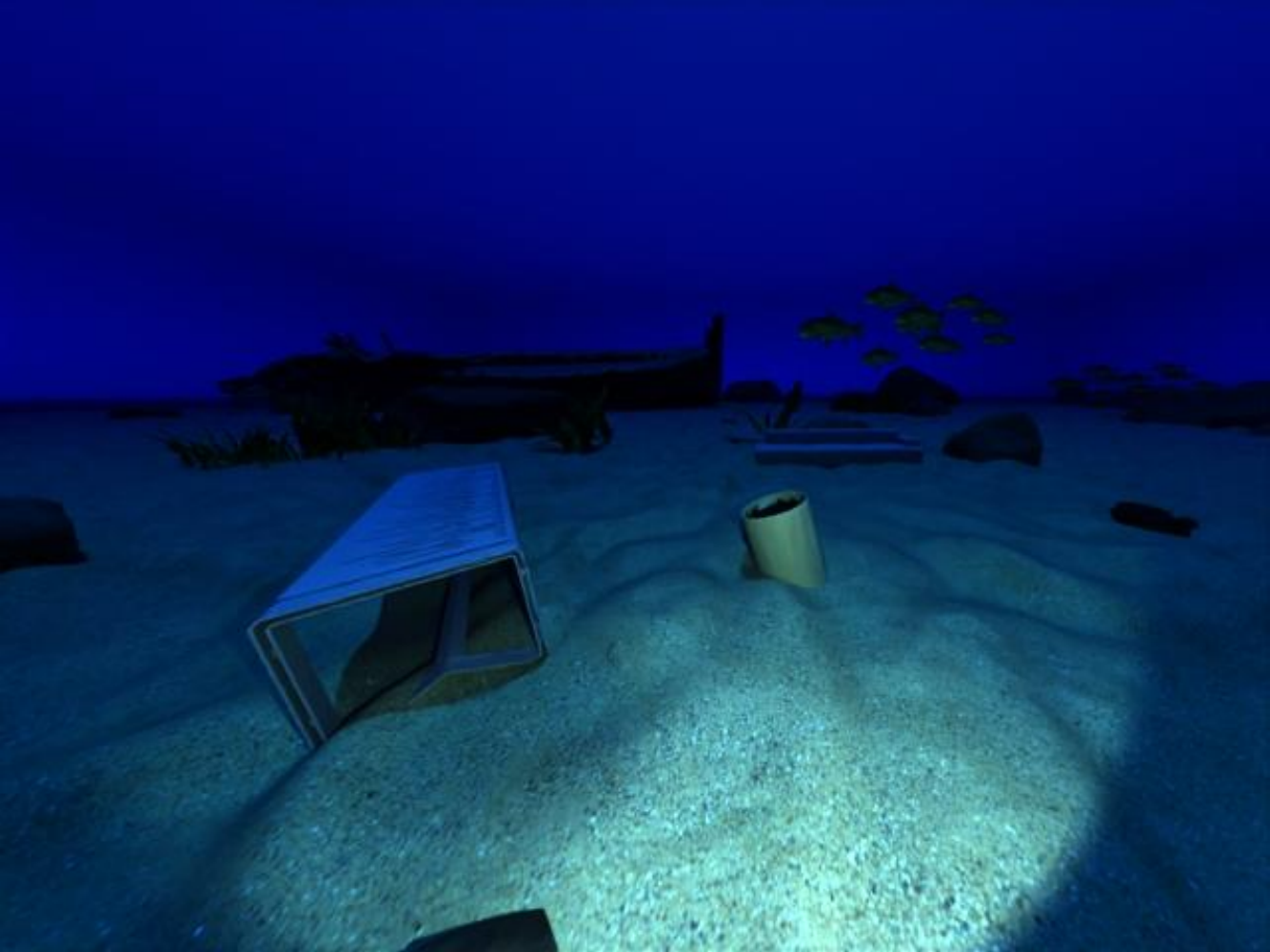} &
        \includegraphics[width=0.1\textwidth]{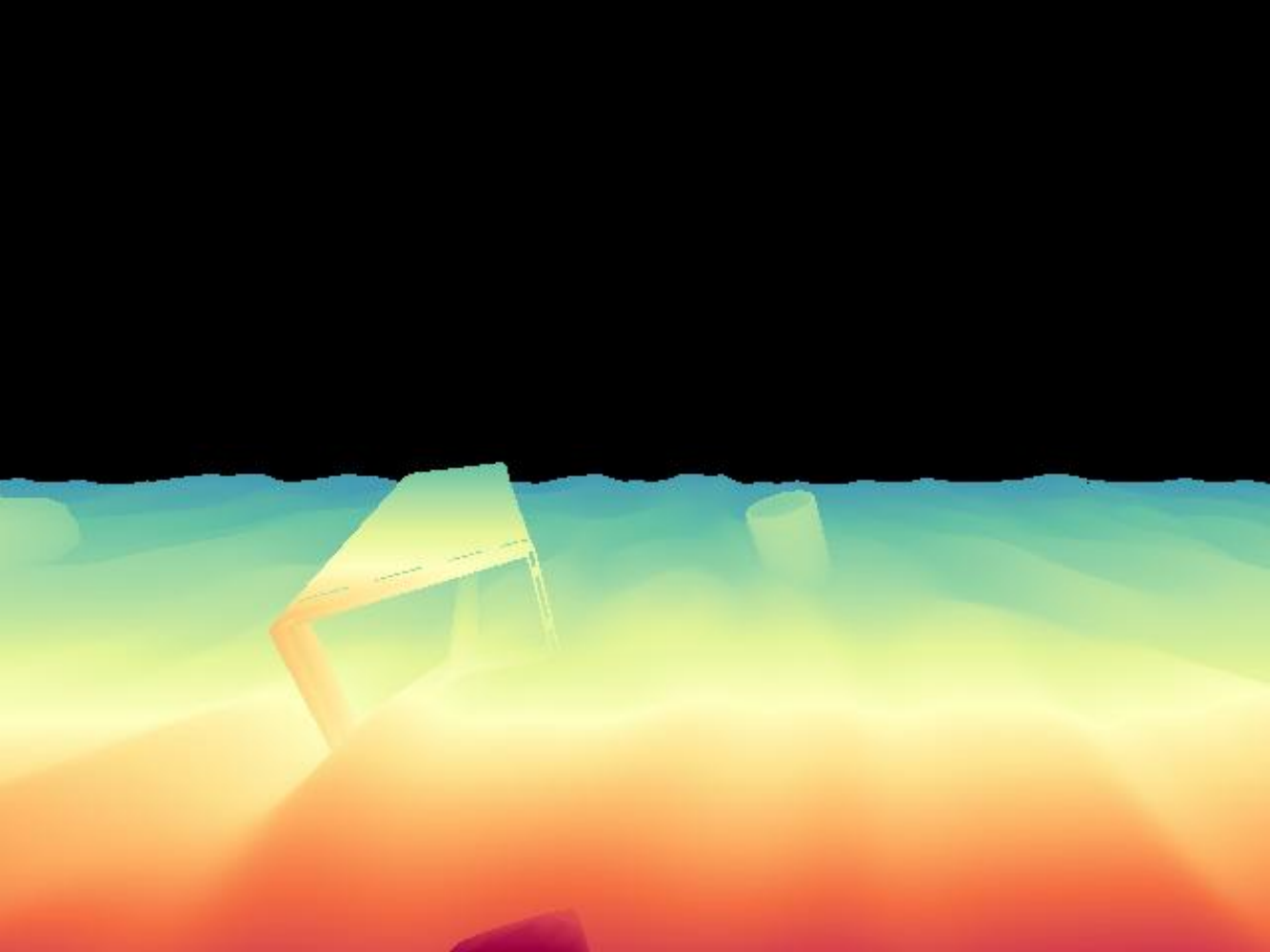} &

        \includegraphics[width=0.1\textwidth]{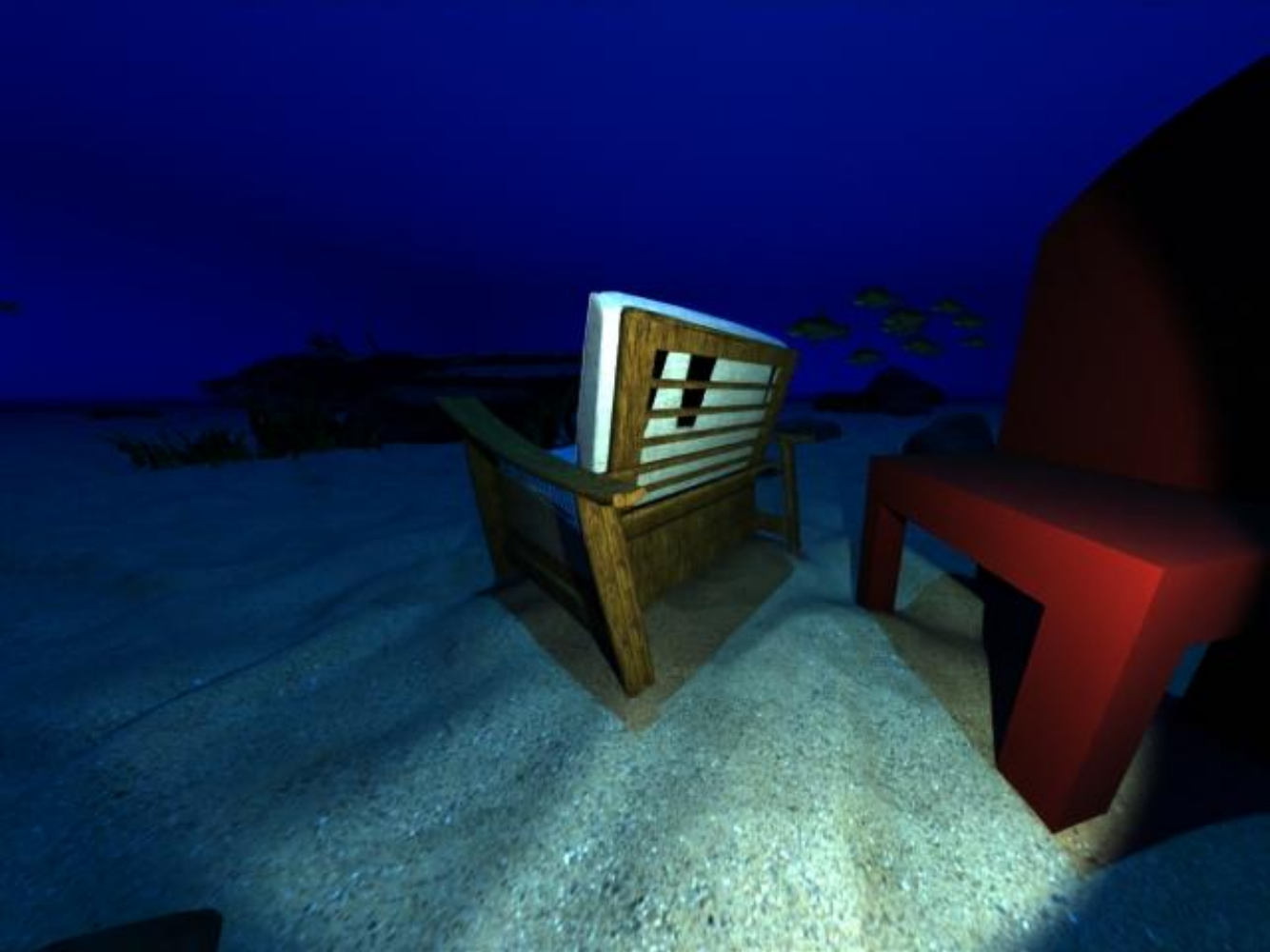} &
        \includegraphics[width=0.1\textwidth]{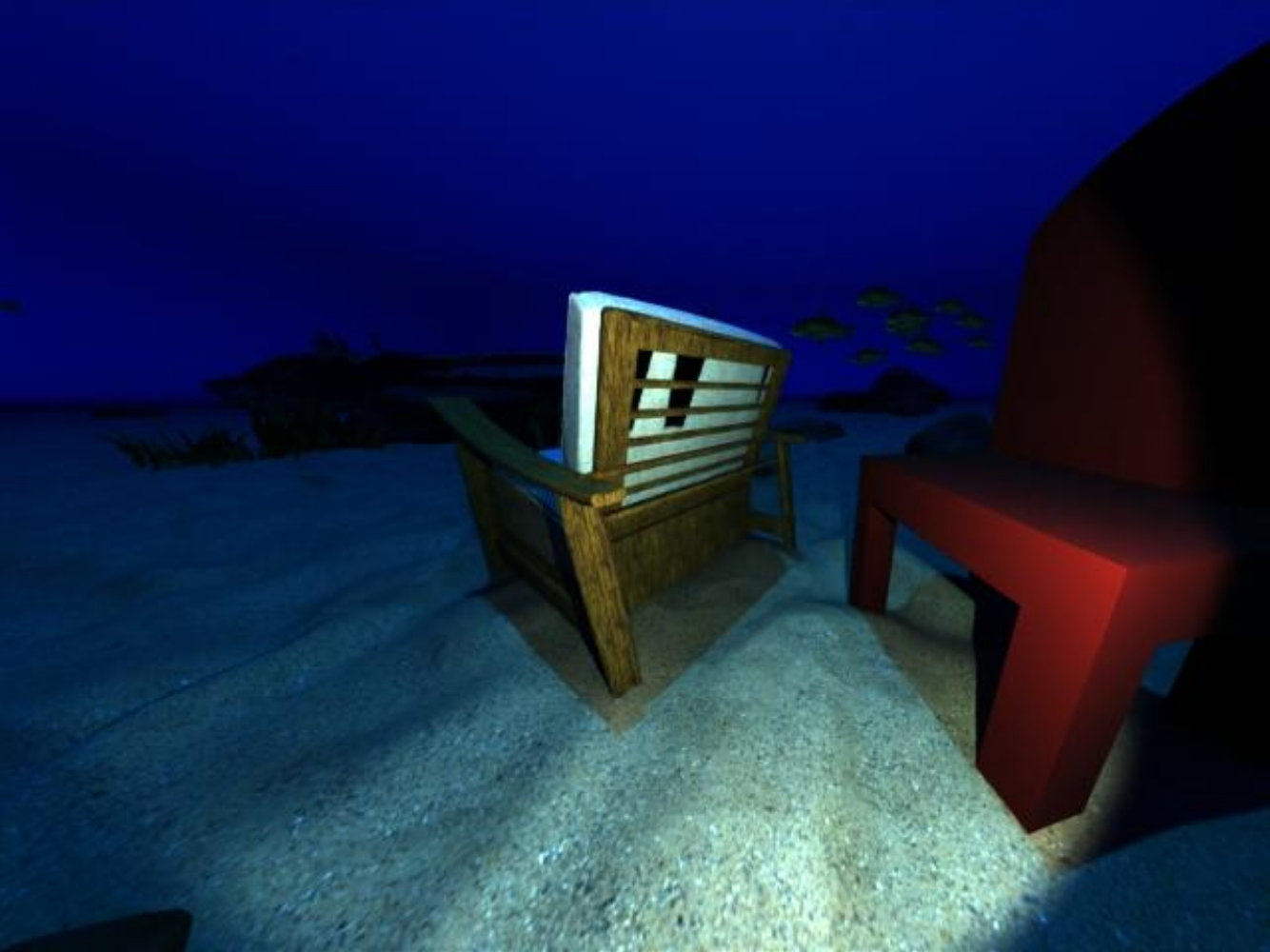} &
        \includegraphics[width=0.1\textwidth]{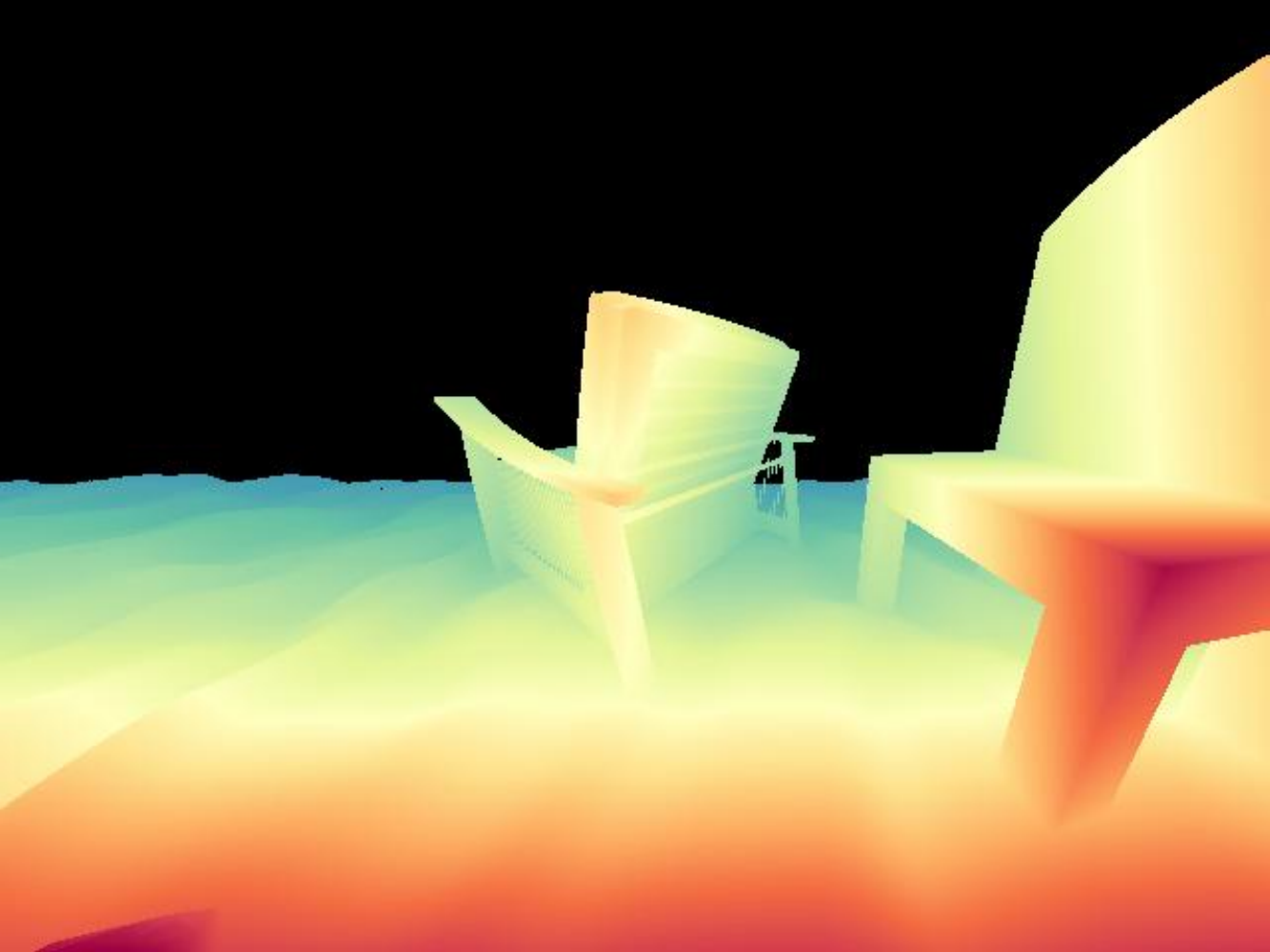} &

        \includegraphics[width=0.1\textwidth]{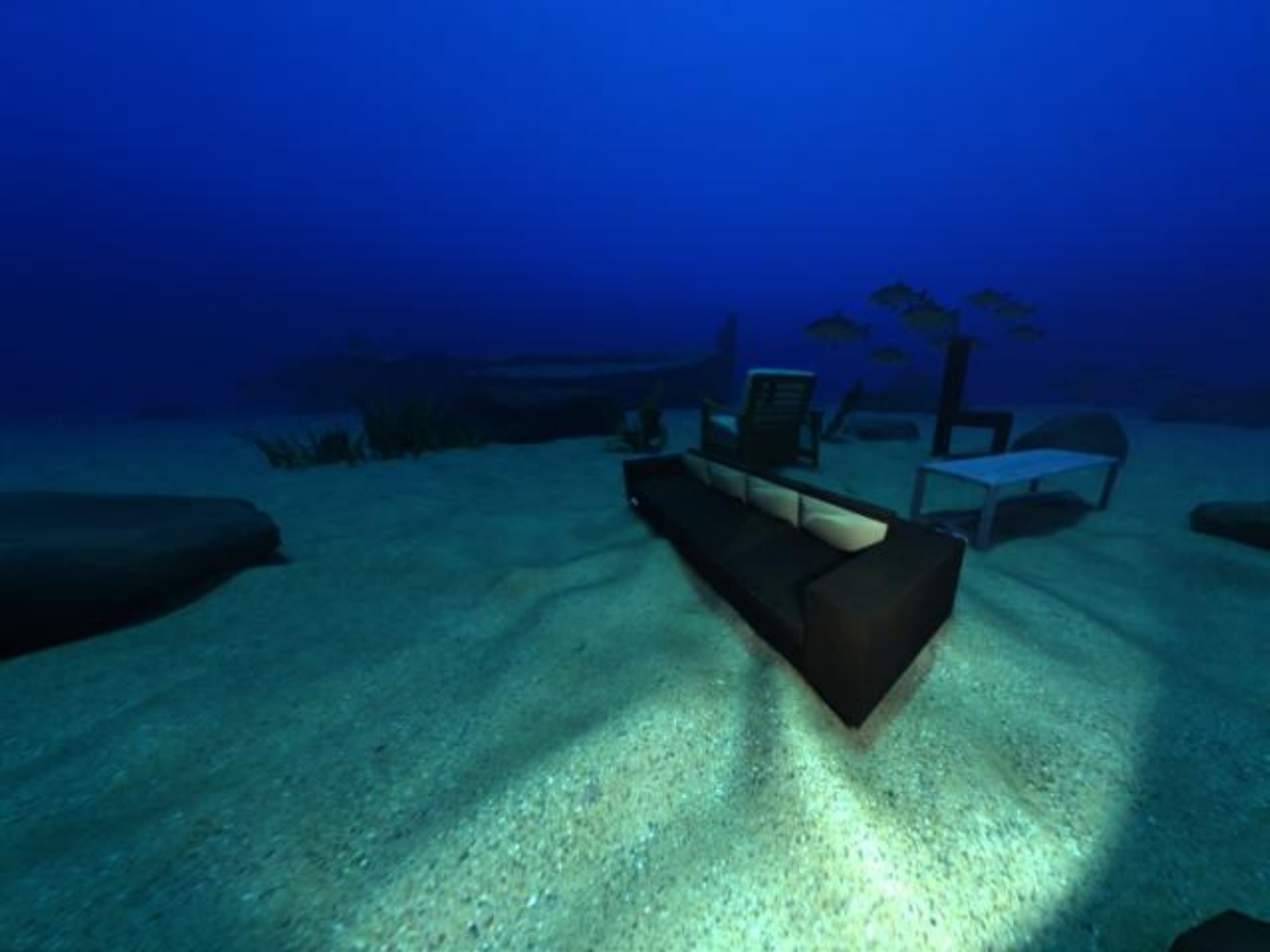} &
        \includegraphics[width=0.1\textwidth]{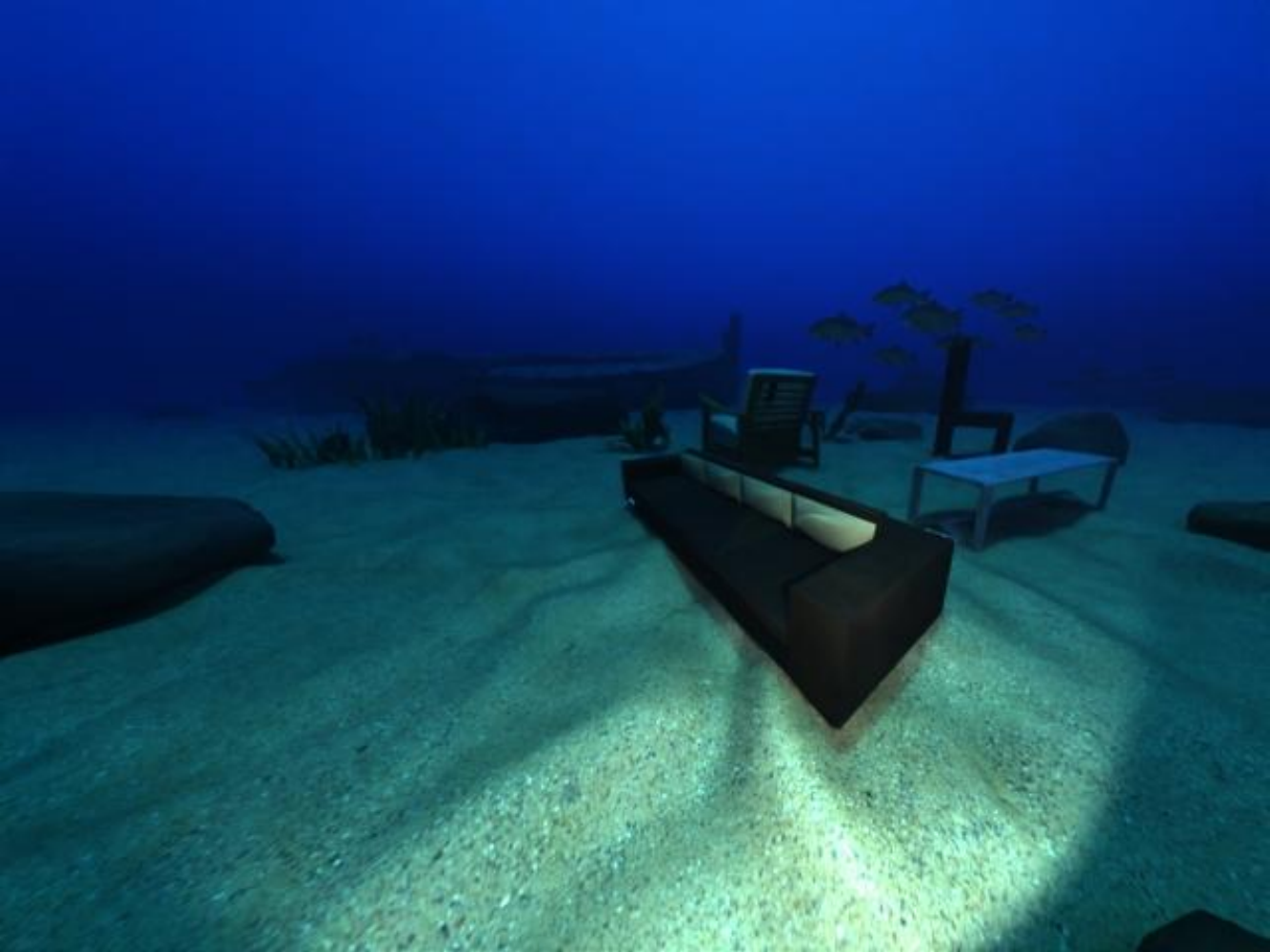} &
        \includegraphics[width=0.1\textwidth]{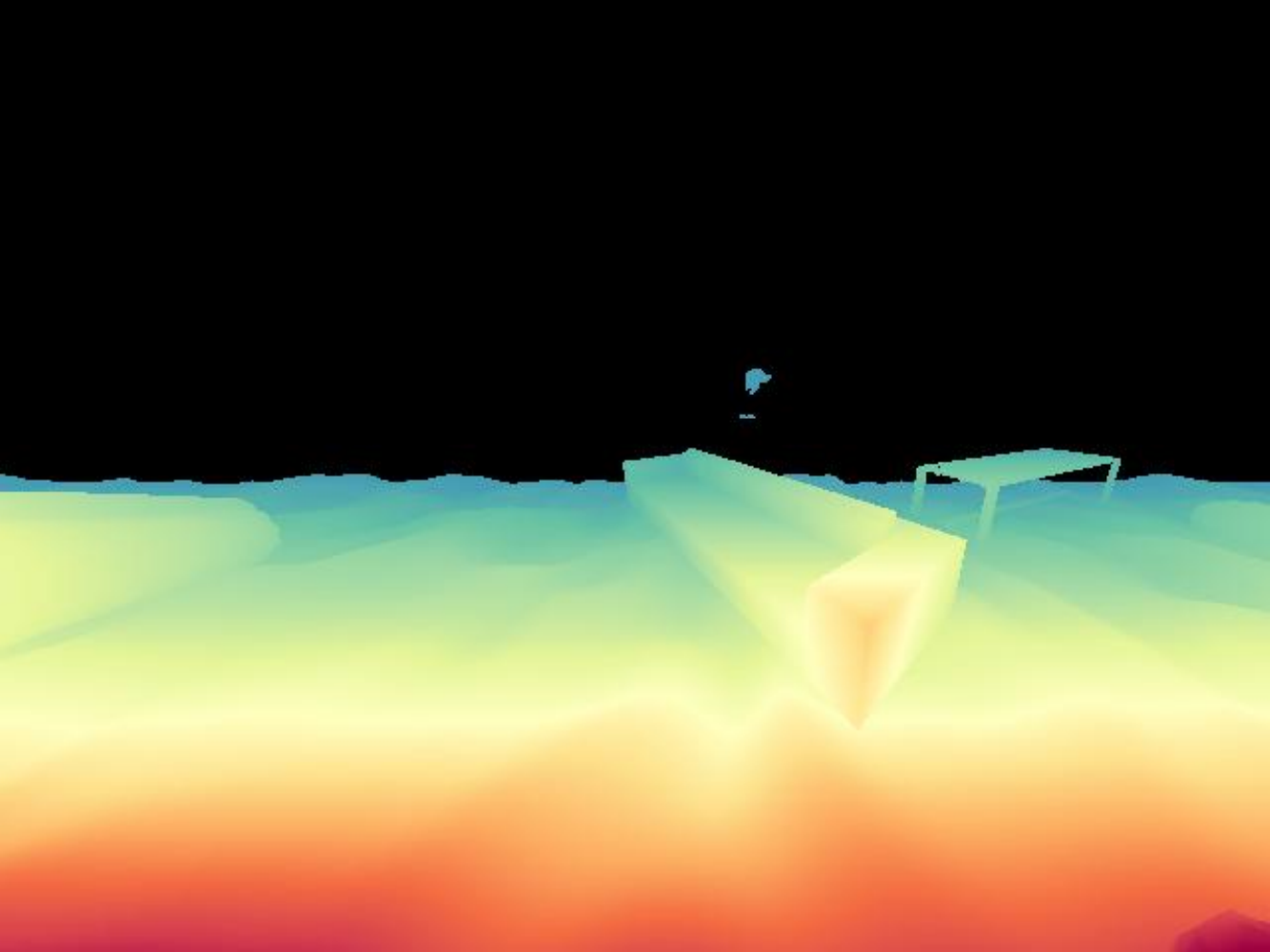} \\

    \end{tabular}}

    \caption{\textbf{Additional SeaStereo-Dataset samples.} Each row shows three sets of stereo RGB images (left, right) and their corresponding disparity annotations under different Jerlov water types, seafloor depths, and cameras. The dataset captures diverse underwater conditions ranging from clear to highly turbid water.}
    \label{fig:sea_synth_additional}
\end{figure*}

\paragraph{Additional Samples.}
\label{sec:appendix_data_synth_vis}
\autoref{fig:sea_synth_additional} presents additional samples from SeaStereo-Dataset, illustrating the diversity of water types, depth ranges, and object configurations. The dataset covers a wide spectrum of underwater visibility conditions, from clear open water (Jerlov~I) to highly turbid coastal water (Jerlov~3C), providing a comprehensive testbed for evaluating stereo methods under varying degrees of photometric degradation.

\section{Efficiency}
\label{sec:appendix_efficient_related}

Reducing inference cost while maintaining accuracy is a central challenge in stereo depth estimation. Predominant strategies simplify the architecture itself.
StereoNet~\cite{khamis2018stereonet}, BGNet~\cite{xu2021bgnet}, MobileStereoNet~\cite{shamsafar2022mobilestereonet}, and DeepPruner~\cite{duggal2019deeppruner} reduce volume resolution or prune the disparity search space;
AANet~\cite{xu2020aanet}, Fast-ACVNet~\cite{xu2024acvnet}, CoEx~\cite{bangunharcana2021coex}, CGI-Stereo~\cite{xu2023cgistereo}, ADStereo~\cite{wang2025adstereo}, and LightStereo~\cite{guo2025lightstereo} replace expensive 3D aggregation with lightweight 2D alternatives;
HITNet~\cite{tankovich2021hitnet} eliminates explicit volumes entirely via tile-based geometric propagation.
These methods achieve impressive speed (LightStereo-S runs at over 100\,FPS on a single GPU), though the architectural simplifications can limit accuracy in challenging regions such as textureless areas, fine structures, and non-Lambertian surfaces.
Our method pursues efficiency from a complementary direction: rather than simplifying the architecture, we reduce the number of refinement iterations required for convergence, maintaining rich contextual modeling while substantially lowering total inference time.

\paragraph{Efficient Methods: Underwater.}
Tab.~\ref{tab:realtime_underwater} compares LinStereo ($T\!=\!2$) against real-time and efficient stereo methods. With only two refinement iterations, LinStereo achieves AbsRel $= 0.09$, RMSE $= 3.84$ on TartanAir-UW, outperforming all efficient baselines running at full iterations. The gap is most pronounced on SQUID: all efficient baselines yield AbsRel $\geq 0.71$, with the best competing method (RAFT-Stereo$^\dagger$) reaching AbsRel $= 0.71$ and RT-IGEV achieving AbsRel $= 0.73$ despite additional training data, while LinStereo at $T\!=\!2$ achieves AbsRel $= 0.05$, RMSE $= 1.03$.

\begin{table*}[t]
\centering
\caption{
  Quantitative comparison of efficient stereo depth estimation on underwater scenes.
  TartanAir-UW: underwater images from TartanAir~\cite{wang2020tartanair}.
  SQUID: real-world underwater dataset~\cite{berman2020underwater}.
  Extra Data ($\checkmark$) indicates the method uses additional training data beyond SceneFlow~\cite{mayer2016large}.
}
\label{tab:realtime_underwater}
\resizebox{\textwidth}{!}{%
\begin{tabular}{l c ccccccc ccccccc}
\toprule
\multirow{2}{*}{Method}
  & \multirow{2}{*}{\begin{tabular}[c]{@{}c@{}}Extra\\Data\end{tabular}}
  & \multicolumn{7}{c}{TartanAir UW}
  & \multicolumn{7}{c}{SQUID} \\
\cmidrule(lr){3-9} \cmidrule(lr){10-16}
  &
  & AbsRel$\downarrow$ & SqRel$\downarrow$ & RMSE$\downarrow$ & LogRMSE$\downarrow$
  & A1$\uparrow$ & A2$\uparrow$ & A3$\uparrow$
  & AbsRel$\downarrow$ & SqRel$\downarrow$ & RMSE$\downarrow$ & LogRMSE$\downarrow$
  & A1$\uparrow$ & A2$\uparrow$ & A3$\uparrow$ \\
\midrule
LightStereo-S~\cite{guo2025lightstereo}
  & ---
  & 0.61 & 8.71 & 11.45 & 1.39 & 0.304 & 0.483 & 0.578
  & 0.91 & 6.00 & 4.17 & 0.65 & 0.452 & 0.572 & 0.674 \\
AANet~\cite{xu2020aanet}
  & ---
  & 0.24 & 2.74 & 5.08 & 0.30 & 0.815 & 0.910 & 0.949
  & 3.55 & 57.88 & 13.34 & 1.34 & 0.048 & 0.114 & 0.206 \\
MobileStereoNet~\cite{shamsafar2022mobilestereonet}
  & ---
  & 0.11 & 0.96 & 3.92 & 0.18 & 0.894 & 0.961 & 0.982
  & 0.88 & 4.32 & 3.35 & 0.60 & 0.416 & 0.574 & 0.701 \\
CoEx~\cite{bangunharcana2021coex}
  & ---
  & 0.13 & 2.04 & 4.84 & 0.22 & 0.890 & 0.951 & 0.972
  & 0.79 & 5.09 & 3.54 & 0.57 & 0.513 & 0.642 & 0.735 \\
CGI-Stereo~\cite{xu2023cgistereo}
  & ---
  & 0.12 & 1.99 & 4.94 & 0.22 & 0.895 & 0.953 & 0.975
  & 0.90 & 6.13 & 3.86 & 0.62 & 0.550 & 0.635 & 0.701 \\
Fast-ACVNet~\cite{xu2024acvnet}
  & ---
  & 0.12 & 2.03 & 5.29 & 0.27 & 0.892 & 0.945 & 0.967
  & 0.82 & 7.24 & 3.84 & 0.57 & 0.557 & 0.684 & 0.770 \\
ADStereo~\cite{wang2025adstereo}
  & ---
  & 0.12 & 1.11 & 4.58 & 0.22 & 0.863 & 0.946 & 0.975
  & 0.94 & 8.33 & 4.46 & 0.64 & 0.500 & 0.643 & 0.733 \\
RAFT-Stereo$^\dagger$~\cite{lipson2021raftstereo}
  & ---
  & 0.14 & 2.29 & 5.75 & 0.23 & 0.867 & 0.934 & 0.963
  & 0.71 & 10.04 & 3.57 & 0.45 & 0.771 & 0.841 & 0.878 \\
RT-IGEV~\cite{xu2025igevpp}
  & $\checkmark$
  & 0.17 & 4.27 & 6.89 & 0.27 & 0.873 & 0.926 & 0.953
  & 0.73 & 6.21 & 3.89 & 0.55 & 0.587 & 0.696 & 0.779 \\
\midrule
\rowcolor[rgb]{.886,.937,.855}
\textbf{Ours}
  & ---
  & \textbf{0.09} & \textbf{0.80} & \textbf{3.84} & \textbf{0.16}
  & \textbf{0.910} & \textbf{0.967} & \textbf{0.986}
  & \textbf{0.05} & \textbf{0.16} & \textbf{1.03} & \textbf{0.09}
  & \textbf{0.965} & \textbf{0.985} & \textbf{0.995} \\
\bottomrule
\end{tabular}}
\end{table*}

\section{Real-World Evaluation: Laboratory Tank}
\label{sec:appendix_rope}

LinStereo is evaluated on a controlled laboratory water tank dataset at close range ($<2$\,m).

\paragraph{Target Design.}
The evaluation target consists of pre-fabricated structural pieces whose geometry is known from precision CAD models created in SolidWorks.
Taut ropes are strung between these pieces, producing thin linear structures (${\sim}3$\,mm diameter) that serve as a stress test for fine-structure depth recovery, an especially challenging scenario for stereo methods, as conventional cost volumes struggle to maintain reliable correspondences on structures narrower than a few pixels.

\paragraph{Data Acquisition.}
The target assembly is fully submerged in a glass water tank and imaged with a calibrated stereo camera.
AprilTag (16h5) fiducial markers are affixed at known positions on the target frame.
For each stereo pair, the camera-to-target pose is recovered by detecting the tags in the left image and solving the Perspective-\textit{n}-Point (PnP) problem using the known tag geometry and the calibrated camera intrinsics.

\paragraph{Ground-Truth Depth Generation.}
Given the recovered camera pose, the high-fidelity CAD mesh of the target (including the ropes) is rendered from the left camera viewpoint using standard z-buffer rasterization.
Pixels that do not project onto any mesh surface are masked as invalid.
This \emph{render-from-CAD} pipeline yields sub-millimetre depth accuracy for the rigid prefabricated components, providing a ground-truth quality far higher than what photogrammetric reconstruction could achieve on the thin, textureless rope segments.
The resulting benchmark complements the moderate-to-large depth scales covered by TartanAir-UW and SQUID with high-precision near-range ground truth under genuine underwater conditions.

\section{Full-Precision Underwater Results}
\label{sec:appendix_underwater_full}

Tab.~\ref{tab:quant_underwater_full} provides full-precision accuracy thresholds for the underwater evaluation in the main paper. Tab.~\ref{tab:quant_labtank_full} reports the laboratory tank results.

\begin{table*}[t]
\centering
\caption{
  Full-precision underwater evaluation results (three decimal places for accuracy thresholds A1--A3).
  This is the unrounded version of the underwater comparison table in the main paper.
}
\label{tab:quant_underwater_full}
\resizebox{\textwidth}{!}{%
\begin{tabular}{l c ccccccc ccccccc}
\toprule
\multirow{2}{*}{Method}
  & \multirow{2}{*}{\begin{tabular}[c]{@{}c@{}}Extra\\Data\end{tabular}}
  & \multicolumn{7}{c}{TartanAir UW}
  & \multicolumn{7}{c}{SQUID} \\
\cmidrule(lr){3-9} \cmidrule(lr){10-16}
  &
  & AbsRel$\downarrow$ & SqRel$\downarrow$ & RMSE$\downarrow$ & LogRMSE$\downarrow$
  & A1$\uparrow$ & A2$\uparrow$ & A3$\uparrow$
  & AbsRel$\downarrow$ & SqRel$\downarrow$ & RMSE$\downarrow$ & LogRMSE$\downarrow$
  & A1$\uparrow$ & A2$\uparrow$ & A3$\uparrow$ \\
\midrule
UniMatch~\cite{xu2023unifying}
  & ---
  & 0.166 & 2.586 & 6.346 & 0.258 & 0.759 & 0.830 & 0.837
  & 1.382 & 20.905 & 6.416 & 0.662 & 0.561 & 0.660 & 0.734 \\
MoCha-Stereo~\cite{chen2024mocha}
  & ---
  & 0.155 & 2.303 & 6.023 & 0.249 & 0.769 & 0.835 & 0.843
  & 0.150 & 0.610 & 1.550 & 0.200 & 0.875 & 0.931 & 0.958 \\
NMRF~\cite{guan2024nmrf}
  & ---
  & 0.106 & 1.084 & 4.633 & 0.212 & 0.811 & 0.856 & 0.870
  & 0.430 & 5.270 & 2.570 & 0.330 & 0.844 & 0.904 & 0.935 \\
PSMNet~\cite{chang2018pyramid}
  & ---
  & 0.103 & 1.018 & 4.558 & 0.210 & 0.813 & 0.857 & 0.871
  & 0.457 & 4.311 & 3.428 & 0.510 & 0.736 & 0.813 & 0.853 \\
RAFT-Stereo~\cite{lipson2021raftstereo}
  & ---
  & 0.083 & 0.647 & 4.361 & 0.196 & 0.902 & 0.963 & 0.984
  & 0.068 & 0.268 & 1.245 & 0.123 & 0.937 & 0.971 & 0.987 \\
MGStereo~\cite{yao2025diving}
  & ---
  & 0.075 & 0.552 & 3.685 & 0.185 & 0.911 & 0.968 & 0.987
  & 0.093 & 0.708 & 1.992 & 0.161 & 0.925 & 0.958 & 0.976 \\
Stereo~Anywhere~\cite{bartolomei2025stereo}
  & ---
  & 0.058 & 0.414 & 3.241 & 0.177 & 0.946 & 0.980 & 0.989
  & 0.072 & 0.401 & 1.461 & 0.132 & 0.937 & 0.976 & 0.985 \\
DEFOM-Stereo~\cite{jiang2025defom}
  & ---
  & 0.054 & 0.406 & 3.127 & 0.170 & 0.953 & 0.982 & 0.990
  & 0.090 & 0.630 & 2.000 & 0.160 & 0.915 & 0.954 & 0.977 \\
\midrule
Selective-Stereo~(RAFT)~\cite{wang2024selective}
  & $\checkmark$
  & 0.092 & 0.670 & 4.312 & 0.214 & 0.902 & 0.962 & 0.981
  & 0.106 & 0.219 & 1.158 & 0.165 & 0.877 & 0.940 & 0.968 \\
Selective-Stereo~(IGEV)~\cite{wang2024selective}
  & $\checkmark$
  & 0.114 & 1.022 & 5.007 & 0.235 & 0.856 & 0.942 & 0.976
  & 0.077 & 0.206 & 1.048 & 0.137 & 0.932 & 0.966 & 0.980 \\
IGEV-Stereo~\cite{xu2023igev}
  & $\checkmark$
  & 0.100 & 0.900 & 4.679 & 0.212 & 0.891 & 0.955 & 0.979
  & 0.198 & 1.346 & 2.679 & 0.455 & 0.760 & 0.821 & 0.863 \\
IGEV++~\cite{xu2025igevpp}
  & $\checkmark$
  & 0.089 & 0.815 & 4.375 & 0.201 & 0.905 & 0.962 & 0.983
  & 0.061 & 0.219 & 1.107 & 0.120 & 0.950 & 0.981 & 0.990 \\
FoundationStereo~\cite{wen2025foundationstereo}
  & $\checkmark$
  & 0.050 & 0.398 & 3.013 & 0.164 & \textbf{0.959} & 0.984 & 0.991
  & 0.066 & 0.304 & 1.364 & 0.128 & 0.940 & 0.976 & 0.987 \\
\midrule
\rowcolor[rgb]{.886,.937,.855}
\textbf{Ours} ($T\!=\!8$)
  & ---
  & \textbf{0.036} & \textbf{0.194} & \textbf{2.079} & \textbf{0.087}
  & \textbf{0.959} & \textbf{0.992} & \textbf{0.998}
  & \textbf{0.045} & \textbf{0.121} & \textbf{0.896} & \textbf{0.084}
  & \textbf{0.970} & \textbf{0.990} & \textbf{0.996} \\
\bottomrule
\end{tabular}}
\end{table*}

\begin{table}[t]
\centering
\caption{
  Full-precision evaluation on close-range laboratory tank images ($<2$\,m depth).
  This is the unrounded version of the laboratory-tank table in the main paper.
  Extra Data ($\checkmark$) indicates training data beyond SceneFlow~\cite{mayer2016large}.
  Best results in \textbf{bold}.
}
\label{tab:quant_labtank_full}
\resizebox{\textwidth}{!}{%
\begin{tabular}{l c c c c c c c c}
\toprule
\textbf{Method}
& \begin{tabular}[c]{@{}c@{}}Extra\\Data\end{tabular}
& \textbf{AbsRel$\downarrow$}
& \textbf{SqRel$\downarrow$}
& \textbf{RMSE$\downarrow$}
& \textbf{Log RMSE$\downarrow$}
& \textbf{A1$\uparrow$}
& \textbf{A2$\uparrow$}
& \textbf{A3$\uparrow$} \\
\midrule
PSMNet~\cite{chang2018pyramid}                      & --- & 0.1805 & 0.1595 & 0.3380 & 0.2841 & 0.8677 & 0.9025 & 0.9300 \\
UniMatch~\cite{xu2023unifying}                      & --- & 0.0815 & 0.0392 & 0.1700 & 0.1594 & 0.9334 & 0.9595 & 0.9803 \\
Stereo~Anywhere~\cite{bartolomei2025stereo}         & --- & 0.0868 & 0.0555 & 0.2009 & 0.1865 & 0.9216 & 0.9347 & 0.9706 \\
RAFT-Stereo~\cite{lipson2021raftstereo}             & --- & 0.0603 & 0.0297 & 0.1501 & 0.1391 & 0.9442 & 0.9585 & 0.9891 \\
MGStereo~\cite{yao2025diving}                     & --- & 0.0816 & 0.0487 & 0.1788 & 0.1661 & 0.9250 & 0.9423 & 0.9777 \\
\midrule
Selective-Stereo~(RAFT)~\cite{wang2024selective}     & $\checkmark$ & 0.0653 & 0.0262 & 0.1401 & 0.1366 & 0.9358 & 0.9673 & 0.9903 \\
Selective-Stereo~(IGEV)~\cite{wang2024selective}     & $\checkmark$ & 0.0819 & 0.0360 & 0.1603 & 0.1604 & 0.9111 & 0.9529 & 0.9867 \\
IGEV-Stereo~\cite{xu2023igev}                       & $\checkmark$ & 0.0577 & 0.0212 & 0.1289 & 0.1246 & 0.9398 & 0.9663 & 0.9949 \\
IGEV++~\cite{xu2025igevpp}                          & $\checkmark$ & 0.0470 & 0.0198 & 0.1241 & 0.1150 & 0.9599 & 0.9723 & 0.9935 \\
FoundationStereo~\cite{wen2025foundationstereo}      & $\checkmark$ & 0.0743 & 0.0441 & 0.1810 & 0.1665 & 0.9315 & 0.9445 & 0.9812 \\
\midrule
\rowcolor[rgb]{.886, .937, .855}
\textbf{Ours} ($T\!=\!8$)                           & --- & \textbf{0.0423} & \textbf{0.0067} & \textbf{0.0672} & \textbf{0.0727} & \textbf{0.9778} & \textbf{0.9943} & \textbf{0.9984} \\
\bottomrule
\end{tabular}}
\end{table}

\section{Additional Qualitative Results}
\label{sec:appendix_qualitative}

We provide additional qualitative comparisons beyond those shown in the main paper. These visualizations further illustrate LinStereo's advantages: PALA's global context propagation recovers accurate depth in regions where local updaters fail, while HSCV's multi-scale correlation preserves fine structural detail.

\begin{figure*}[t]
\centering
\includegraphics[width=0.8\textwidth]{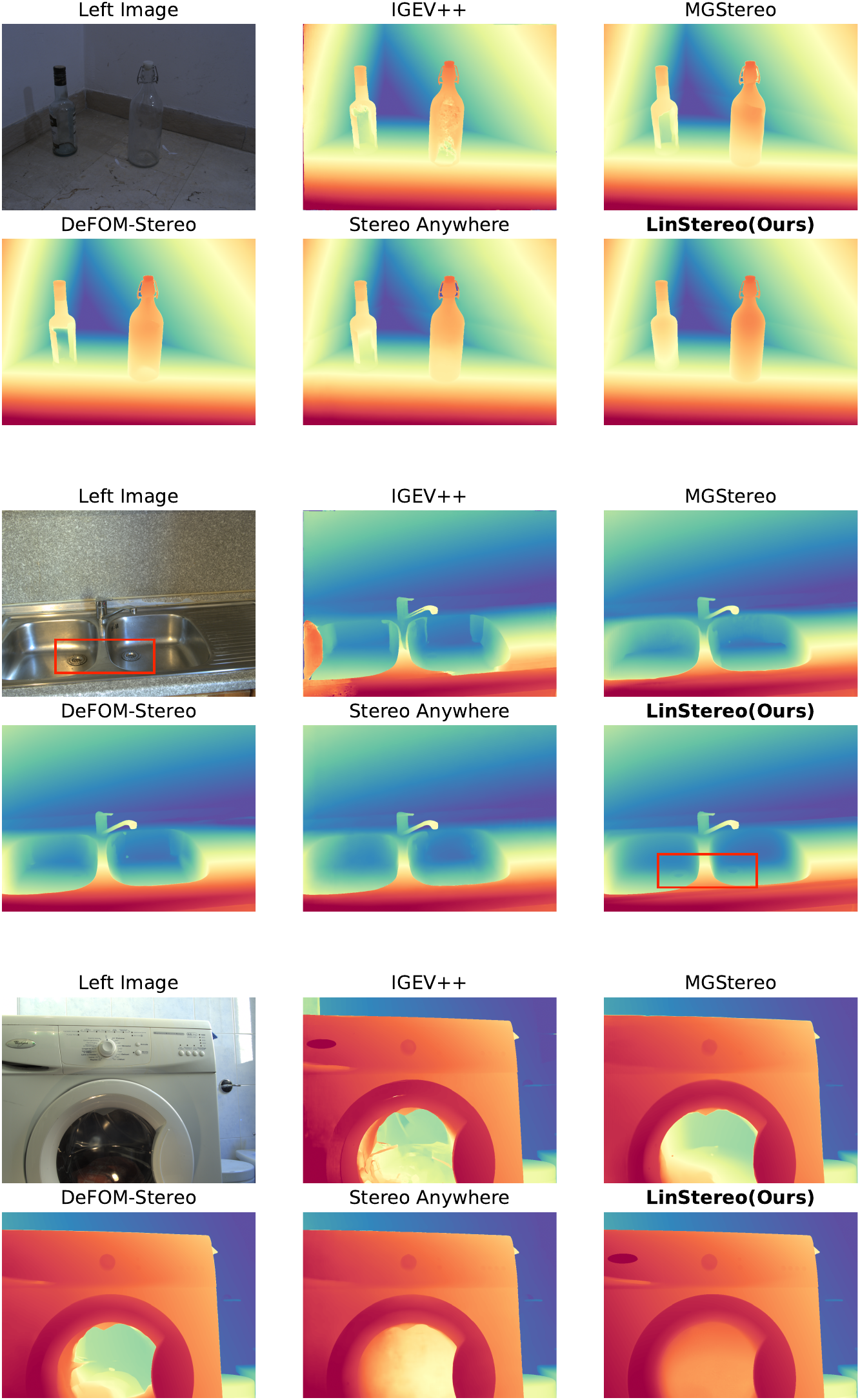}
\caption{\textbf{Additional qualitative comparison on Booster~(Q).} Scenes contain transparent glass bottles, specular metallic sinks, and reflective appliance surfaces. Note that for transparent objects, the expected depth typically corresponds to the physical surface (\eg, the glass pane of the washing machine door) rather than the depth of objects visible through it. Competing methods produce depth bleeding around glass boundaries and noisy estimates on specular metal, while LinStereo recovers clean object contours and consistent surface depth on these non-Lambertian materials (red boxes).}
\label{fig:qual_booster}
\end{figure*}

\begin{figure*}[t]
\centering
\includegraphics[width=0.98\textwidth]{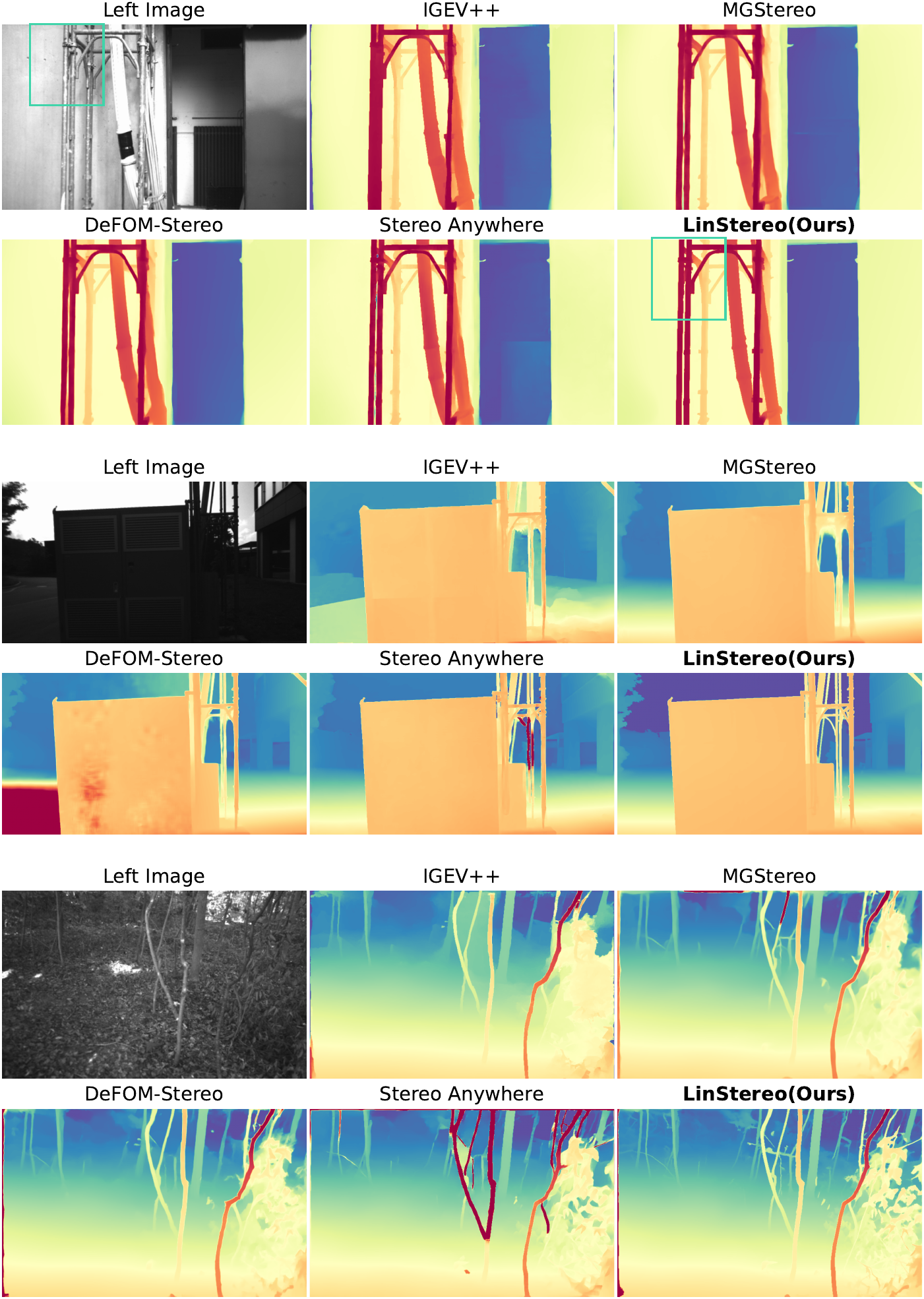}
\caption{\textbf{Additional qualitative comparison on ETH3D.} Scenes span industrial interiors with thin pipes, low-light building exteriors, and dense outdoor vegetation. Baselines lose thin-structure continuity and produce large-area depth errors in dark regions, while LinStereo preserves fine geometric detail and remains stable under low-light conditions (cyan boxes).}
\label{fig:qual_eth3d}
\end{figure*}

\begin{figure*}[t]
\centering
\includegraphics[width=0.98\textwidth]{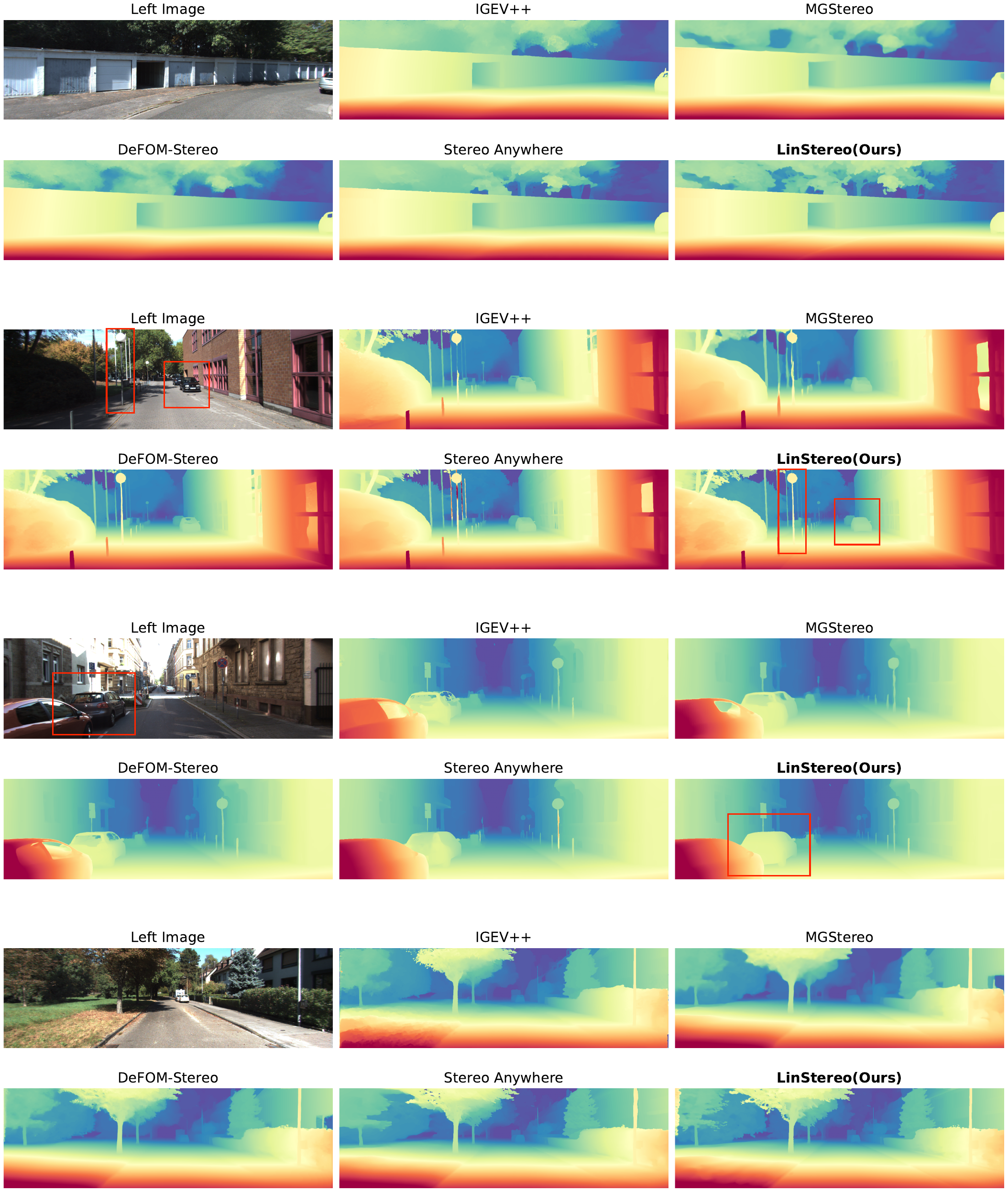}
\caption{\textbf{Additional qualitative comparison on KITTI~2012.} LinStereo produces sharper vehicle boundaries and preserves thin structures (\eg, poles, traffic signs) that competing methods blur or fragment (red boxes).}
\label{fig:qual_kitti2012}
\end{figure*}

\begin{figure*}[t]
\centering
\includegraphics[width=0.98\textwidth]{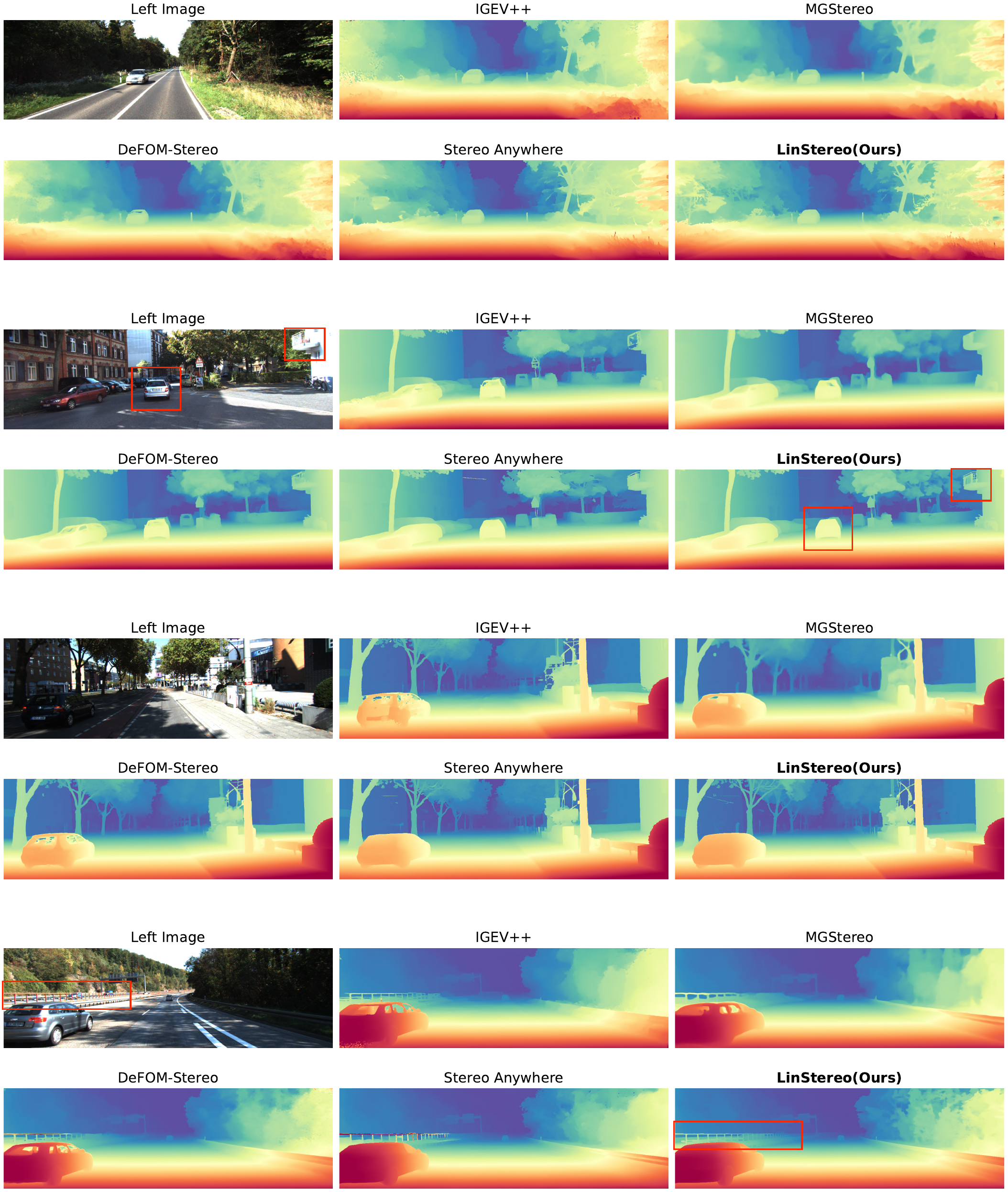}
\caption{\textbf{Additional qualitative comparison on KITTI~2015.} LinStereo resolves small distant objects and road-level details (\eg, lane markings, guard rails) that baselines over-smooth or lose entirely (red boxes).}
\label{fig:qual_kitti2015}
\end{figure*}

\paragraph{Standard Benchmarks.}
Figs.~\ref{fig:qual_booster}--\ref{fig:qual_kitti2015} present additional qualitative comparisons on Booster~(Q), ETH3D, KITTI~2012, and KITTI~2015. LinStereo's advantages are consistently visible across these benchmarks: (1)~on non-Lambertian surfaces (transparent glass, specular metal, and reflective car bodies), competing methods produce depth bleeding or erroneous estimates, whereas LinStereo recovers clean, geometrically consistent depth; (2)~on thin structures such as pipes, poles, and traffic signs, baselines tend to blur or break the structure, while LinStereo preserves continuous, sharp boundaries; and (3)~in low-light or low-texture regions, other methods exhibit large-area depth collapse (\eg, spurious foreground estimates), whereas LinStereo maintains stable predictions. Red and cyan boxes in the figures highlight representative regions where these differences are most pronounced.

\begin{figure*}[t]
\centering
\includegraphics[width=0.8\textwidth]{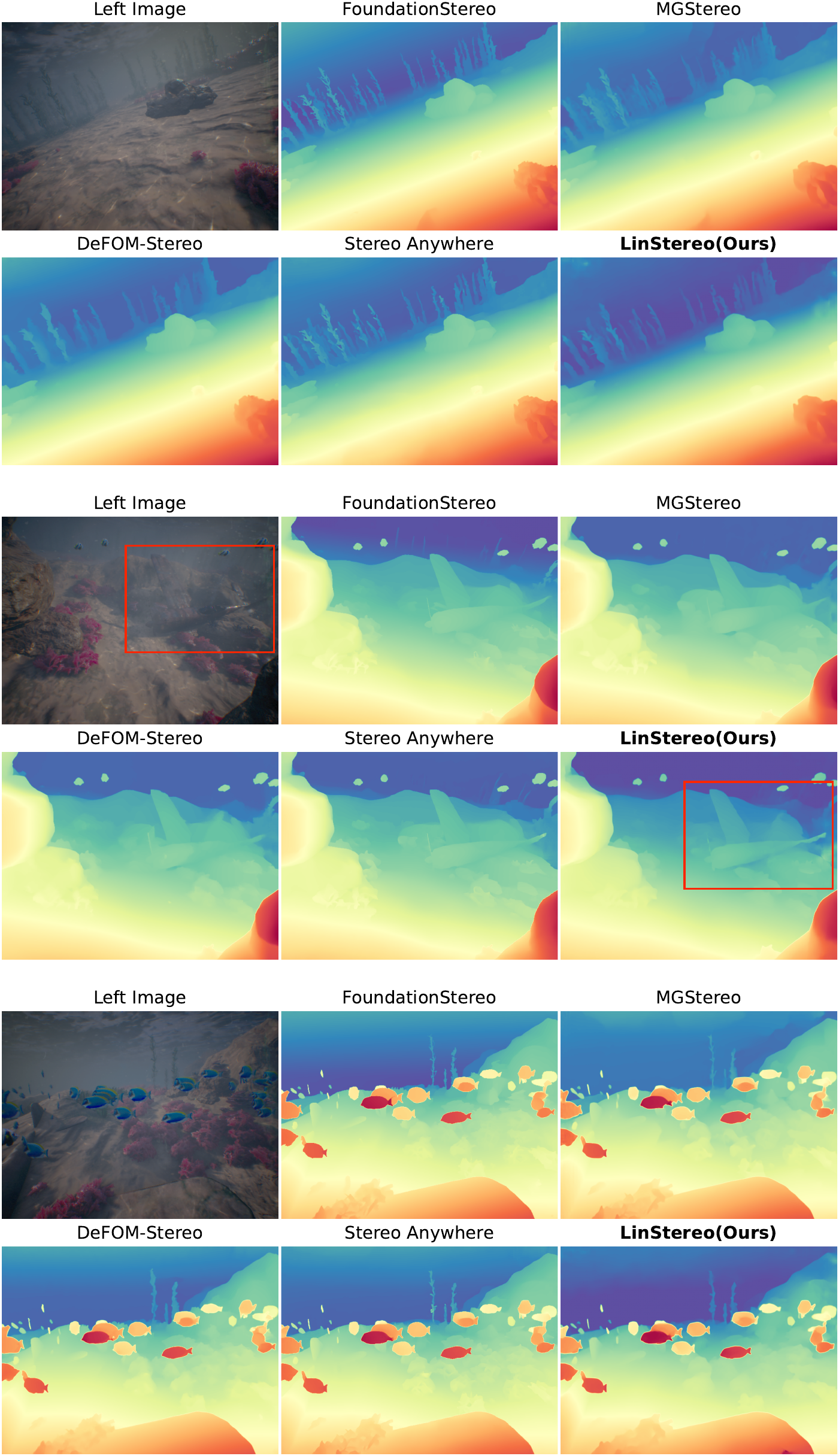}
\caption{\textbf{Additional qualitative comparison on TartanAir-UW (1/2).} Scenes include rocky seafloor terrain, underwater cliffs with vegetation, and coral reefs with fish. Baselines overestimate depth in backscatter-affected distant regions and lose small foreground objects, while LinStereo preserves accurate depth-scale transitions and resolves individual object depth (red boxes).}
\label{fig:qual_tartanair_easy}
\end{figure*}

\begin{figure*}[t]
\centering
\includegraphics[width=0.8\textwidth]{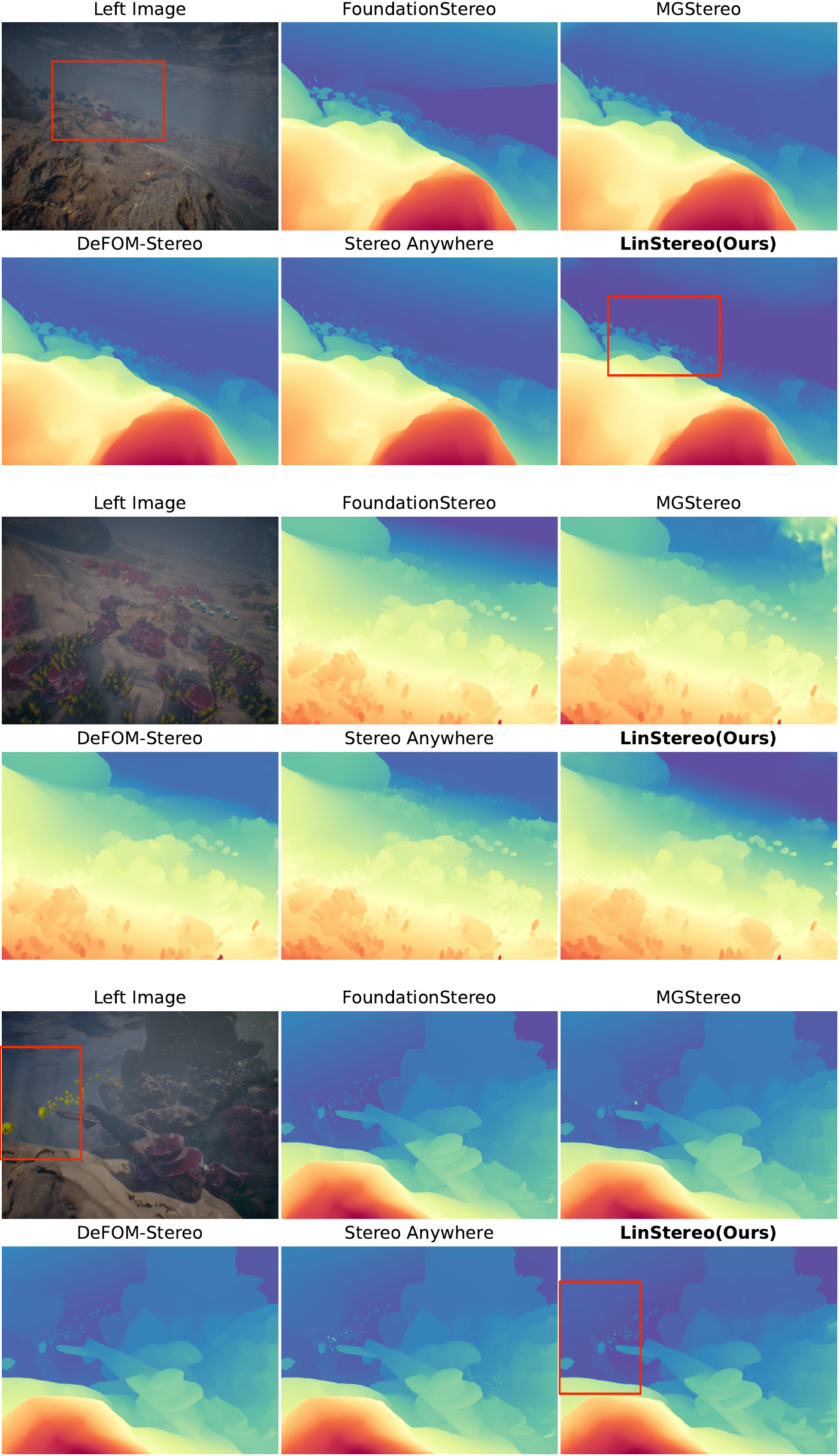}
\caption{\textbf{Additional qualitative comparison on TartanAir-UW (2/2).} Under severe scattering and low visibility, baselines produce depth collapse, blocky artifacts, or over-smoothed predictions that erase terrain and vegetation structure. LinStereo still recovers coherent depth gradients and detects small foreground objects in extremely degraded conditions (red boxes).}
\label{fig:qual_tartanair_hard}
\end{figure*}

\begin{figure*}[t]
\centering
\includegraphics[width=0.8\textwidth]{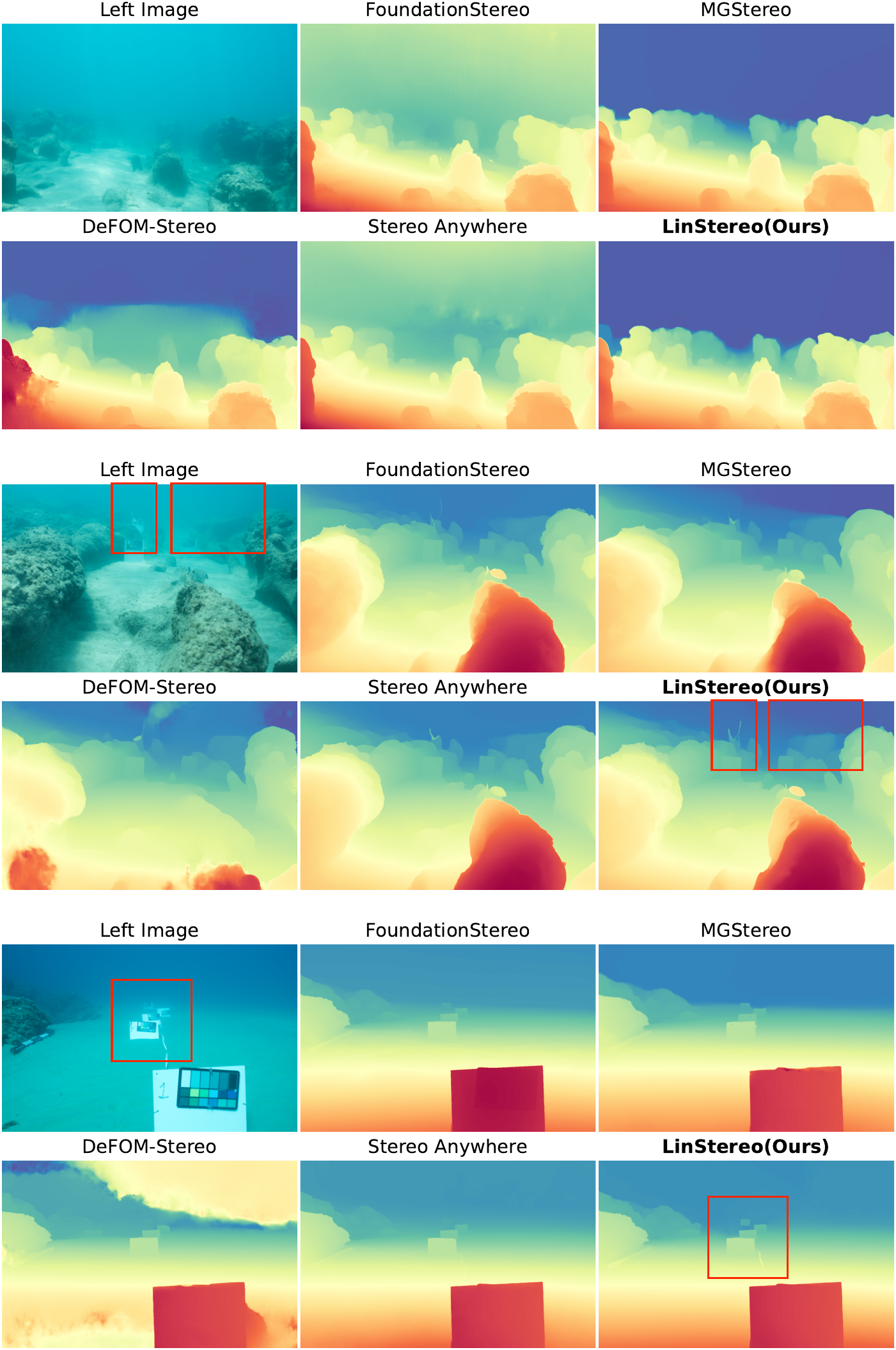}
\caption{\textbf{Additional qualitative comparison on SQUID.} Scenes cover shallow sandy seafloor, near-field coral and rock formations, and man-made equipment. Under strong color attenuation, baselines exhibit depth-scale inconsistencies, over-smooth fine structures, and produce depth bleeding around object boundaries. LinStereo maintains stable depth and recovers fine structural details (red boxes). Black regions denote unavailable depth.}
\label{fig:qual_squid}
\end{figure*}

\paragraph{Underwater Benchmarks.}
Figs.~\ref{fig:qual_tartanair_easy}--\ref{fig:qual_squid} show additional predictions on TartanAir-UW and SQUID. LinStereo's advantages are evident across both datasets: (1)~in backscatter-affected distant regions (TartanAir-UW), baselines overestimate depth or produce large-area depth collapse, whereas LinStereo maintains smooth and accurate near-to-far depth transitions; (2)~under severe scattering and low visibility, baselines yield blocky artifacts or over-smoothed predictions that lose scene structure, while LinStereo still recovers coherent depth gradients for terrain and vegetation; and (3)~in SQUID's near-field scenes with strong color attenuation, baselines exhibit depth-scale inconsistencies and fail to resolve fine structural boundaries such as rock crevices and equipment contours, whereas LinStereo preserves stable, continuous depth (red boxes).

\begin{figure*}[t]
\centering
\includegraphics[width=0.78\textwidth]{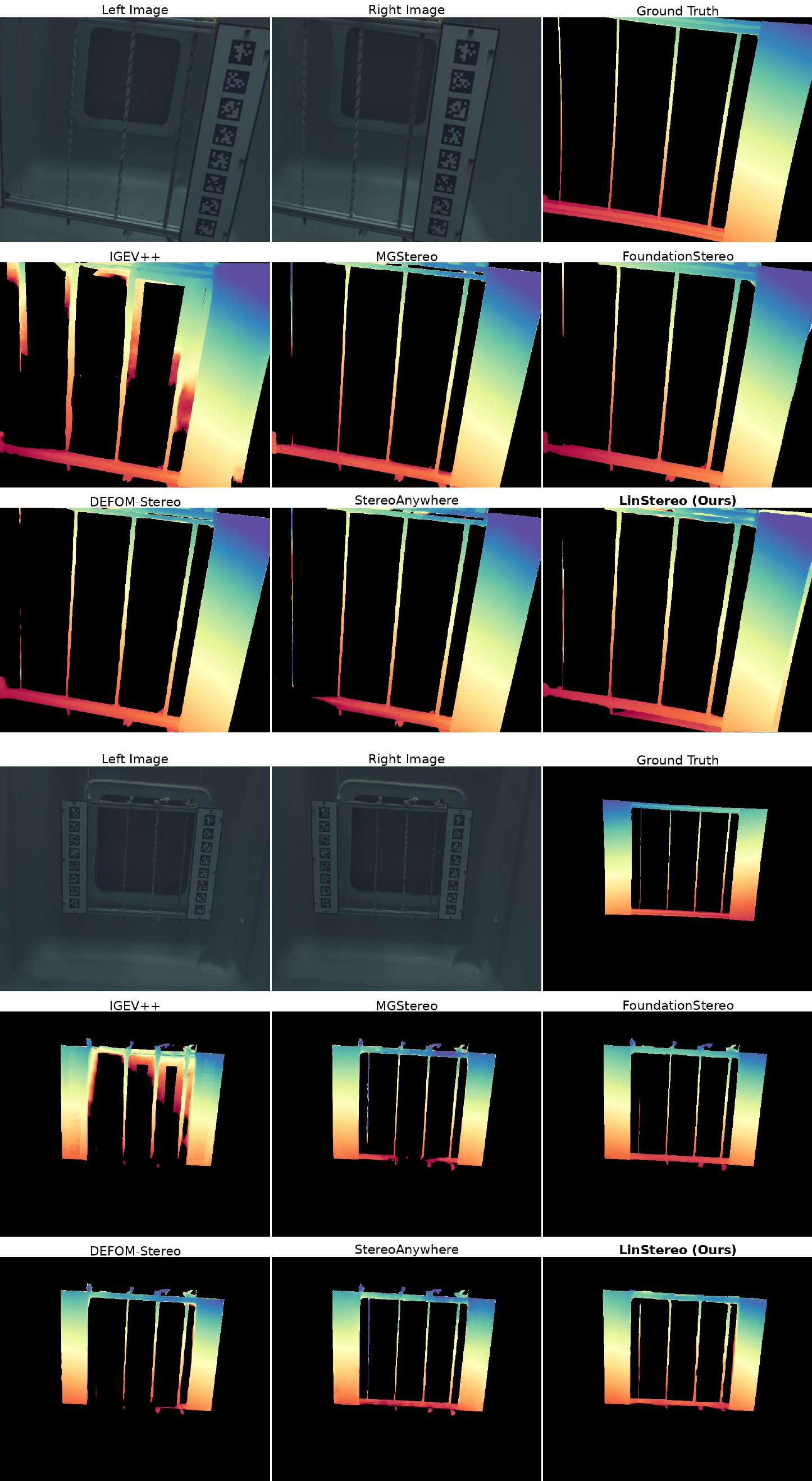}
\caption{\textbf{Qualitative results on the laboratory tank dataset.} Baselines fail to recover the thin ropes and produce incomplete frame geometry. LinStereo recovers continuous rope depth and the complete frame structure under genuine near-range underwater conditions.}
\label{fig:qual_labtank}
\end{figure*}

\paragraph{Laboratory Tank.}
Fig.~\ref{fig:qual_labtank} presents qualitative results from the controlled laboratory tank experiment (see Sec.~\ref{sec:appendix_rope} for dataset construction details). Two key advantages are visible: (1)~on the thin rope segments (${\sim}3$\,mm diameter), baselines either produce severe noise and fragmentation or lose the ropes entirely, whereas LinStereo recovers continuous rope depth; and (2)~for the metal frame structure, baselines fail to fully recognize the frame geometry, resulting in missing or incomplete depth predictions, while LinStereo recovers the complete frame structure.

\begin{figure*}[t]
\centering
\includegraphics[width=0.8\textwidth]{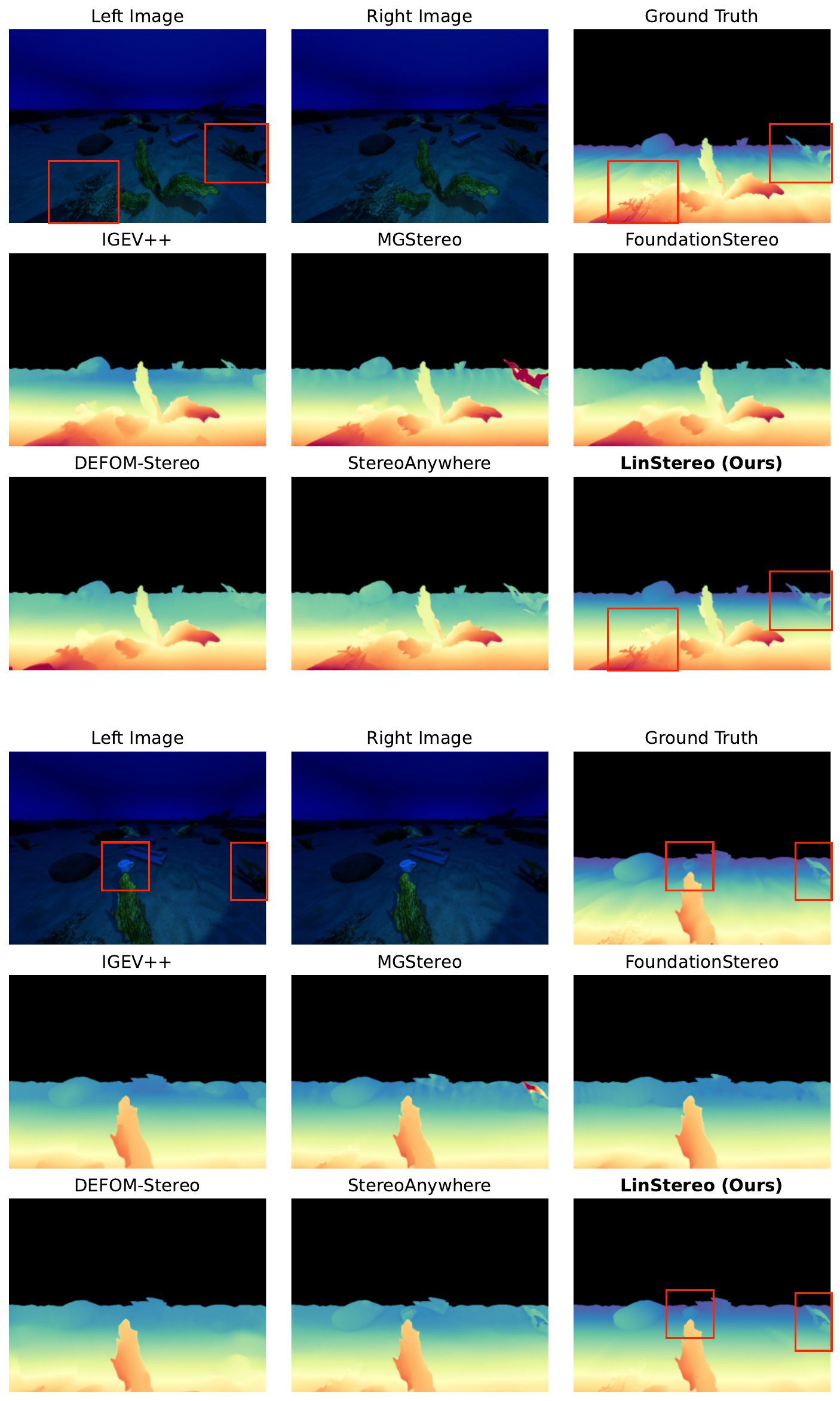}
\caption{\textbf{Qualitative comparison on SeaStereo-Dataset (1/2).} Baselines merge small foreground objects into the seafloor or produce artifacts as turbidity increases. LinStereo cleanly separates foreground objects and maintains consistent depth across visibility conditions (red boxes).}
\label{fig:qual_seasynth_supp_1}
\end{figure*}

\begin{figure*}[t]
\centering
\includegraphics[width=0.8\textwidth]{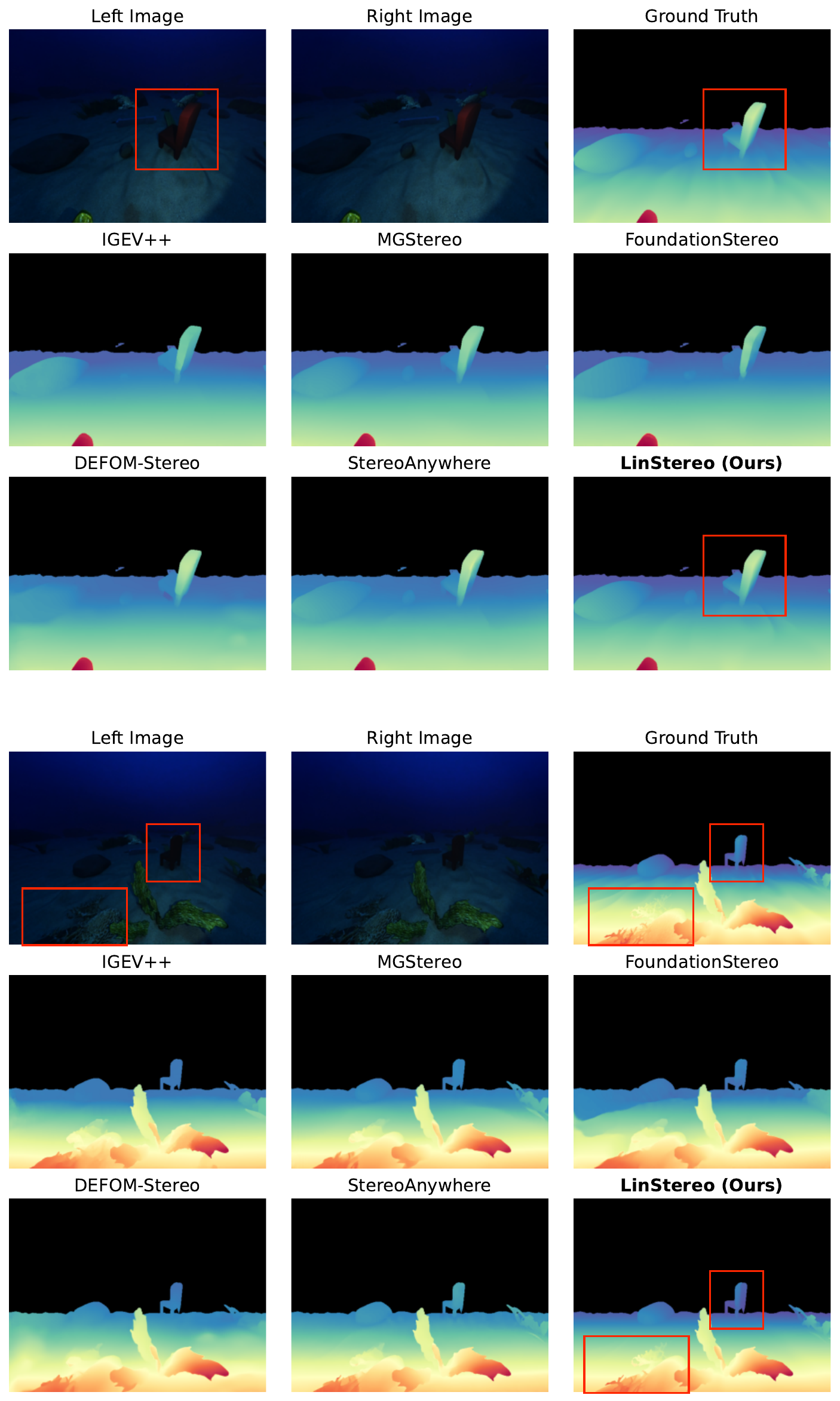}
\caption{\textbf{Qualitative comparison on SeaStereo-Dataset (2/2).} Baselines produce edge noise or shape distortion on elongated structures and over-smooth vegetation, losing depth layering. LinStereo preserves sharp object contours and foreground--background depth separation (red boxes).}
\label{fig:qual_seasynth_supp_2}
\end{figure*}

\paragraph{Synthetic Underwater Data.}
Figs.~\ref{fig:qual_seasynth_supp_1}--\ref{fig:qual_seasynth_supp_2} show LinStereo predictions on randomly selected samples from the SeaStereo-Dataset (Sec.~\ref{sec:appendix_data_synth}), covering clear to highly turbid Jerlov water types. Baselines struggle with small or elongated foreground objects (merging them into the seafloor or distorting their shape) and degrade rapidly as turbidity increases, producing artifacts, unstable depth, or over-smoothed predictions that lose foreground--background layering. LinStereo cleanly separates object depth from the background and maintains consistent depth structure across all water types (red boxes).

\end{document}